\documentclass[wcp]{jmlr}

\usepackage{longtable}% for long tables

\usepackage{booktabs}
\if0
\usepackage{hyperref}
\usepackage{color, colortbl}
\usepackage{url}
\usepackage{amsmath}
\usepackage{amsfonts}
\usepackage{ulem}
\usepackage{breqn}
\usepackage{autobreak}
\usepackage{authblk}
\usepackage{wrapfig}
\usepackage[hang,small,bf]{caption}
\usepackage{pdflscape}
\fi
\usepackage{color, colortbl}
\usepackage{breqn}
\usepackage{algorithm}
\usepackage{algorithmic}
\newcommand{\argmax}{\mathop{\rm arg~max}\limits}

\newtheorem{definition1}{Definition}
\newtheorem{theorem1}{Theorem}

\newtheorem{corollary1}{Corollary}
\newtheorem{lemma1}{Lemma}
\definecolor{LightCyan}{rgb}{0.88,1,1}

\makeatletter
\let\Ginclude@graphics\@org@Ginclude@graphics 
\makeatother

\jmlrvolume{157}
\jmlryear{2021}
\jmlrworkshop{ACML 2021}

\title[Meta-Model-Based Meta-Policy Optimization]{Meta-Model-Based Meta-Policy Optimization}

  \author{\Name{Takuya Hiraoka} \Email{takuya-h1@nec.com}\\
  \addr Central Research Laboratories, NEC Corporation, Kanagawa, Japan
  \AND
  \Name{Takahisa Imagawa} \Email{imagawa.t@aist.go.jp}\\
  \addr Artificial Intelligence Research Center, National Institute of Advanced Industrial Science and Technology, Tokyo, Japan
  \AND
  \Name{Voot Tangkaratt} \Email{voot.tangkaratt@riken.jp}\\
  \addr Center for Advanced Intelligence Project, RIKEN, Tokyo, Japan
  \AND
  \Name{Takayuki Osa} \Email{osa@brain.kyutech.ac.jp}\\
  \addr Department of Human Intelligence Systems, Kyushu Institute of Technology, Fukuoka, Japan \\
  \addr Center for Advanced Intelligence Project, RIKEN, Tokyo, Japan
  \AND
  \Name{Takashi Onishi} \Email{takashi.onishi@nec.com}\\
  \addr Central Research Laboratories, NEC Corporation, Kanagawa, Japan
  \AND
  \Name{Yoshimasa Tsuruoka} \Email{tsuruoka@logos.t.u-tokyo.ac.jp}\\
  \addr Department of Information and Communication Engineering, The University of Tokyo, Tokyo, Japan
 }

\editors{Vineeth N Balasubramanian and Ivor Tsang}

\begin{document}

\maketitle

\begin{abstract}
%-[背景/課題]
Model-based meta-reinforcement learning (RL) methods have recently been shown to be a promising approach to improving the sample efficiency of RL in multi-task settings. 
%-[問題提起]
However, the theoretical understanding of those methods is yet to be established, and there is currently no theoretical guarantee of their performance in a real-world environment. 
%-[研究内容]%--[理論拡張]
In this paper, we analyze the performance guarantee of model-based meta-RL methods by extending the theorems proposed by Janner et al. (2019). 
On the basis of our theoretical results, we propose Meta-Model-Based Meta-Policy Optimization (M3PO), a model-based meta-RL method with a performance guarantee. 
%--[実験]
We demonstrate that M3PO outperforms existing meta-RL methods in continuous-control benchmarks. 
\end{abstract}

\section{Introduction}
\label{sec:intro}
%-[メタ学習＋RL？の導入]
Reinforcement learning (RL) in multi-task problems requires a large number of training samples, and this is a serious obstacle to its practical application~\citep{rakelly2019efficient}. 
%-[RL in multi-task problemsとは]
In the real world, we often want a policy that can solve multiple tasks. 
For example, in robotic arm manipulation~\citep{yu2019meta}, we may want the policies for controlling the arm to ``grasp the wood box,'' ``grasp the metal box,'' ``open the door,'' and so on. 
%-[RL in multi-task problemsでlargenumber of training samplesが必要な理由]
In such multi-task problems, standard RL methods independently learn a policy for individual tasks with millions of training samples~\citep{doi:10.1177/0278364919887447}. 
%-[それがなぜ実問題上やばいのか]
This independent policy learning with a large number of samples is often too costly in practical RL applications. 

%-＊[メタRL重要・役に立つ]
Meta-RL methods have recently gained attention as promising methods to reduce the number of samples required in multi-task problems~\citep{finn2017model}. 
%-[メタ学習とは] 
In meta-RL methods, the structure shared in the tasks is learned by using samples collected across the tasks. 
Once learned, it is leveraged for adapting quickly to new tasks with a small number of samples. 
Various meta-RL methods have previously been proposed in both model-free and model-based settings. 

%-＊[モデルフリーのメタRLはサンプル効率が悪い]
Many \textbf{model-free meta-RL} methods have been proposed, but they usually require a large number of samples to learn a useful shared structure. 
%--[先行研究 Model-free meta-RL methodsのレビュー]
%---[同学習して、同アダプトするのかを先行研究を交えて]
In model-free meta-RL methods, the shared structure is learned as being embedded into the parameters of a context-aware policy~\citep{duan2016rl,mishra2018a,rakelly2019efficient,wang2016learning}, or into the prior of policy parameters~\citep{al2017continuous,finn2017meta,finn2017model,gupta2018meta,rothfuss2018promp,stadie2018some}. 
The policy adapts to a new task by updating context information or parameters with recent trajectories. 
%--[reinforcement learningの課題]
Model-free meta-RL methods have better sample efficiency than independently learning a policy for each task. 
However, these methods still require millions of training samples in total to learn the shared structure sufficiently usefully for the adaptation~\citep{mendonca2019guided}. 

%-＊[モデルベースのメタRLはサンプル効率は良いことが知られている]
On the other hand, \textbf{model-based meta-RL} methods have been demonstrated to be more sample efficient than model-free counterparts. 
%-[Model-based x meta-learning]
In model-based meta-RL methods, the shared structure is learned as being embedded into the parameters of a context-aware transition model~\citep{saemundsson2018meta,perez2020generalized}, or into the prior of the model parameter~\citep{nagabandilearning,nagabandi2018deep}. 
The model adapts to a new task by updating context information or its parameters with recent trajectories. 
The adapted models are used for action optimization in model predictive control (MPC). 
In the aforementioned literature, model-based meta-RL methods have been empirically shown to be more sample efficient than model-free meta-RL methods. 

%-[モデルベースの欠点の指摘]
Despite these empirical findings, these model-based meta-RL methods lack performance guarantees. 
%-[理論保障がないとは]
Here, the performance guarantee means the theoretical property that specifies the relationship between the planning performance in the model and the actual planning performance in a real environment. 
Note that the planning performance in the model is generally different from the actual performance due to the model bias. 
That is, even if planning performs well in the model, it does not necessarily perform well in a real environment. 
%-[理論保障がないことのやばさ]
Such a performance guarantee for model-based meta-RL methods has not been analyzed in the literature. 

\begin{figure*}[t]
\begin{center}
\includegraphics[clip, width=1.0\hsize]{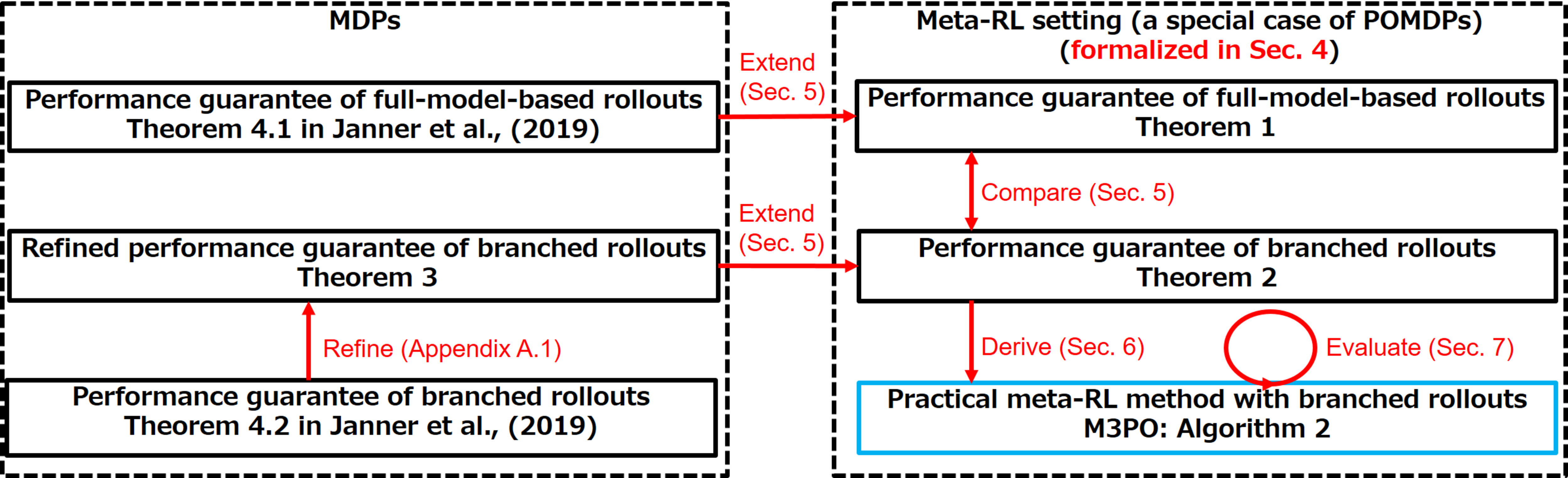}
\end{center}
 %\vspace{-1.2\baselineskip}
  \vspace{-1.8\baselineskip}
\caption{The executive summary of our work. 
Our primary contribution is proposing a model-based meta-RL method (M3PO: Algorithm~\ref{alg1:Meta-MBPO}) with a performance guarantee. 
%-[sec 4]
In Section~\ref{sec:formulatingMBRL}, we formulate our model-based meta-RL setting as solving (a special case of) POMDPs. 
%-[sec 5]
In Section~\ref{sec:monotonic}, we derive the theorems of performance guarantees by extending the theorems proposed in \citet{janner2019trust} into the meta-RL setting. 
We then compare the derived theorems to justify using branched rollouts for model-based meta-RL. 
%-
In Section~\ref{sec:practicalMBPO}, on the basis of our theoretical results, we derive a practical model-based meta-RL method (M3PO: Algorithm~\ref{alg1:Meta-MBPO}). 
In Section~\ref{sec:experiments}, we experimentally evaluate M3PO. 
}
\label{fig:hlsummary}
  \vspace{-1.0\baselineskip}
\end{figure*}

%-[本研究の話mod]
In this paper, we propose a model-based meta-RL method called Meta-Model-based Meta-Policy Optimization (M3PO) which is equipped with a performance guarantee (Figure~\ref{fig:hlsummary}). 
Specifically, we develop M3PO by firstly formulating a meta-RL setting on the basis of partially observable Markov decision processes (POMDPs) in Section~\ref{sec:formulatingMBRL}. 
Then, in Section~\ref{sec:monotonic}, we conduct a theoretical analysis to justify the use of branched rollouts, which is a promising model-based rollout method proposed by \citet{janner2019trust}, in the meta-RL setting. 
We compare the theorems for performance guarantees of the branched rollouts and full-model based rollouts, which is a naive baseline model-based rollout method, in the meta-RL setting. 
Our comparison result shows that the performance of the branched rollout is more tightly guaranteed than that of the full model-based rollouts. 
%deppre We derive the theorems of performance guarantees by extending the theorems proposed in \citet{janner2019trust} into the meta-RL setting. 
%deppre We then compare the derived theorems to justify using branched rollouts for model-based meta-RL. 
%deppre\emph{the one-step branched rollout can reduce the discrepancy more significantly than the full model-based rollout}
Based on this, in Section~\ref{sec:practicalMBPO} we derive M3PO, a practical method that uses the branch rollouts. 
Lastly, in Section~\ref{sec:experiments} we experimentally demonstrate the performance of M3PO in continuous-control benchmarks. 
%
%deppre The executive summary of our work is shown in Figure~\ref{fig:hlsummary}. 
%

We make four primary contributions in both theoretical and practical frontiers.\\
\textbf{Theoretical contributions:} 
(1) We provide the performance guarantee for model-based meta-RL (Section~\ref{sec:monotonic}). To the best of our knowledge, our work is the first attempt to provide a performance guarantee for model-based meta-RL. 
(2) We conduct theoretical analysis and provide the insight that \textit{short-steps branched rollouts are better than full-model based rollouts} (Section~\ref{sec:monotonic} and Appendix~\ref{sec:appen}). 
We refine analyses in \citet{janner2019trust} by considering multiple-model-based rollout factors (Appendix~\ref{sec:boundinmdp}), and extend our analyses results into a meta-RL setting (Section~\ref{sec:monotonic}). 
%de Some readers may concern that \citet{janner2019trust} have already provided similar insight in the MDP case. 
%de However, we refine their analyses by considering the important premise (multiple-model-based rollout factors) ignored in their work (Appendix~\ref{sec:boundinmdp}), and extend our analyses results into a meta-RL setting (Section~\ref{sec:monotonic}). 
Therefore, our work complements and strengthens their analyses and insight.\\
%理論解析：
%１．性能保証を付けた。モデルベースでのメタ方策の中では最初のもの。
%２．理論解析により、（我々のメタRL）短い分枝ロールアウトが有効であることを示した。
%この理論的な結果はMDPにおけるJannerらの分枝ロールアウトと概ね同じであるが、
%我々はJannerらのMDP解析で考慮されていなかった重要な前提（multi-rollout factor)を考慮し再解析し、それでも彼らの帰結は概ね変化しないことを示し、彼らの解析を補強した（Appendix）.
%また、同様の結果はメタRLにも適用できることを示した。
%
\textbf{Practical contributions:} 
(3) We propose a practical model-based RL method (M3PO) for meta-RL (Section~\ref{sec:practicalMBPO}) and demonstrate that it outperforms existing meta-RL methods (Section~\ref{sec:experiments}). 
Notably, we demonstrate that M3PO can work successfully even on complex tasks (e.g., Humanoid-direc). 
%deppre (4) Our experiments demonstrate that our proposed method (M3PO) outperforms existing meta-RL methods, and provides practically interesting insights (Section~\ref{sec:practicalMBPO}) (Section~\ref{sec:experiments}). 
%deppre The first insight is that, in our setting, the one-step branched rollout is better than N-steps (N > 1) branched rollouts. 
%deppre This is different from the insight of XX, in which N-steps (N > 1) branched rollouts can be better than one-step branched rollout. 
(4) We provide a practically interesting finding that, in M3PO (Dyna-style RL method), dynamically adjusting the mixture ratio of real/fictitious training data substantially improves long-term learning performance (Section~\ref{sec:compari} and Appendix~\ref{sec:m3po-h}). 
In existing Dyna-style RL methods (e.g. \cite{janner2019trust,NEURIPS2020_1dc3a89d,NEURIPS2020_a322852c}), this mixture ratio is fixed during the training phase. 
Our finding promotes assessing the effect of dynamically adjusting the mixture ratio for the other Dyna-style RL methods to improve these methods. 
%経験的な解析：
%deppre １．我々の手法が有用であることを示し、これまで明らかになっていなかった実用上興味深い知見を得た。
%JannerらのMDPでの実験的な解析にでは明らかではなかった知見を得られた。
%２．我々のめたRL の設定ではk=1のときが効果的であり、kを増大させるとことを示した。（Jannerらは環境に応じて１～25）。
%３．最終的な疑似ロールアウトと実データの動的な比率がより重要であることを例示した（Dynaベースの手法では、基本的に比率は固定しているが、動的にやったほうがいいことを奨励している）。
%}

\section{Related work}\label{sec:relatedworks}
%[Meta RL]
Meta-RL is a popular approach for solving multi-tasks RL problems. 
Recently, researchers have formulated meta-RL as solving special cases of POMDPs~\citep{duan2016rl,humplik2019meta,zintgraf2020varibad}. 
However, the performance guarantee of meta-RL under the POMDPs framework has not been established. %deppre , and only the guarantee of standard RL methods has been established~\citep{ghavamzadeh2015bayesian,sun2019towards,sun2019model}. 
In this paper, we presents bounds of the performance of meta-RL under the POMDPs framework. 

%[MBRL]
We derive the above-mentioned performance bound in the context of model-based meta-RL. Existing works have focused on the performance bound of model-based RL~\citep{feinberg2018mbve,henaff2019explicit,janner2019trust,luo2018algorithmic,rajeswaran2020game}, while ignoring model-based meta-RL. 
%deppre Existing works have been focused on the performance bound of model-based RL~\citep{feinberg2018mbve,henaff2019explicit,janner2019trust,luo2018algorithmic,rajeswaran2020game}, while ignoring model-based meta-RL. 
Specifically, in existing works, Markov decision processes (MDPs) are assumed, and a meta-RL setting is not discussed.

\section{Preliminaries}\label{sec:preliminaries}
%-[meta-RL]
\subsection{Meta-reinforcement learning} 
%-[一般的なmeta-RLとは]
Meta-RL aims to learn the structure shared in tasks~\citep{finn2017meta,nagabandilearning}. 
%-[メタ学習の説明]%\textcolor{red}{(メタ学習の説明：メタ学習の概要のレビュー)}
Here, a task is defined as a MDP $\langle \mathcal{S}, \mathcal{A}, p_{\text{st}}, r, \gamma \rangle$ with a state space $\mathcal{S}$, an action space $\mathcal{A}$, a transition probability $p_{\text{st}} : \mathcal{S} \times \mathcal{S} \times \mathcal{A} \rightarrow [0,1]$, a reward function $r: \mathcal{S} \times \mathcal{A} \rightarrow \mathbb{R}$, and a discount factor $\gamma \in [0,1)$. 
We assume that there are infinitely many tasks with the same state and action spaces but different transition probabilities and reward functions. 
%deppreHere, a task specifies the transition probability and the reward function. 
Information about task identity cannot be observed by the agent. 
Hereinafter, the task identity and its set are denoted by $\tau$ and $\mathcal{T}$, respectively. 

%--meta rl
A meta-RL process is composed of meta-training and meta-testing. 
In meta-training, the structure shared in the tasks is learned as being embedded in either or both of a policy and a transition model. 
In meta-testing, on the basis of the meta-training result, they adapt to a new task. 
For the adaptation, the trajectory observed from the beginning of the new task to the current time step is leveraged. 

\subsection{Partially observable Markov decision processes}
%[POMDP]
The POMDP framework extends the MDP framework by assuming that the state itself is hidden, and the agent receives an observation instead of a state. 
Formally, a POMDP is defined as a tuple $\langle \mathcal{O}, \mathcal{S}, \mathcal{A}, p_{\text{ob}}, r, \gamma, p_{\text{st}} \rangle$. 
Here, $\mathcal{O}$ is an observation space, and $p_{\text{ob}} : \mathcal{O} \times \mathcal{S} \times \mathcal{A} \rightarrow [0, 1]$ is the observation probability. 
%deppre A POMDP is defined as a tuple $\langle \mathcal{O}, \mathcal{S}, \mathcal{A}, p_{\text{ob}}, r, \gamma, p_{\text{st}} \rangle$. 
%deppre Here, $\mathcal{O}$ is a set of observations, $\mathcal{S}$ is a set of hidden states, $\mathcal{A}$ is a set of actions, $p_{\text{ob}} = \mathcal{O} \times \mathcal{S} \times \mathcal{A} \rightarrow [0, 1]$ is the observation probability, $p_{\text{st}} = \mathcal{S} \times \mathcal{S} \times \mathcal{A} \rightarrow [0, 1]$ is the state transition probability, $r : \mathcal{S} \times \mathcal{A} \rightarrow \mathbb{R}$ is a reward function, and $\gamma \in [0, 1)$ is a discount factor. 
At time step $t$, the functions contained in POMDP are used as $p(s_t |s_{t-1}, a_{t-1})$, $p(o_t| s_t,~ a_{t-1})$, and $r_t = r(s_t,~ a_t)$. 
%deppre At time step $t$, these functions are used as $p(s_t |s_{t-1}, a_{t-1})$, $p(o_t| s_t,~ a_{t-1})$, and $r_t = r(s_t,~ a_t)$. 
%deppre The agent cannot directly observe the hidden state but receives the observation instead. 
%
The agent selects an action on the basis of a policy $\pi(a_{t} | h_t)$. Here, $h_t$ is a history (of the past trajectory) defined as $h_t = \left\{o_0, a_0, ..., o_t \right\}$. 
We denote the set of the histories by $\mathcal{H}$. 
Given the definition of the history, the history transition probability $p(h_{t+1} | a_{t}, h_t)$ can be evaluated by using $p(o_{t+1} | a_t, h_{t})$. 
Here, $p(o_{t+1}| a_t, h_{t}) = \sum_{s_{t+1}} \sum_{s_t}p(s_t | h_t) p(s_{t+1}|s_t, a_t)p(o_{t+1} | s_{t+1}, a_t)$. 
%- model based RL 
The goal of RL in the POMDP is to find the optimal policy $\pi^*$ that maximizes the expected return: $\pi^*= \argmax_{\pi} \mathbb{E}_{a\sim\pi, h\sim p} \left[ R \right]$, where $R = \sum_{t=0}^{\infty} \gamma^t r_t$. 

\subsection{Branched rollouts~\citep{janner2019trust}}
%--[Branched rolloutsとは]
Branched rollouts are Dyna-style rollouts~\citep{sutton1991dyna}, in which model-based rollouts are run as being branched from real trajectories. 
It is seemingly similar to the MPC-style rollout used in the previous model-based meta-RL methods~\citep{saemundsson2018meta,perez2020generalized,nagabandilearning,nagabandi2018deep}. 
However, it can be used for off-policy optimization with respect to an infinite planning horizon, which is a strong advantage over the MPC-style rollout. 
A theorem for the performance guarantee of the branched rollout in MDPs are provided as Theorem 4.2  in \citet{janner2019trust}. 
We refer to the formal statement of the theorem in Appendix~\ref{sec:boundinmdp}. 

\section{Formulating model-based meta-RL}\label{sec:formulatingMBRL}
%[導入]%[話の概要]
In this section, we formulate model-based meta-RL as a special case of POMDPs. 
%
%-[(state) タスク、タスク遷移、新しいタスクでの学習]
In our formulation, the hidden state is assumed to be composed by the task and observation: $\mathcal{S}=\mathcal{T} \times \mathcal{O}$. 
%deppre Here $\mathcal{T}$ is the set of task $\tau$, which cannot be observed by the agent. 
%
%-[functions]
With this assumption, the transition probability, observation probability, and reward function can be written as follows:  $p(s_{t+1}|s_t, a_t) = p(\tau_{t+1}, o_{t+1} | \tau_{t}, o_{t}, a_t)$, $p(o_{t+1}|s_{t+1}, a_t) = p(o_{t+1} | \tau_{t+1}, o_{t+1}, a_t) = 1$ and $r(s_t, a_t) = r(\tau_t, o_t, a_t)$. 
%- assumption 
As in \citet{nagabandilearning,nagabandi2018deep}, we assume that the task can change during an episode (i.e., the value of $\tau_{t+1}$ is not necessarily equal to that of $\tau_t$). 
In addition, following previous studies~\citep{finn2017meta,finn2017model,rakelly2019efficient}, we assume that the task set $\mathcal{T}$ and the initial task distribution $p(\tau_0)$ do not change in meta-training and meta-testing. 
Owing to this assumption, in our analysis and method, meta-training and meta-testing can be seen as identical. 

%-[定式化]\textcolor{red}{(meta-policyとmodelの設定のモデル化)}
We define a parameterized policy and model as $\pi_{\phi}(a_{t} | h_{L:t})$ and $p_{\theta}(r_{t}, o_{t+1} | a_t, h_{L:t})$, respectively. 
Here, $\phi$ and $\theta$ are learnable parameters, 
%-[Truncating history]
and $h_{L:t}$ is the truncated history $h_{L:t} = \left\{o_{\max(t-L,0)}, a_{\max(t-L,0)}, ..., o_t \right\}$, where $L$ is a hyper-parameter. 
As with \citet{nagabandilearning}, % we assume that, for the agent, the current task has been invariant for the last $L$-time steps, where $L$ is a hyper-parameter, and use the history for this period for policy and model adaptation. 
we use the history for this period for policy and model adaptation. % \footnote{Similar truncation approaches have been introduced to limit the complexity of POMDPs in Blocked MDP literature~\citep{du2019provably}.}. %,pmlr-v119-zhang20t}.}. 
In addition, we assume that $r_{t}$ and $o_{t+1}$ are conditionally independent given $a_t$ and $h_{L:t}$, i.e., $p_{\theta}(r_{t}, o_{t+1} | a_t, h_{L:t}) = p_{\theta}(r_{t}| a_t, h_{L:t}) \cdot p_{\theta}(o_{t+1} | a_t, h_{L:t})$. 
For analysis in the next section, we use the model for the history $p_{\theta}(h_{t+1} | a_{t}, h_{t})$ that can be evaluated by using $p_{\theta}(o_{t+1} | a_t, h_{L:t})$. 

%[モデルの使い方について]\textcolor{red}{(本研究で対象とするタイプのアルゴリズム)}
We use the parameterized model and policy as shown in Algorithm~\ref{alg1:MonotonicMeta-MBPO}. 
This algorithm is composed of 1) training the model, 2) collecting trajectories from the real environment, and 3) training policy to maximize the model return $\mathbb{E}_{(a,h) \sim m(\pi_\phi, p_\theta, \mathcal{D}_{\text{env}})} \left[ R \right]$. 
Here, $m(\pi_\phi, p_\theta, \mathcal{D}_{\text{env}})$ is the history-action visitation probability based upon an abstract model-based rollout scheme, which can be calculated on the basis of $\pi_\phi$, $p_\theta$, and $\mathcal{D}_{\text{env}}$. 
In the next section, we will analyze the algorithm equipped with two specific model-based rollout schemes: the full-model-based rollout and branched rollout. 

\begin{algorithm}[t]
\caption{Model-based meta-RL}
\label{alg1:MonotonicMeta-MBPO}
\begin{algorithmic}[1]
\STATE Initialize policy $\pi_{\phi}$, model $p_{\theta}$, environment dataset $\mathcal{D}_{\text{env}}$
\FOR{$N$ epochs}
 \STATE Train $p_\theta$ with $\mathcal{D}_{\text{env}}$ 
 \STATE Collect trajectories from environment in accordance with $\pi_{\phi}$: $\mathcal{D}_{\text{env}} = \mathcal{D}_{\text{env}} \cup \left\{ \left( h_{t}, a_t, o_{t+1}, r_{t}\right) \right\}$
 \STATE Train $\pi_{\phi}$ to maximize $\mathbb{E}_{(a,h) \sim m(\pi_\phi, p_\theta, \mathcal{D}_{\text{env}})} \left[ R \right]$ 
\ENDFOR
\end{algorithmic}
\end{algorithm}

\begin{figure}[t]
\begin{center}
\includegraphics[clip, width=0.5\hsize]{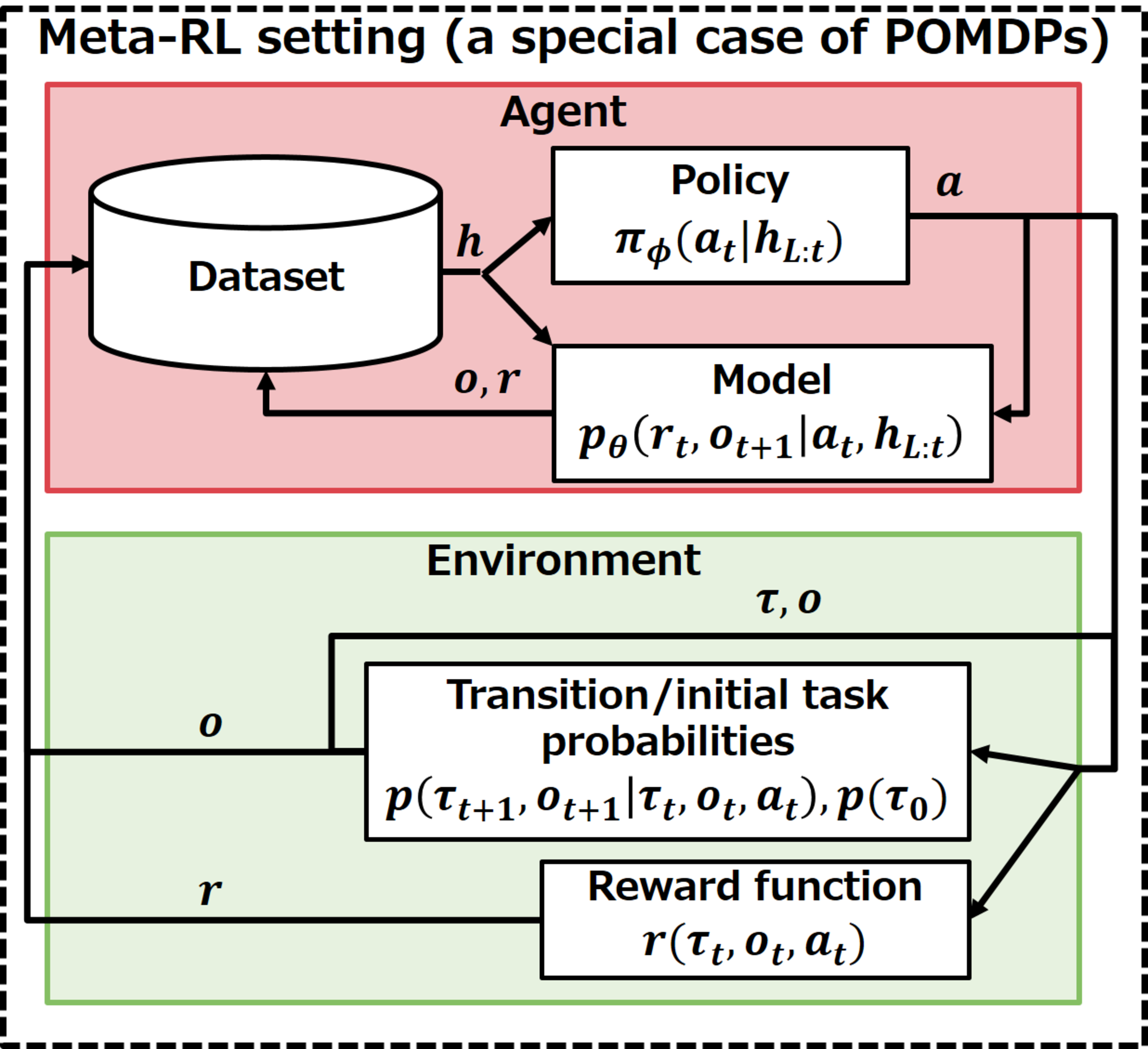}
\end{center}
 %\vspace{-1.2\baselineskip}
   \vspace{-1.8\baselineskip}
\caption{
Model-based meta-RL formulation via POMDPs. 
Here, $\tau$ is a task, $o$ is an observation, $h$ is a (truncated) history, $a$ is an action, and $r$ is a reward. 
}
\label{fig:ourformulation}
  \vspace{-1.0\baselineskip}
\end{figure}

%-[summary]
Our model-based meta-RL formulation via POMDPs introduced in this section is summarized in Figure~\ref{fig:ourformulation}. 
In the following sections, we present theoretical analysis and empirical experiments on the basis of this formulation. 

\section{Performance guarantees}\label{sec:monotonic}
%-[intro]
In the previous section, we formulate our meta-RL setting. 
In this section, we justify the use of the branched rollout in the meta-RL setting. 
To do so, we show the performance guarantee of the branched rollout can be tighter than that of full-model based rollout in the meta-RL setting. %deppre, and show that the former can be tighter than the latter. 

%[Branched rolloutの性能保証の説明]%-[性能保証とは]
The performance guarantees is defined as ``planning performance in a real environment $\geq$ planning performance in the model -- discrepancy.'' 
Formally, it is represented as: 
\vspace{-0.2\baselineskip}\begin{dmath}
\mathbb{E}_{a \sim \pi_\phi, h \sim p} \left[ R \right] \geq \mathbb{E}_{(a,h) \sim m(\pi_\phi, p_\theta, \mathcal{D}_{\text{env}})} \left[ R \right] - C(\epsilon_m, \epsilon_\pi).
\end{dmath}\vspace{-0.2\baselineskip}
Here, $\mathbb{E}_{a \sim \pi_\phi, h \sim p}\left[ R \right]$ is a true return (i.e., the planning performance in the real environment). 
$\mathbb{E}_{(a,h) \sim m(\pi_\phi, p_\theta, \mathcal{D}_{\text{env}})} \left[ R \right]$ is a model return (i.e., the planning performance in the model). 
%
%-[Cとそれを計算するためのエラーの定義]
$C(\epsilon_m, \epsilon_\pi)$ is the discrepancy between the returns, which can be expressed as the function of two error quantities $\epsilon_m$ and $\epsilon_\pi$. 
%-[二つのエラーの定義]
%\vspace{-0.2\baselineskip}
\begin{definition1}[Model error]\label{def:epsm} A model error of model $p_{\theta}$ is defined as \\
$\epsilon_m = \max_t \mathbb{E}_{a_{t} \sim \pi_{\mathcal{D}}, h_{t} \sim p} \left[ D_{TV}\begin{pmatrix} p || p_{\theta}) \end{pmatrix} \right]$. 
Here, $D_{TV}(p || p_{\theta})$ is the total variation distance between $p(h_{t+1} | a_{t}, h_{t})$ and $p_{\theta}(h_{t+1} | a_{t}, h_{t})$. 
$\pi_{\mathcal{D}}$ is the data-collection policy. %deppre , such that the trajectory distribution generated under $\pi_{\mathcal{D}}(a_{t} | h_{t})$ and $p(h_{t+1} | a_{t}, h_{t})$ is equal to the one in $\mathcal{D}_{\text{{\normalfont env}}}$. 
\end{definition1}%\vspace{-0.2\baselineskip}
%
%\vspace{-0.3\baselineskip}
\begin{definition1}[Maximal policy divergence]\label{def:epspi}
The maximal policy divergence between $\pi_D$ and $\pi_\phi$ is defined as $\epsilon_\pi = \max_{h_t} D_{TV}\left( \pi_{\mathcal{D}} || \pi_{\phi} \right)$. 
Here, $D_{TV}\left( \pi_{\mathcal{D}} || \pi_{\phi} \right)$ is the total variation distance of $\pi_{\mathcal{D}}(a_{t} | h_{t})$ and $\pi_{\phi}(a_{t} | h_{L:t})$. 
\end{definition1}%\vspace{-0.4\baselineskip}

%-[outline]
In the following, we firstly provide the performance guarantees in the meta-RL setting by deriving the discrepancies $C(\epsilon_m, \epsilon_\pi)$ of the full model-based and branched rollouts.
Then, we analyze these discrepancies and show that the discrepancy of the branched rollout can yield a tighter performance guarantee than that of the full model-based rollout. 

\subsection{Performance guarantee of the full-model-based rollout}\label{sec:thfulmodel}
%[Full model-based rolloutとは]
In the full-model-based rollout scheme, model-based rollouts are run from the initial state to the infinite horizon without any interaction with the real environment. 
%deppre The full-model-based rollout is the model-based rollout scheme in which model-based rollouts are run from the initial state to the infinite horizon. 
The performance guarantee of model-based meta-RL with the full-model-based rollout scheme is as follow: 
%-[boud full-model-based rollouts]%-[Janner's theorem]
\begin{theorem1}\label{th:1pomdp}
Under the full-model-based rollouts, the following inequality holds, 
%\vspace{-0.2\baselineskip}
\begin{equation}
\mathbb{E}_{a \sim \pi_\phi, h \sim p}\left[ R \right] \geq \mathbb{E}_{(a,h) \sim m_{f}(\pi_\phi, p_\theta, \mathcal{D}_{\text{env}})}[R] - C_{\text{{\normalfont Th}}1}(\epsilon_m, \epsilon_\pi). \nonumber
\end{equation}%\vspace{-0.2\baselineskip}
\end{theorem1}
Here $\mathbb{E}_{(a,h) \sim m_{f}(\pi_\phi, p_\theta, \mathcal{D}_{\text{env}})} \left[ R \right]$ is a model return in the full model-based rollout scheme. 
$C_{\text{Th}1}(\epsilon_m, \epsilon_\pi)$ is the discrepancy between the returns, and its expanded representation is shown in Table~\ref{tab:discrepancies}. 
$r_{max}$ is the constant that bounds the expected reward: \\ $r_{max} > \max_{a_t, h_t} \left| \sum_{\tau_t} p(\tau_t | h_t) r(\tau_t, o_t, a_t) \right|$. 
The proof of Theorem~\ref{th:1pomdp} is given in Appendix~\ref{sec:proofs}, where we extend the result of \citet{janner2019trust} to the meta-RL setting.

\if0
%-[本節の概要] (introと同じ)
Next, we conduct a theoretical analysis to justify the use of the branched rollout in the meta-RL setting. 
%--[なに？]
In section~\ref{subsec:inter-mbrlandmfrl}, we extend the notion of the branched rollout proposed in \citet{janner2019trust} into the meta-RL setting and provide a theorem to guarantee performance of the extended branched rollout. 
In section~\ref{sec:thfulmodel}, we compare the theorem with that for a full-model-based rollout. 
The comparison results indicate that, in the meta-RL setting, the performance in the branched rollout is more tightly guaranteed than that in the full model-based rollout. 
\fi
%
%-Summary of discrepancies in the theorems 
\begin{table*}\caption{Return discrepancies in our theorems in Section \ref{sec:monotonic}. 
The discrepancies measure how much different the model return is from the true return. 
A low discrepancy means that the difference between them is small and the performance is tightly guaranteed. 
$C_{\text{Th}1}$ and $C_{\text{Th}2}$ are discrepancies in Theorems~\ref{th:1pomdp} and \ref{th:2pomdp}, respectively.}\label{tab:discrepancies}
 \vspace{-0.5\baselineskip}
  \vspace{-0.7\baselineskip}
\hrulefill
%
%-[バウンドの証明]
\begin{equation}
C_{\text{Th}1} = r_{\text{max}} \left\{ \frac{2 \gamma \left(\epsilon_m + 2 \epsilon_\pi \right)}{(1-\gamma)^{2}} + \frac{4 \epsilon_\pi}{(1 - \gamma)} \right\} \nonumber
\end{equation}\vspace{-0.2\baselineskip}
\hrulefill
%
%\begin{equation}
%\begin{align}\begin{autobreak}
\begin{dmath*}
C_{\text{Th}2} =  r_{\text{max}} \left\{ \frac{1+\gamma^2}{(1-\gamma)^2} 2 \epsilon_\pi + \frac{\gamma-k\gamma^{k} + (k-1) \gamma^{k+1}}{(1-\gamma)^2} \left(\epsilon_\pi + \epsilon_m \right) + \frac{\gamma^{k} - \gamma}{\gamma - 1} (\epsilon_{\pi} + \epsilon_m) + \frac{\gamma^k}{1-\gamma} (k+1) (\epsilon_{\pi} + \epsilon_m) \right\} \nonumber
\end{dmath*}
%\end{autobreak}\end{align}
%\end{equation}
%
\hrulefill
\end{table*}

\subsection{Performance guarantee of the branched rollout}\label{subsec:inter-mbrlandmfrl}
\begin{figure}[t]
\begin{center}
\includegraphics[clip, width=0.6\hsize]{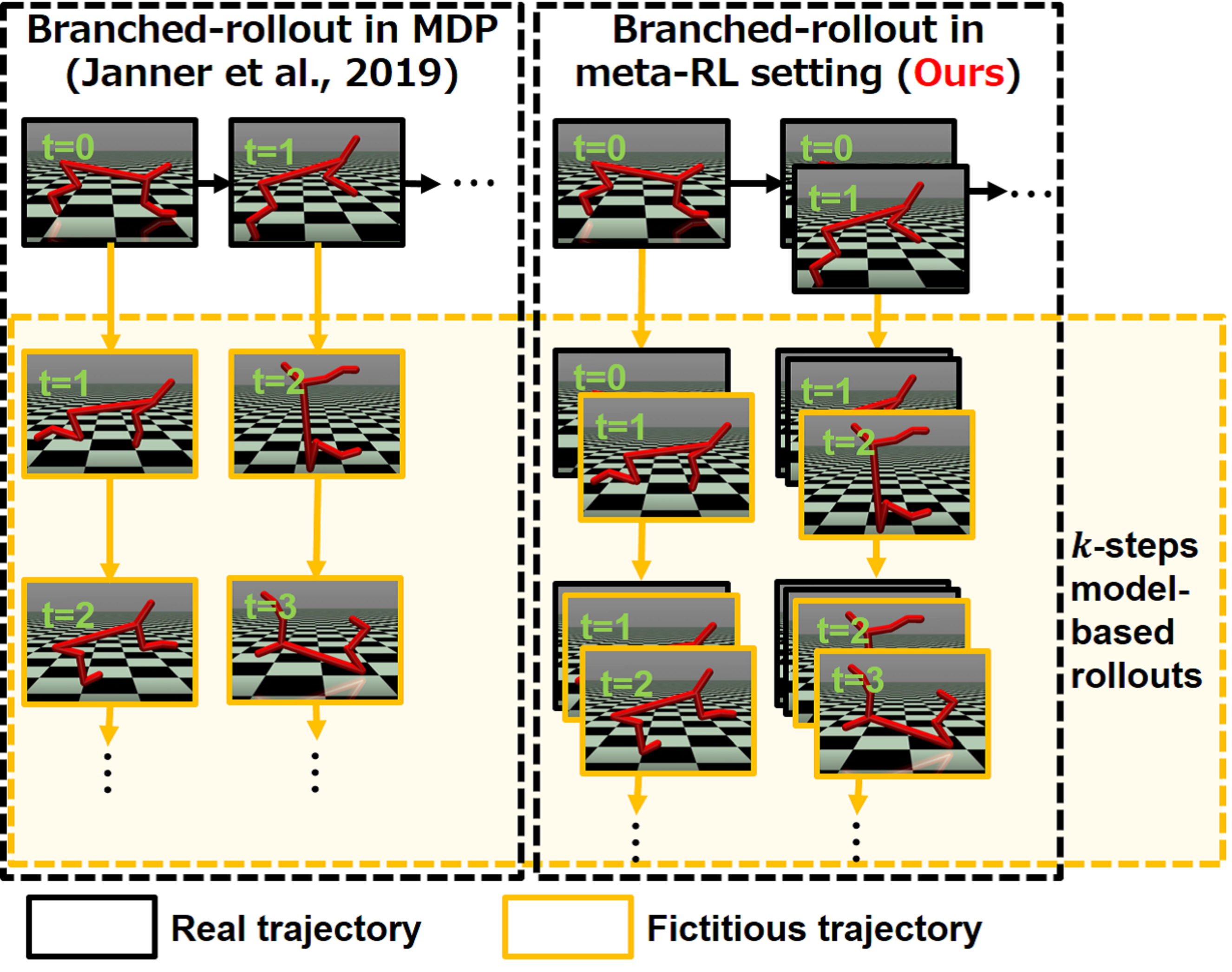}
\end{center}
 %\vspace{-1.2\baselineskip}
   \vspace{-1.8\baselineskip}
\caption{The branched rollouts in a MDP and the meta-RL setting. 
%-[Branched rollouts in MDPの説明]
The branched rollout~\citep{janner2019trust} is a kind of Dyna-style rollouts~\citep{sutton1991dyna}, in which $k$-step model-based rollouts are run as being branched from real trajectories. 
The fictitious trajectories generated by the model-based rollouts are used for policy optimization. 
%-[Branched rollouts in POMDP]
In \citet{janner2019trust}, branched rollouts are defined on a MDP (i.e., the state-action space). 
We extend it to a meta-RL setting (the history-action space). 
}
\label{fig:branchedrollouts}
  \vspace{-1.0\baselineskip}
\end{figure}

%[導入]
Next, we provide a theorem for the performance guarantee of the branched rollout in the meta-RL setting, where $k$-step model-based rollouts are run as being branched from real trajectories (see Figure~\ref{fig:branchedrollouts}). 
%deppre Next, we provide a theorem for the performance guarantee of the branched rollout in the meta-RL setting. 

%[Branched rolloutの拡張](intro)
%-[Branched rolloutとは]%-[Janner etl al ]
To provide our theorem, we extend the notion and theorem of the branched rollout proposed in \citet{janner2019trust} into the meta-RL setting. 
Specifically, we extend the branched rollout defined originally in the state-action space~\citep{janner2019trust} to a branched rollout defined in the history-action space (Figure~\ref{fig:branchedrollouts}). 
%[既存の定理の拡張方針]：
Then, we translate our meta-RL setting into equivalent MDPs with histories as states, and then utilize the theorem to obtain a result in the meta-RL setting (more details are given in Appendix~\ref{sec:proofs}). 

%[Theorem4.2の拡張結果（Theorem ~）]
Our resulting theorem bounds the performance of model-based meta-RL under $k$-steps branched rollouts as follows. 
%-[バウンドの証明]
\begin{theorem1}\label{th:2pomdp}
Under the $k \in \mathbb N_{> 0}$ steps branched rollouts, the following inequality holds, 
%\vspace{-0.2\baselineskip}
\begin{equation}
\mathbb{E}_{a \sim \pi_\phi,h \sim p} \left[ R \right] \geq \mathbb{E}_{(a,h) \sim m_{b}(\pi_\phi, p_\theta, \mathcal{D}_{\text{env}})} \left[ R \right] - C_{\text{{\normalfont Th}}2}(\epsilon_m, \epsilon_\pi). \nonumber
\end{equation}%\vspace{-0.2\baselineskip}
\end{theorem1}
Here $\mathbb{E}_{(a,h) \sim m_{b}(\pi_\phi, p_\theta, \mathcal{D}_{\text{env}})} \left[ R \right]$ is a model return in the branched rollout scheme. 
$C_{\text{Th}2}(\epsilon_m, \epsilon_\pi)$ is the discrepancy between the returns, and its expanded representation is shown in Table~\ref{tab:discrepancies}. 
%-analysis
The value of $C_{\text{Th}2}(\epsilon_m, \epsilon_\pi)$ tends to monotonically increase as the value of $k$ increases, regardless of the values of $\epsilon_m$ and $\epsilon_\pi$. 
This implies that the optimal value of $k$ is 1.

\subsection{Comparison of the performance guarantees}\label{sec:comparison_performance}
%-[比較方法]
We conduct theoretical and empirical comparisons, focusing on the magnitude correlation of discrepancies in Theorem~\ref{th:1pomdp} and that of Theorem~\ref{th:2pomdp} (i.e., $C_{\text{{\normalfont Th}}1}$ and $C_{\text{{\normalfont Th}}2}$). 
In theoretical analysis, we compare the magnitude correlation between the terms relying on $\epsilon_m$ in $C_{\text{Th}1}$ and ones in $C_{\text{Th}2}$. 
In empirical analysis, we compare the empirical value of $C_{\text{Th}1}$ and $C_{\text{Th}2}$ by varying the values of various factors. 

\begin{figure}[t]
\begin{minipage}{0.49\hsize}\begin{center}
   \includegraphics[clip, width=0.99\hsize]{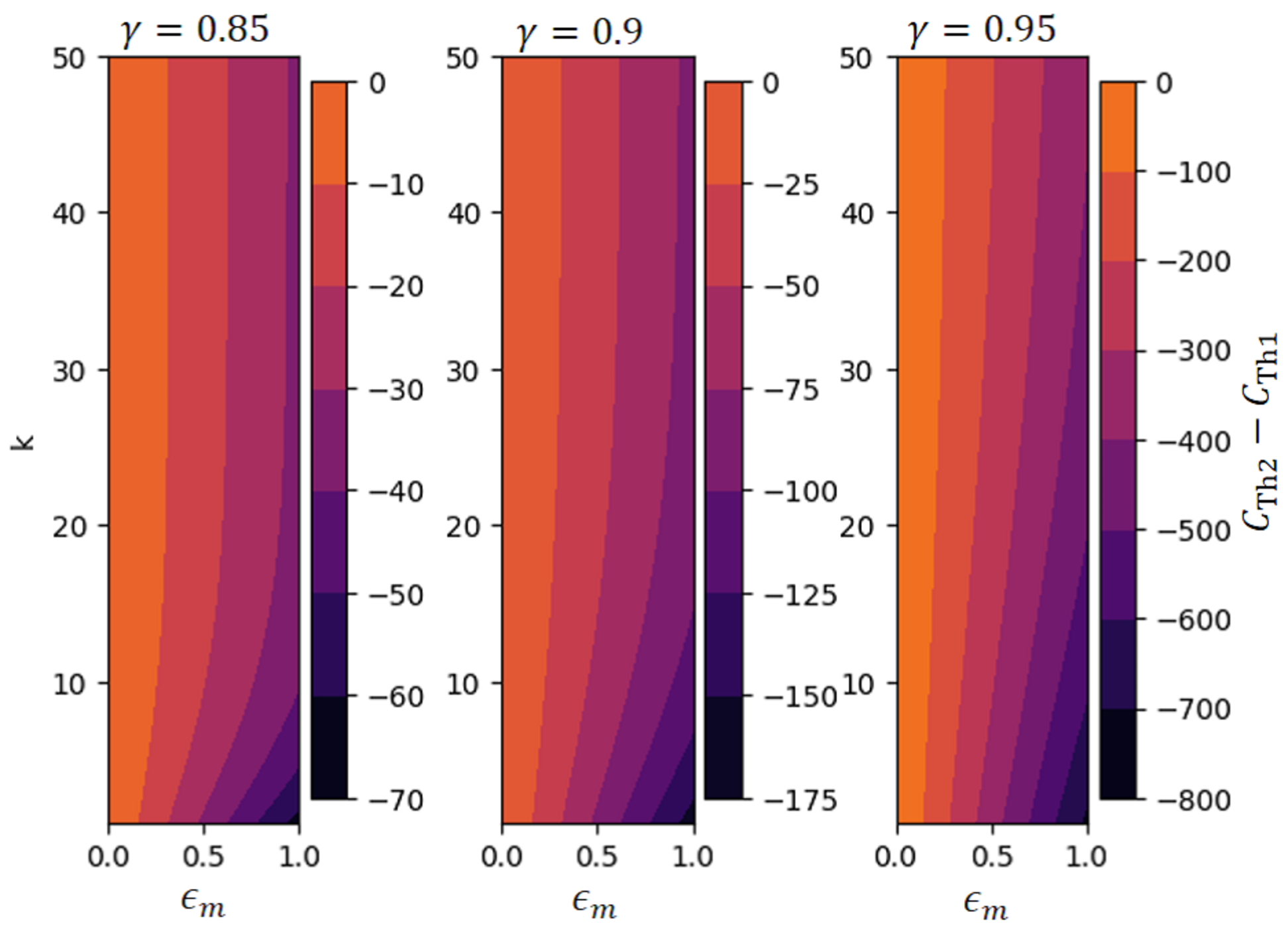}
{(a) $\epsilon_\pi = 0$}
\end{center}%\vspace{-1\baselineskip}
\end{minipage}
\begin{minipage}{0.49\hsize}\begin{center}
   \includegraphics[clip, width=0.99\hsize]{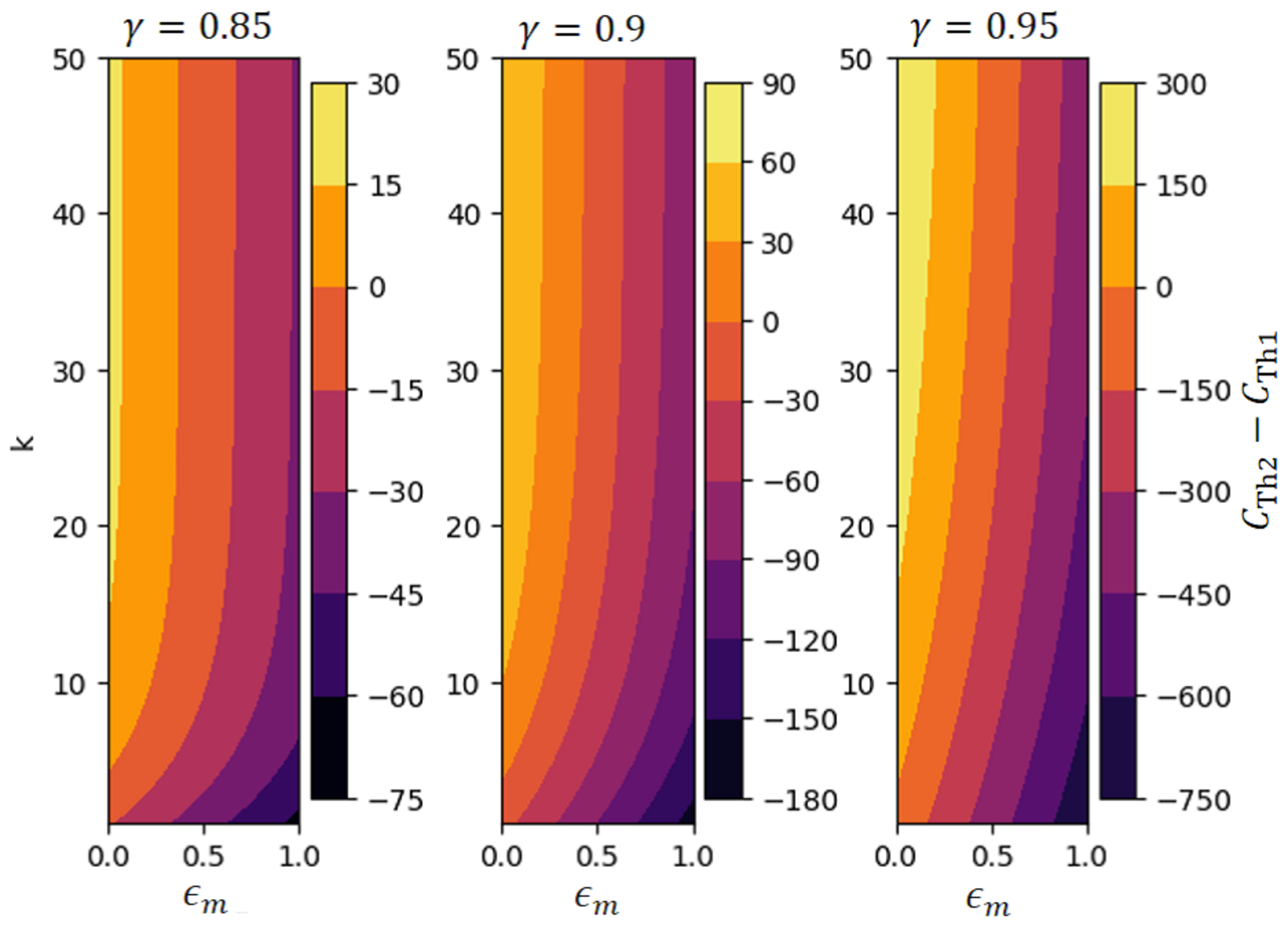}
{(b) $\epsilon_\pi = 1 - \epsilon_m$}
\end{center}%\vspace{-1\baselineskip}
\end{minipage}
 \vspace{-0.5\baselineskip}
  %\vspace{-0.8\baselineskip}
\caption{Comparison results of discrepancies in our theorems in Section~\ref{sec:monotonic}. 
We compare the discrepancy ($C_{\text{Th}1}$) in Theorem~\ref{th:1pomdp} with that ($C_{\text{Th}2}$) in Theorem~\ref{th:2pomdp} by subtracting the former from the later. 
The comparison results in cases of $\gamma=0.85$, $\gamma=0.9$, and $\gamma=0.95$ are shown in the figures. 
In each figure, the vertical axis represents the model-based rollout length $k$ (from 1 to 50), and the horizontal axis represents the model errors $\epsilon_m$ (from 0.0 to 1.0). 
In Figure (a), to focus on the model error $\epsilon_m$ effect, we set $\epsilon_\pi$ value as $\epsilon_\pi = 0$. In Figure (b), to illustrate the effect of both $\epsilon_m$ and $\epsilon_\pi$, we set $\epsilon_\pi$ value as $\epsilon_\pi = 1 - \epsilon_m$. In both figures, we set $r_{\text{max}}$ value as $r_{\text{max}} = 1$. 
%deppre For evaluating the discrepancy values, we set the values as $r_{\text{max}} = 1$, $\epsilon_\pi = 1 - \epsilon_m$. 
A key insight with the figures is that, by setting $k$ to 1, the branched rollouts reduce the discrepancy more significantly than the full model-based rollouts (i.e., $C_{\text{Th}2} - C_{\text{Th}1}$ is always smaller than zero). 
}\label{fig:desc_trend}
 %\vspace{-1\baselineskip}
   \vspace{-1.0\baselineskip}
\end{figure}

%[Branched rolloutとmodel-based rolloutの比較結果]
%-theory
A theoretical comparison result is provided as follows: 
\begin{corollary1}\label{cor:disc1anddisc2}
The terms relying on $\epsilon_m$ at $k=1$ in $C_{\text{Th}2}$ are equal to or smaller than those relying on $\epsilon_m$ in $C_{\text{Th}1}$. 
\end{corollary1}
The proof is given in Appendix~\ref{sec:disc1anddisc2}. 
This result implies that the model error is less harmful in the branched rollout with $k=1$. 
%deppre This result means that the terms relying on $\epsilon_m$ in $C_{\text{Th}2}$ with $k = 1$ are always smaller than the ones in $C_{\text{Th}1}$. 
%-empirical result
In addition, in empirical comparison results shown in Figure~\ref{fig:desc_trend}, we can see that $C_{\text{Th}2}$ with $k=1$ is always smaller than $C_{\text{Th}1}$. 
%
%deppre The comparison results show that \emph{the one-step branched rollout can reduce the discrepancy more significantly than the full model-based rollout}. 
These results motivate us to use \emph{\textit{branched rollouts with $k=1$ for the meta-RL setting}}. 
%
%-[summary of comparison result]

\section{A practical model-based meta-RL method with the branched rollout}\label{sec:practicalMBPO}

%-[導入](最初の導入と同じ)
In the previous section, we demonstrated the usefulness of the branched rollout. 
In this section, we propose Meta-Model-Based Meta-Policy Optimization\footnote{The ``meta-model'' and ``meta-policy'' come from the use of $p_{\theta}(r_{t}, o_{t+1} | a_t, h_{L:t})$ and $\pi_{\phi}(a_{t} | h_{L:t})$ in this method. Following previous work~\citep{clavera2018model}, we refer to this type of history-conditioned policy as a \textbf{meta-policy}. Similarly, we refer to this type of history-conditioned model as a \textbf{meta-model}.} (M3PO), which is a model-based meta-RL method with the branched rollout. 
\begin{algorithm}[t]
\caption{Meta-Model-Based Meta-Policy Optimization (M3PO)}
\label{alg1:Meta-MBPO}
\begin{algorithmic}[1]
\STATE Initialize policy $\pi_{\phi}$, model $p_{\theta}$, environment dataset $\mathcal{D}_{\text{env}}$, model dataset $\mathcal{D}_{\text{model}}$ 
\FOR{$N$ epochs}
 \STATE Train $p_\theta$ with $\mathcal{D}_{\text{env}}$: $\theta \leftarrow \argmax_\theta \mathbb{E}_{r_t, o_{t+1}, a_t, h_{L:t} \sim \mathcal{D}_{\text{env}}} \left[ p_{\theta}(r_{t}, o_{t+1} | a_t, h_{L:t}) \right]$ 
 \FOR{$E$ steps}
  \STATE Collect trajectories from environment in accordance with $\pi_{\phi}$: $\mathcal{D}_{\text{env}} = \mathcal{D}_{\text{env}} \cup \left\{ \left( h_{L:t}, a_t, o_{t+1}, r_{t}\right) \right\}$
  
  \FOR{$M$ model-based rollouts}
  \STATE Sample $h_{L:t}$ uniformly from $\mathcal{D}_{\text{env}}$
  \STATE Perform $k$-step model-based rollouts starting from $h_{L:t}$ using $\pi_\phi$, and then add fictitious trajectories to $\mathcal{D}_{\text{model}}$
  \ENDFOR
  \FOR{$G$ gradient updates}
  \STATE Update policy parameters with $\mathcal{D}_{\text{model}}$: $\phi \leftarrow \phi - \nabla_{\phi} J_{\mathcal{D}_{\text{model}}}(\phi)$
  \ENDFOR
 \ENDFOR
\ENDFOR
\end{algorithmic}
\end{algorithm}

%-[M3POの概要]
M3PO is described in Algorithm~\ref{alg1:Meta-MBPO}. 
%-[モデルの学習方法] 
In line 3, the model is learned via maximum likelihood estimation. 
In line 5, trajectories are collected from the environment with policy $\pi_\phi$ and stored into the environment dataset $\mathcal{D}_{\text{env}}$. 
%--[branched rolloutの導入]
In lines 6--9, $k$-step branched rollouts are run to generate fictitious trajectories, and the generated ones are stored into the model dataset $\mathcal{D}_{\text{model}}$. 
%--[方策の学習方法]
In lines 10--12, the policy $\pi_\phi$ is learned to improve the model return. 
%-- 
%deppre Below, we explain our implementation of the model and policy. 

%-[実装の詳細MBPOとの比較をしながら説明]
%我々の手法は主にモデルと方策が~で条件つけられている点が異なる(異なる点はハイライト)。
%-[モデルの表現方法]
%-[Predictive modesl]
The model is implemented as a bootstrap ensemble of $B$ diagonal Gaussian distributions:  $p_{\theta}(r_{t}, o_{t+1} | a_t, h_{L:t}) = \frac{1}{B} \sum_{i=1}^{B} \mathcal{N}\left(r_{t+1}, o_{t+1} | \mu_{\theta}^{i}\left(  a_t, h_{L:t} \right), \sigma_{\theta}^{i}\left(  a_t, h_{L:t} \right) \right)$. 
Here, $\mu_\theta^i$ and $\sigma_\theta^i$ are the mean and standard deviation, respectively. 
The model ensemble technique is used to consider both epistemic and aleatoric uncertainty of the model into prediction~\citep{NEURIPS2018_3de568f8}. 
%-[]
$\mu_\theta^i$ and $\sigma_\theta^i$ are implemented with a recurrent neural network (RNN) and a feed-forward neural network (FFNN). 
At each prediction, $\{o_{\max(t-L,0)}, a_{\max(t-L,0)}, ..., o_{t-1}, a_{t-1} \}$ in $h_{L:t}$ is fed to the RNN. Then its hidden-unit outputs $z_t$ are fed to the FFNN together with $o_t$ and $a_t$. 
The FFNN outputs the mean and standard deviation of the Gaussian distribution, and the next observation and reward are sampled from its ensemble. 
\begin{figure}[t]
\begin{center}
\includegraphics[clip, width=0.5\hsize]{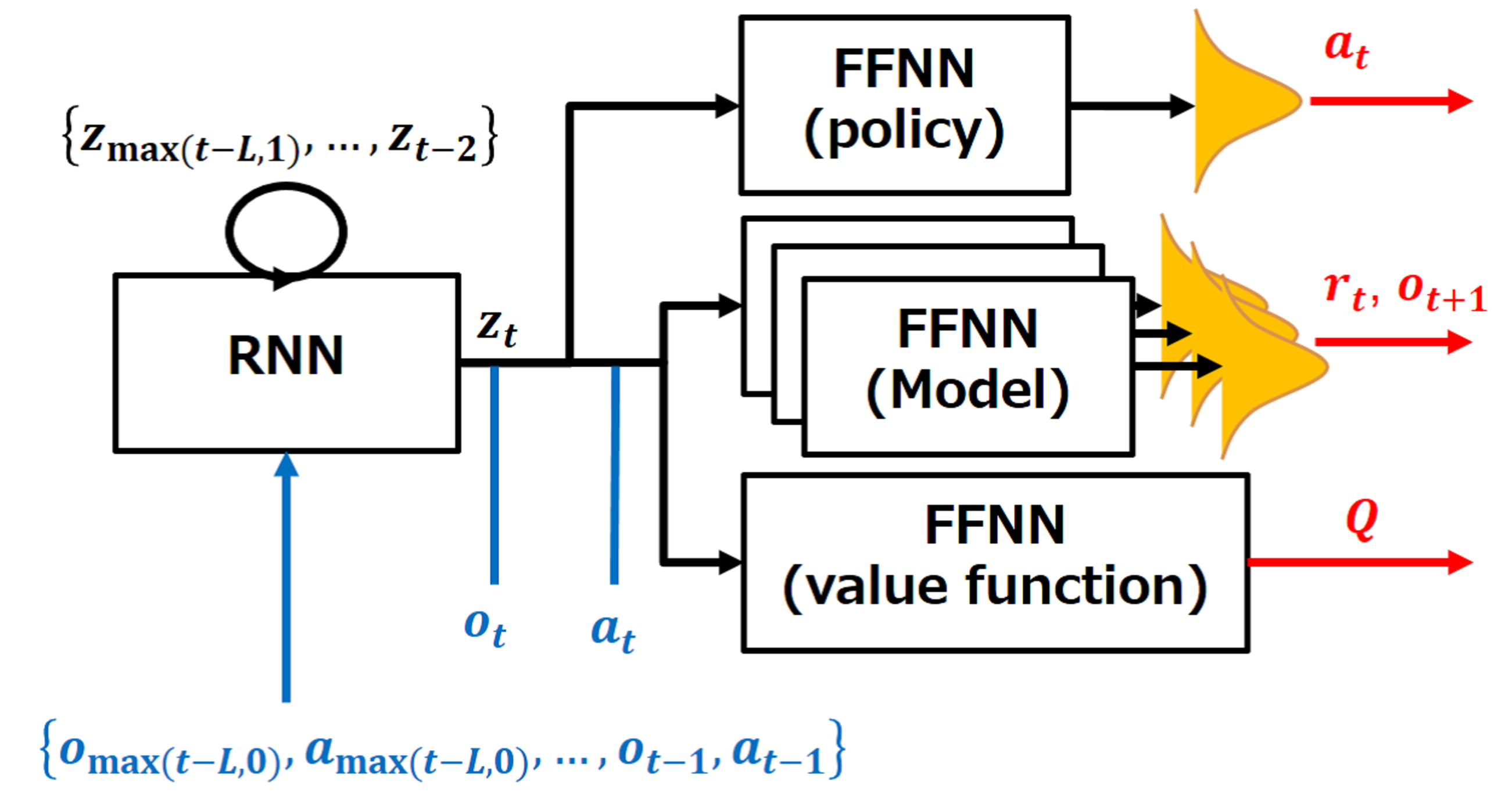}
\end{center}
 %\vspace{-1.5\baselineskip}
   \vspace{-1.8\baselineskip}
\caption{A summary of our implementation of the model, policy, and value network for M3PO. We use a recurrent neural network (RNN) to encode the history. RNN's hidden-unit output $z$ is fed into the model, policy and value network. 
We use a feed-forward neural network (FFNN) for the model, policy and value network. 
Especially, we use an ensemble of the networks with $B$ network heads for the model. 
} 
\label{fig:networkimplementation}
 \vspace{-1.0\baselineskip}
\end{figure}

%-[方策の表現方法]
%-[Policy optimization]
The policy is implemented as $\pi_\phi(a_{t} | h_{L:t}) = \mathcal{N}\left(a_{t} | \mu_{\phi}\left( h_{L:t} \right), \sigma_{\phi}\left( h_{L:t} \right) \right)$. 
Here, $\mu_\phi$ and $\sigma_\phi$ are the mean and standard deviation, respectively. 
The network architecture for them is the same as that for the model, except that the former does not contain $a_t$ as input. 
%
%-value functions[]
To train $\phi$, we use a soft-value policy improvement objective~\citep{haarnoja2018soft}:  $J_{\mathcal{D}_{\text{model}}}(\phi)= \mathbb{E}_{ h_{L:t} \sim \mathcal{D}_{\text{model}}}\left[ D_{KL}\left( \pi_\phi || \exp\left( Q - V \right) \right) \right]$. 
Here, $D_{KL}$ is the Kullback-Leibler divergence, and $Q$ and $V$ are soft-value functions: 
$Q(a_{t}, h_{L:t}) = \mathbb{E}_{(r_t,h_{L:t+1}) \sim \mathcal{D}_{\text{model}}} \left[ r_t + \gamma V (h_{L:t+1}) | a_{t}, h_{L:t} \right]$ and $V(h_{L:t}) = \mathbb{E}_{a \sim \pi_{\phi}}\left[ Q(a, h_{L:t}) - \log \pi_\phi(a | h_{L:t}) | h_{L:t} \right]$. 
The network architecture for $Q$ is identical to that for the model. 
We do not prepare the network for $V$, and it is directly calculated by using $Q$ and $\pi_\phi$. 

%-[conclusion]
Figure~\ref{fig:networkimplementation} gives a summary of our implementation of the model, policy and value network for M3PO. 
%deoore The model and policy (and value function) are implemented as shown in Figure~\ref{fig:networkimplementation}. 

\section{Experiments}\label{sec:experiments}
%-[概要]
In this section, we report our experiments~\footnote{Source code to replicate the experiments is available at \url{https://github.com/TakuyaHiraoka/Meta-Model-Based-Meta-Policy-Optimization}}. %deppre The source code to replicate the experiments will be open to the public. 

\subsection{Comparison against meta-RL baselines}\label{sec:compari}
%-[methods]
In our first experiments, we compare our method (M3PO) with two baseline methods: \textbf{probabilistic embeddings for actor-critic reinforcemnet learning (PEARL)}~\citep{rakelly2019efficient} and \textbf{learning to adapt (L2A)}~\citep{nagabandilearning}. 
More detailed information on the baselines is described in  Appendix~\ref{sec:baselines}. 
%-[evaluation criteria]
As with \citet{nagabandilearning}, our primary interest lies in improving the performance of meta-RL methods in short-term training. 
Thus, we compare the aforementioned methods on the basis of their performance in short-term training (within 200k training samples). 
Nevertheless, for complementary analysis, we compare the performances of PEARL and M3PO in long-term training (with 2.5m to 5m training samples, or until earlier learning convergence). 
Their long-term performances are denoted by \textbf{M3PO-long} and \textbf{PEARL-long}. 

\begin{figure}[t]
\begin{center}
\begin{tabular}{c}
\begin{minipage}{0.7\hsize}
      \includegraphics[clip, width=0.325\hsize]{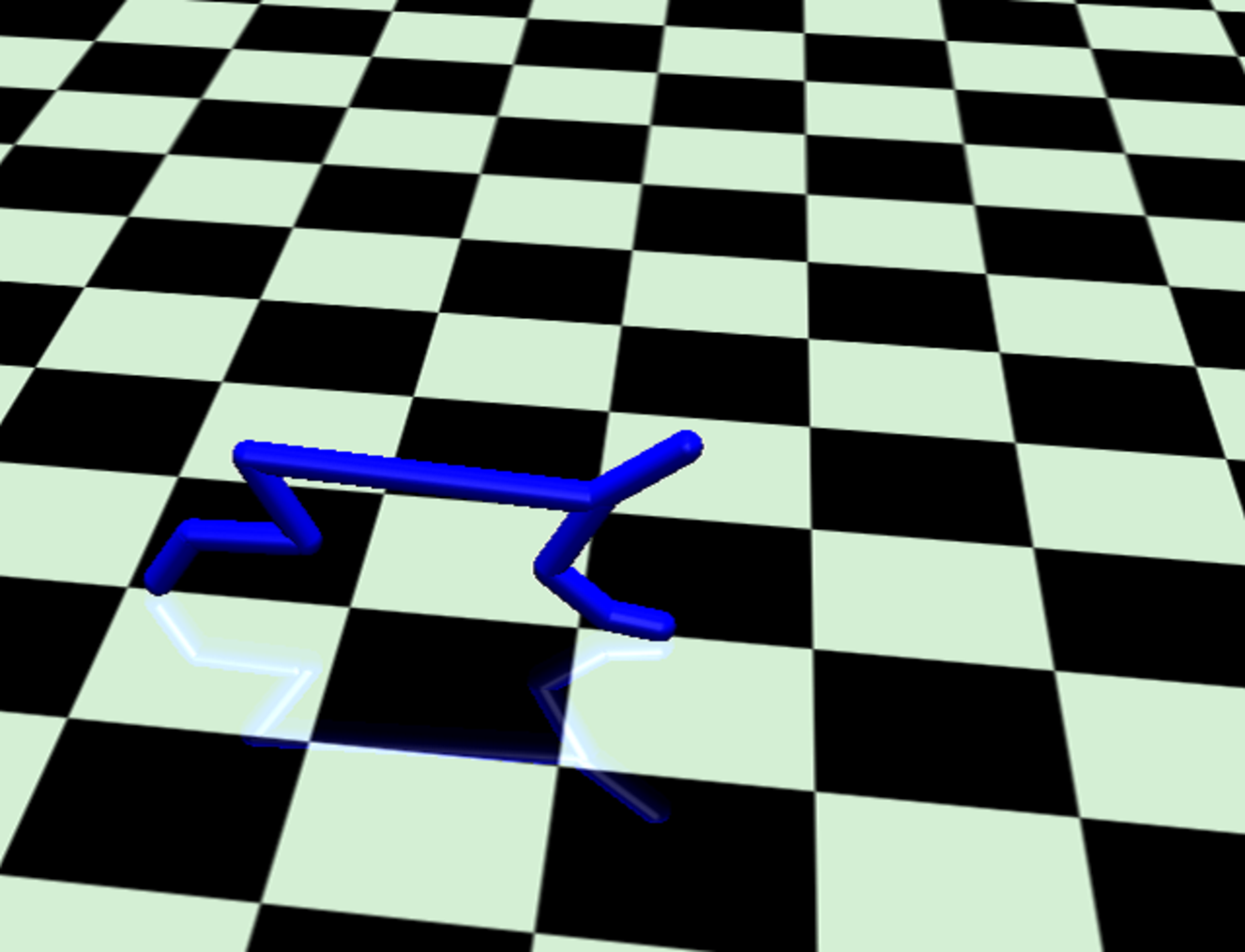}
      \includegraphics[clip, width=0.325\hsize]{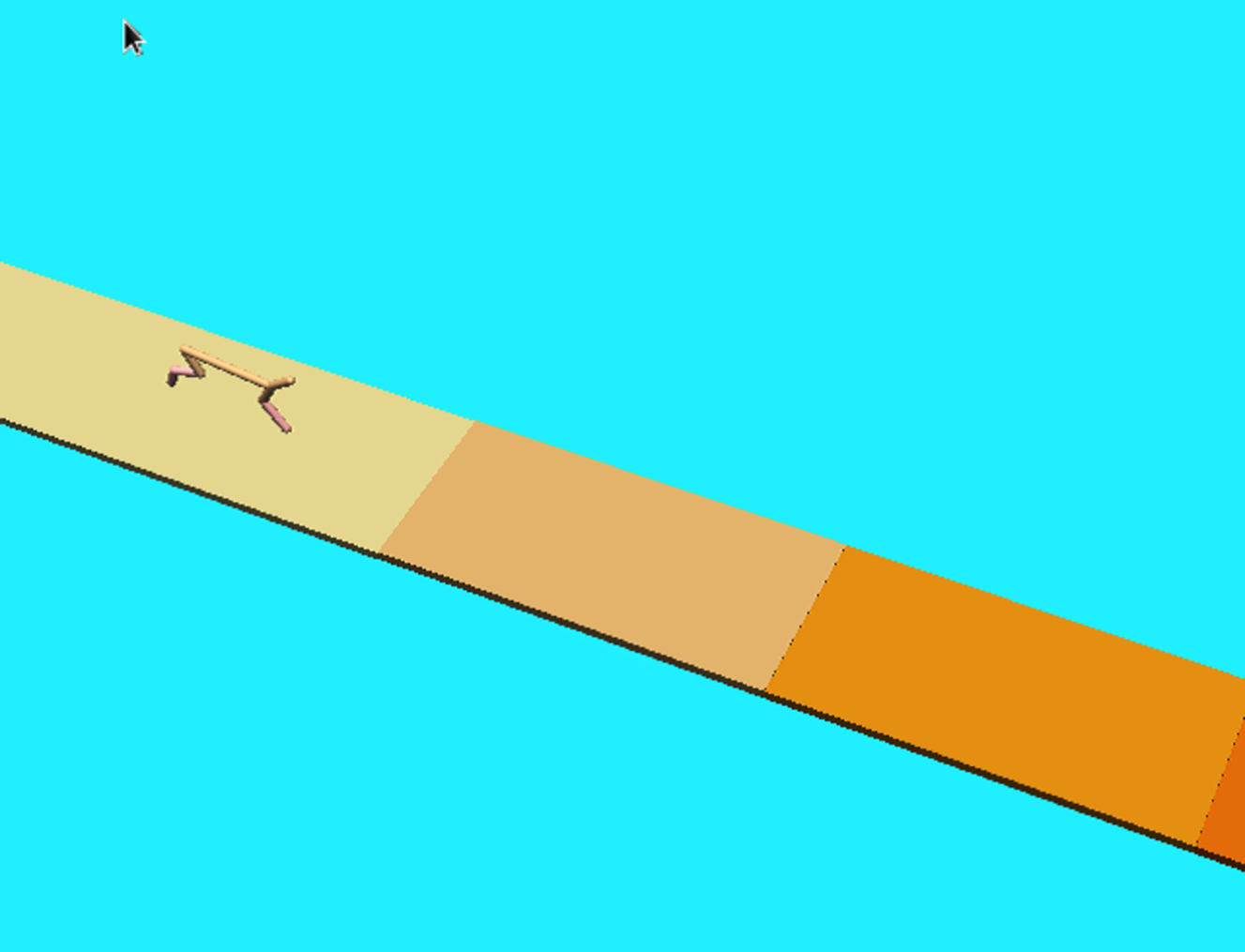}
      \includegraphics[clip, width=0.32\hsize]{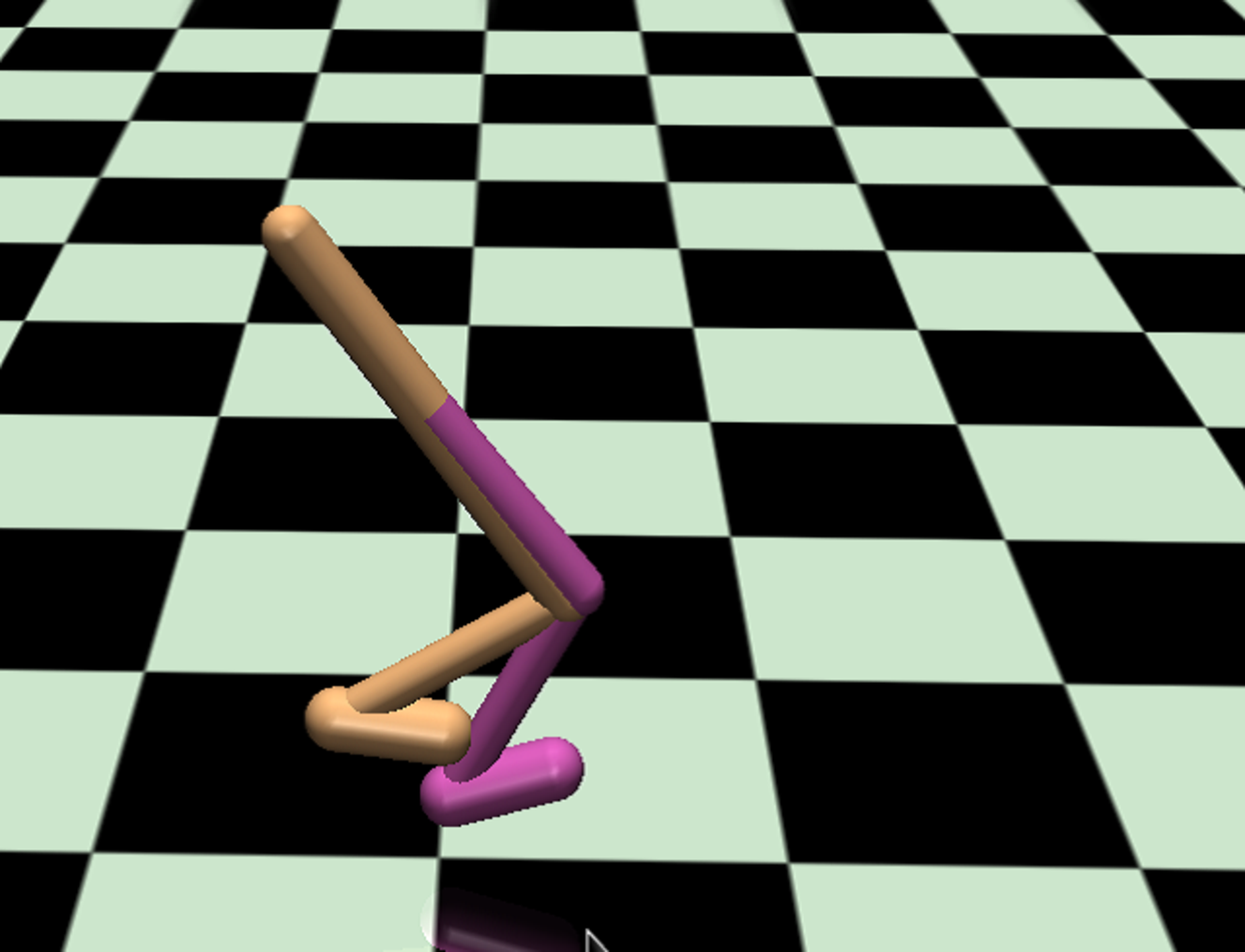}\end{minipage}\\\arrayrulecolor{white}\hline\arrayrulecolor{white}\hline\arrayrulecolor{white}\hline\arrayrulecolor{white}\hline\arrayrulecolor{white}\hline
\begin{minipage}{0.7\hsize}
      \includegraphics[clip, width=0.325\hsize]{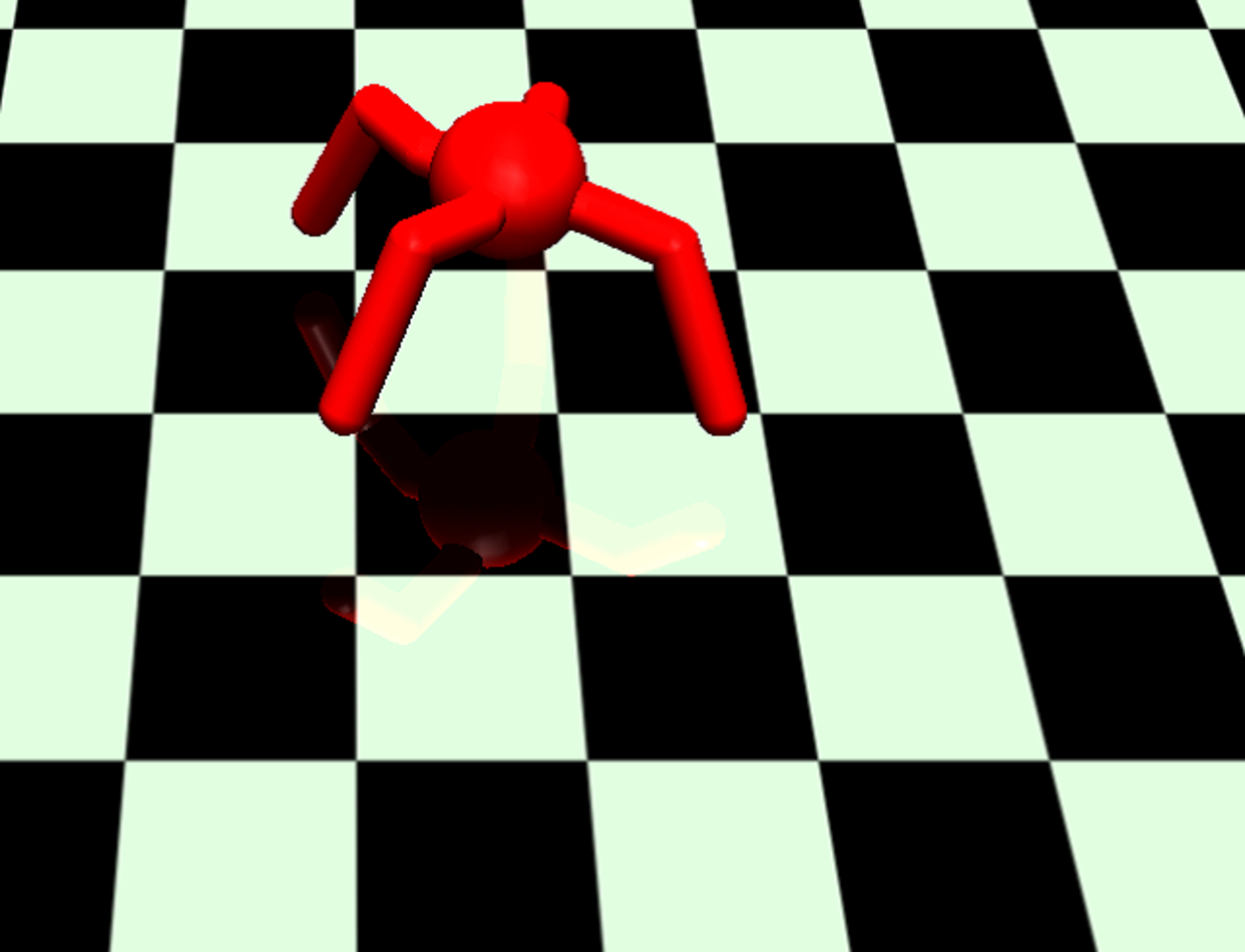}
      \includegraphics[clip, width=0.325\hsize]{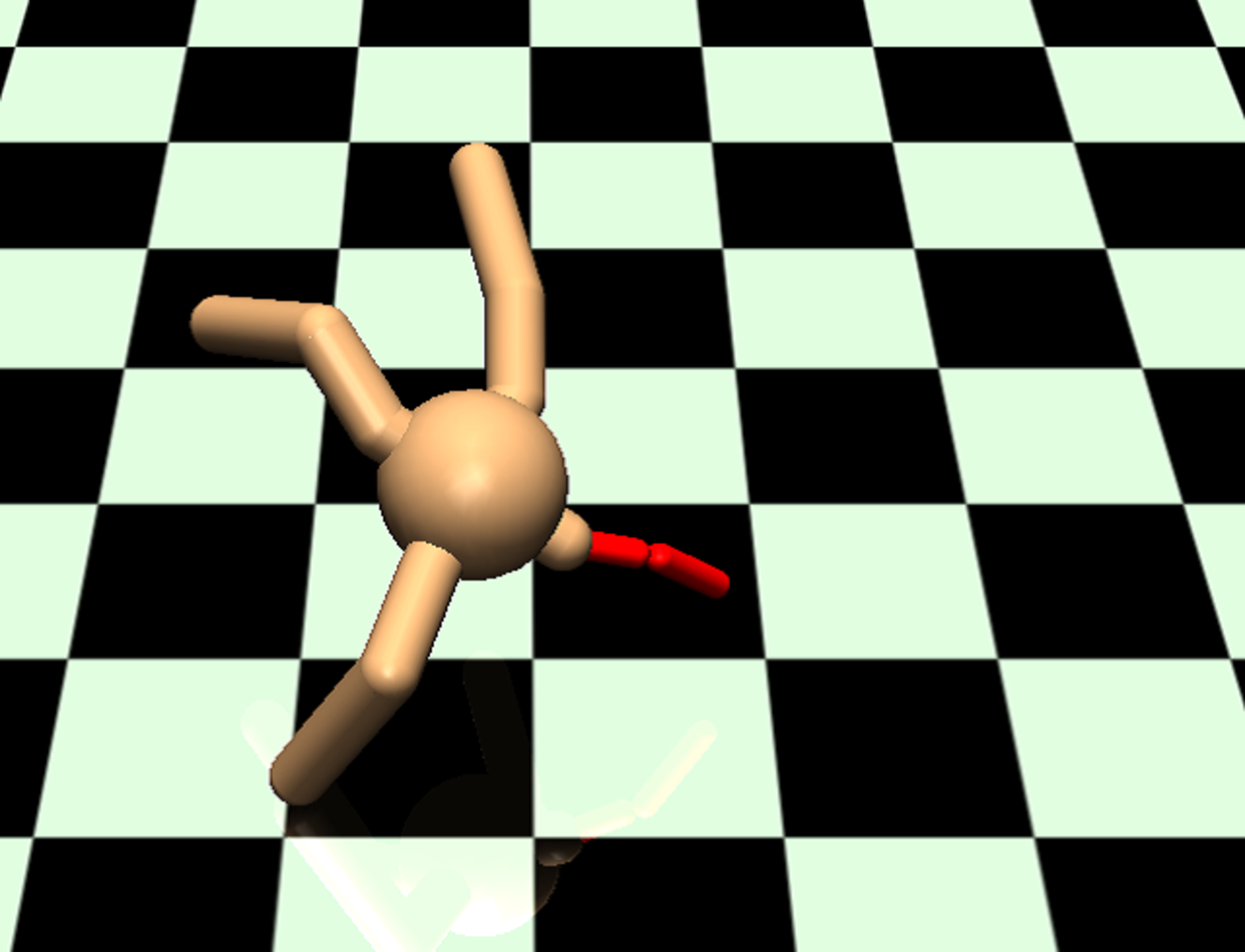}
      \includegraphics[clip, width=0.325\hsize]{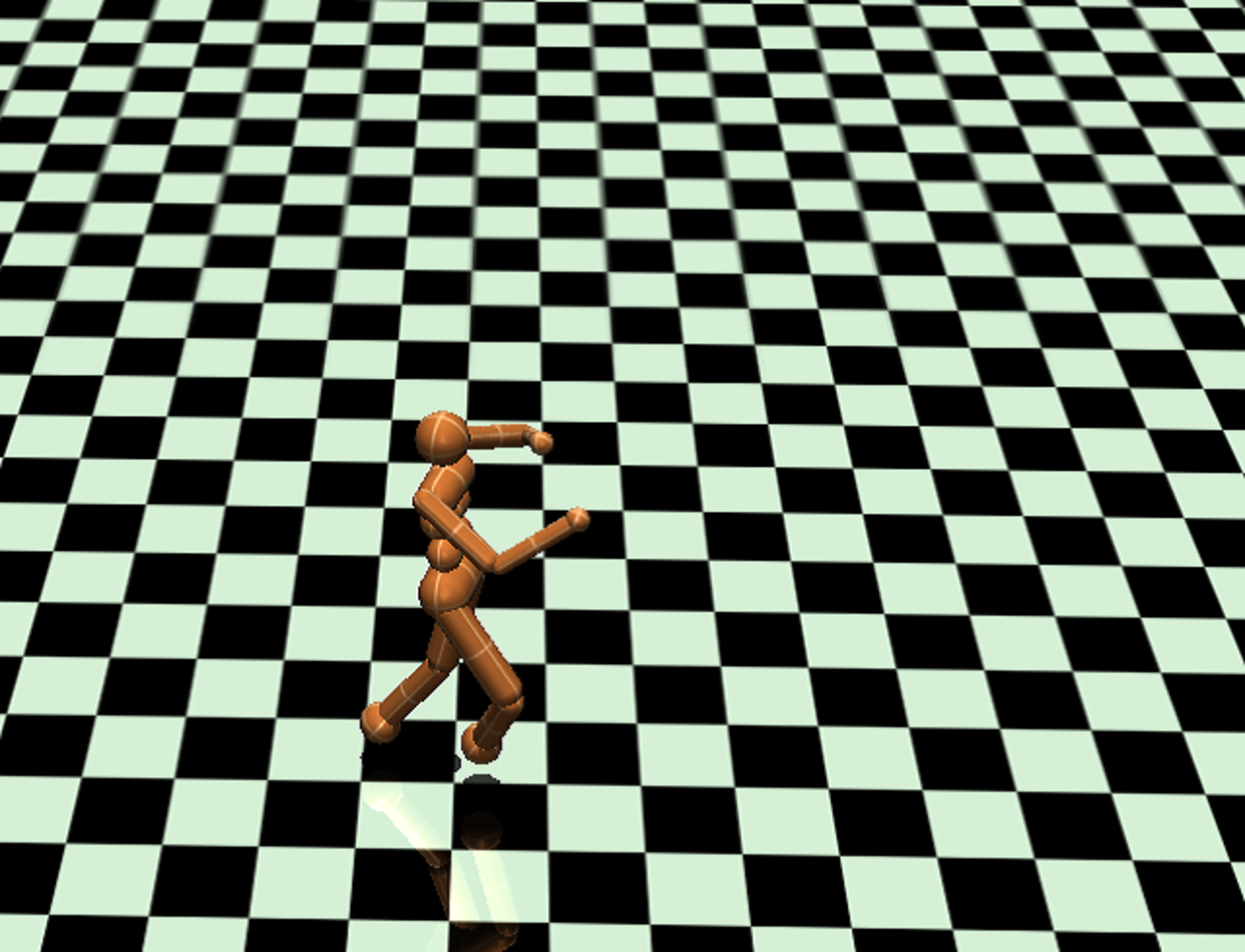}
\end{minipage}
\end{tabular}
\end{center}
 %\vspace{-1.2\baselineskip}
  \vspace{-1.5\baselineskip}
\caption{Environments for our experiment. Halfcheetah-fwd-bwd (upper left), Halfcheetah-pier (upper center), Walker2D-randomparams (upper right), Ant-fwd-bwd (lower left), Ant-crippled-leg (lower center) and Humanoid-direc (lower right). }
\label{fig:envs}
 \vspace{-1.0\baselineskip}
\end{figure}

%[実験タスク]
We compare the methods in simulated robot environments based on the MuJoCo physics engine~\citep{todorov2012mujoco}. 
For this, we consider environments proposed in the literature of meta-RL~\citep{finn2017meta,nagabandilearning,rakelly2019efficient,rothfuss2018promp}: \textbf{Halfcheetah-fwd-bwd}, \textbf{Halfcheetah-pier}, \textbf{Ant-fwd-bwd}, \textbf{Ant-crippled-leg}, \textbf{Walker2D-randomparams} and \textbf{Humanoid-direc} (Figure~\ref{fig:envs}). 
In these environments, the agent is required to adapt to a fluidly changing task that the agent cannot directly observe. 
Detailed information about each environment is described in Appendix~\ref{sec:environment_desc}. 
%--hypara
In addition, the hyperparameter settings of M3PO for each environment are shown in Appendix~\ref{sec:hypara}. 

\begin{figure*}[t]
\begin{center}
\begin{tabular}{c}
\begin{minipage}{1.0\hsize}
\begin{center}
\includegraphics[clip, width=0.6\hsize]{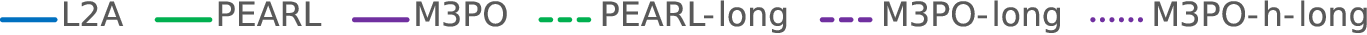}
\end{center}
\end{minipage}\\\hline
\begin{minipage}{1.0\hsize}
      \includegraphics[clip, width=0.333\hsize]{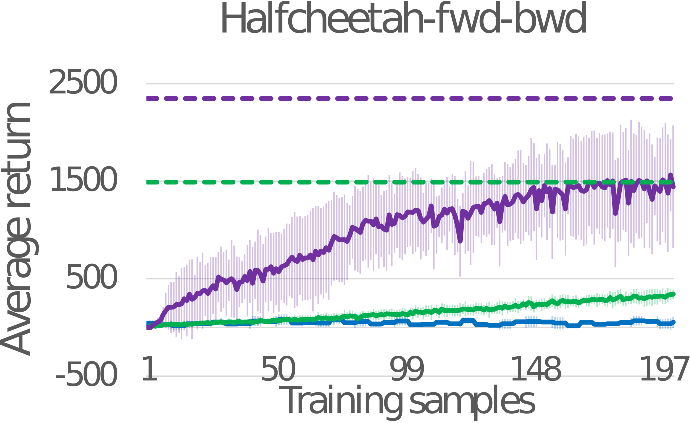}\vline
      \includegraphics[clip, width=0.333\hsize]{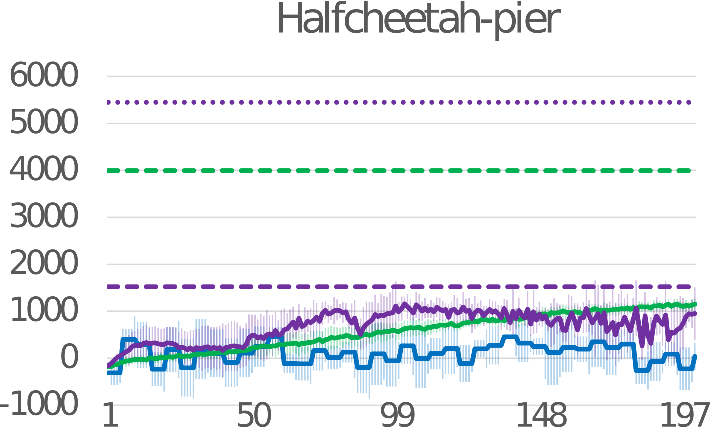}\vline
      \includegraphics[clip, width=0.333\hsize]{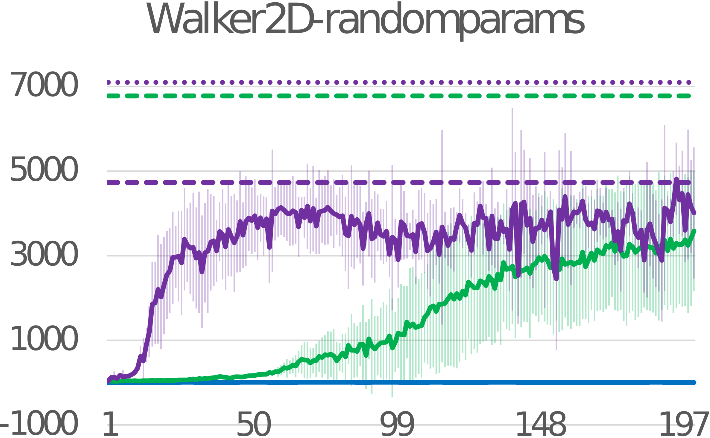}
\end{minipage}\\\hline
\begin{minipage}{1.0\hsize}
      \includegraphics[clip, width=0.333\hsize]{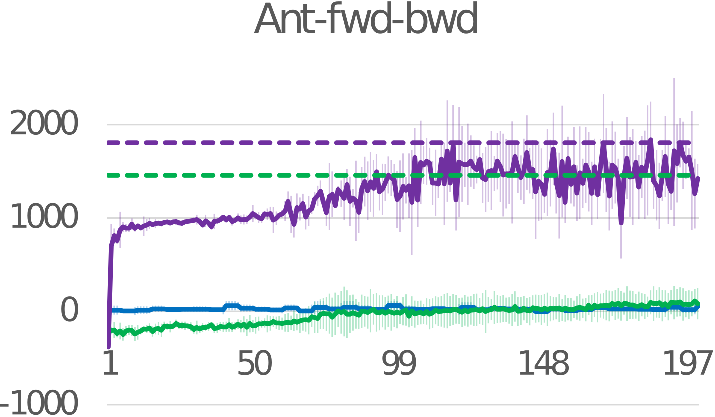}\vline
      \includegraphics[clip, width=0.333\hsize]{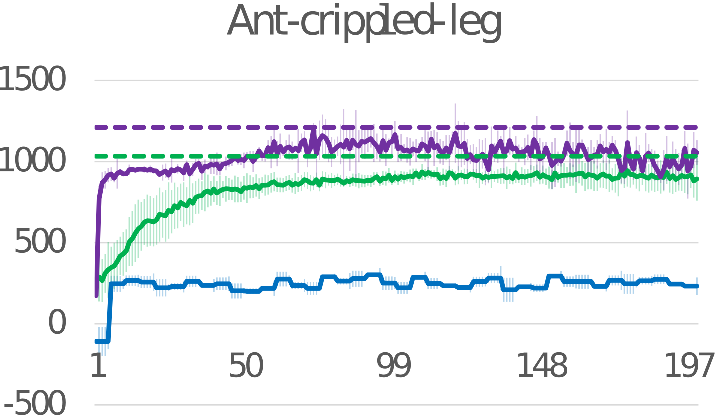}\vline
      \includegraphics[clip, width=0.333\hsize]{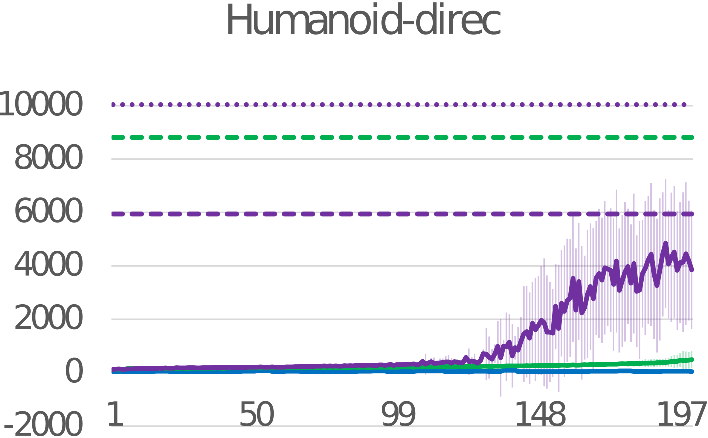}
\end{minipage}
\end{tabular}
\end{center}
%\vspace{-0.9\baselineskip}
  \vspace{-1.8\baselineskip}
\caption{Learning curves. 
In each figure, the vertical axis represents returns, and the horizontal axis represents numbers of training samples (one on the scale is equal to 1000 samples). 
The policy and model are fixed and evaluated in terms of their average return on 50 episodes at every 5000 training samples for L2A and 1000 training samples for the other methods. 
Each method is evaluated in six trials, and average returns on the 50 episodes are further averaged over the trials. 
The averaged returns and their standard deviations are plotted. 
}
\label{fig:lc}
%\vspace{-0.5\baselineskip}
\end{figure*}

%-[Q1]
%deppre Regarding Q1, 
Our experimental results demonstrate that M3PO outperforms existing meta-RL methods. 
%- [L2AとPEARLとの比較]
%--[図２の説明]
Figure~\ref{fig:lc} shows learning curves of M3PO and existing meta-RL methods (L2A and PEARL). 
These learning curves indicate that M3PO has better sample efficiency than L2A and PEARL. 
%---[M3PO vs L2A]
The performance (average return) of L2A remains poor and does not improve even when the number of training samples increases. 
%---[M3PO vs PEARL]
On the other hand, the performance of PEARL improves as the number of training samples increases, but the degree of improvement is smaller. 
%--
Note that, in an early stage of the training phase, unseen tasks appear in many test episodes. 
Therefore, the improvement of M3PO over L2A and PEARL at the early stage of training indicates M3PO's high adaptation capability for unseen tasks. 

%--[Eval. of M3PO-h]
Interestingly, in Halfcheetah-pier, Walker2D-randomparams, and Humanoid-direc, M3PO (M3PO-long) has worse long-term performance than PEARL (PEARL-long). 
That is, the improvement of M3PO over PEARL, which is a model-free approach and does not depend on the model, becomes smaller as the learning epoch elapses. 
 This is perhaps due to the model bias. 
%deppre This indicates that gradually transitioning from M3PO to PEARL (or other model-free approaches) needs to be considered to improve overall performance.
This result indicates that gradually making M3PO less dependent on the model needs to be considered to improve overall performance. 
Motivated by this observation, we equip M3PO with a gradual transition mechanism. 
Specifically, we replace $\mathcal{D}_{\text{model}}$ in line 11 in Algorithm~\ref{alg1:Meta-MBPO} with the mixture of $\mathcal{D}_{\text{model}}$ and $\mathcal{D}_{\text{env}}$. 
During training, the mixture ratio of $\mathcal{D}_{\text{model}}$ is gradually reduced. 
As this ratio reduces, the M3PO becomes less dependent on the model. %deppre and is transitioned to a model-free approach. 
More details of our modification for M3PO are described in Appendix~\ref{sec:m3po-h}. 
The long-term performance of the modified M3PO is shown as \textbf{M3PO-h-long} in Figure~\ref{fig:lc}. 
We can see that it is the same as or even better than that of PEARL-long. 

%-[Behavior analysis]
Figure~\ref{fig:example} shows an example of policies learned by M3PO with 200k training samples in Humanoid-direc and indicates that the learned policy successfully adapts to tasks. 
Additional examples of the policies learned by PEARL and M3PO are shown in the video at the following link: \url{https://drive.google.com/file/d/1DRA-pmIWnHGNv5G_gFrml8YzKCtMcGnu/view?usp=sharing} 
\begin{figure*}[t]
%\vspace{-10pt}
%\vspace{-1.0\baselineskip}
  \begin{center}
\includegraphics[clip, width=0.13\hsize]{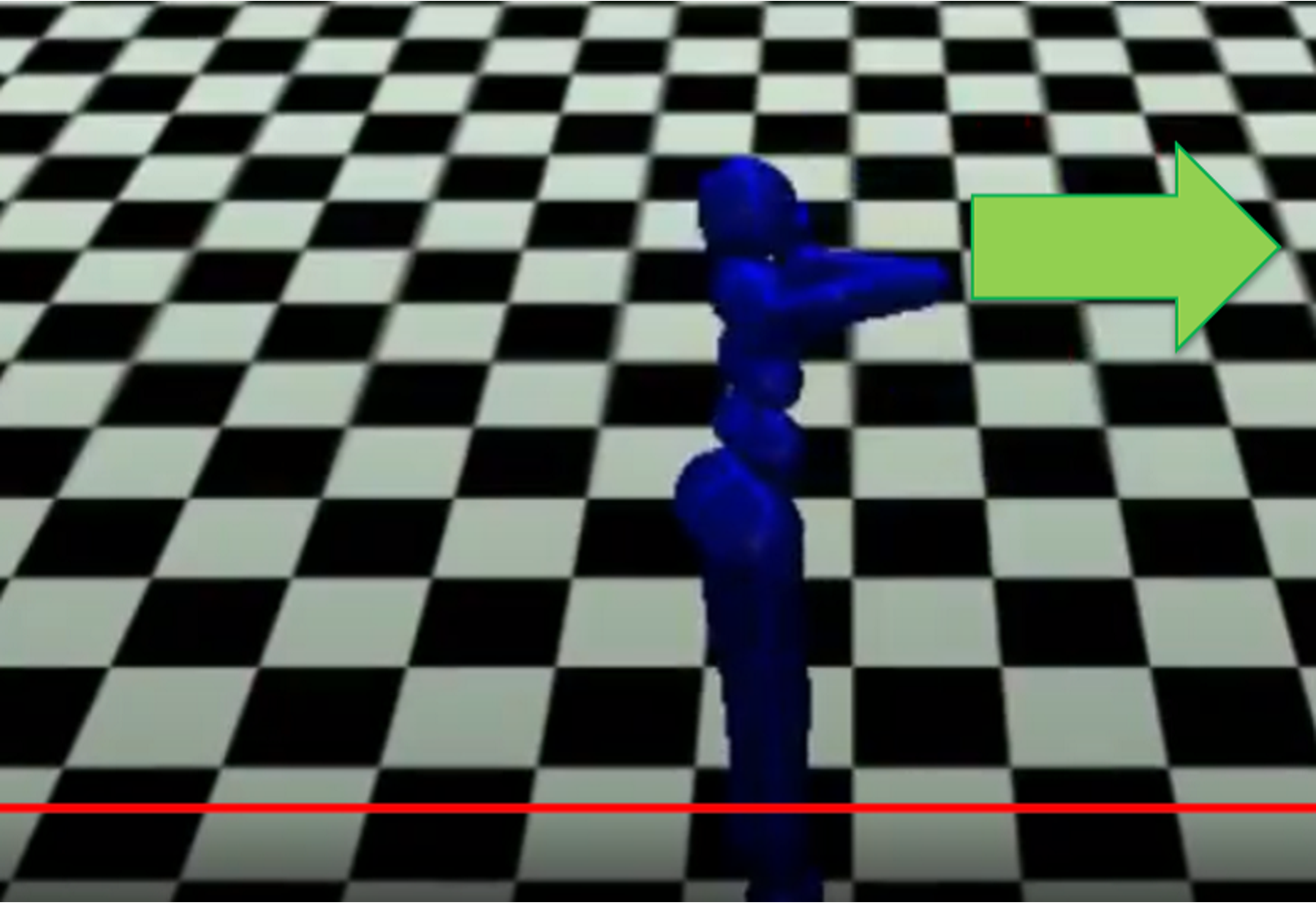}
\includegraphics[clip, width=0.13\hsize]{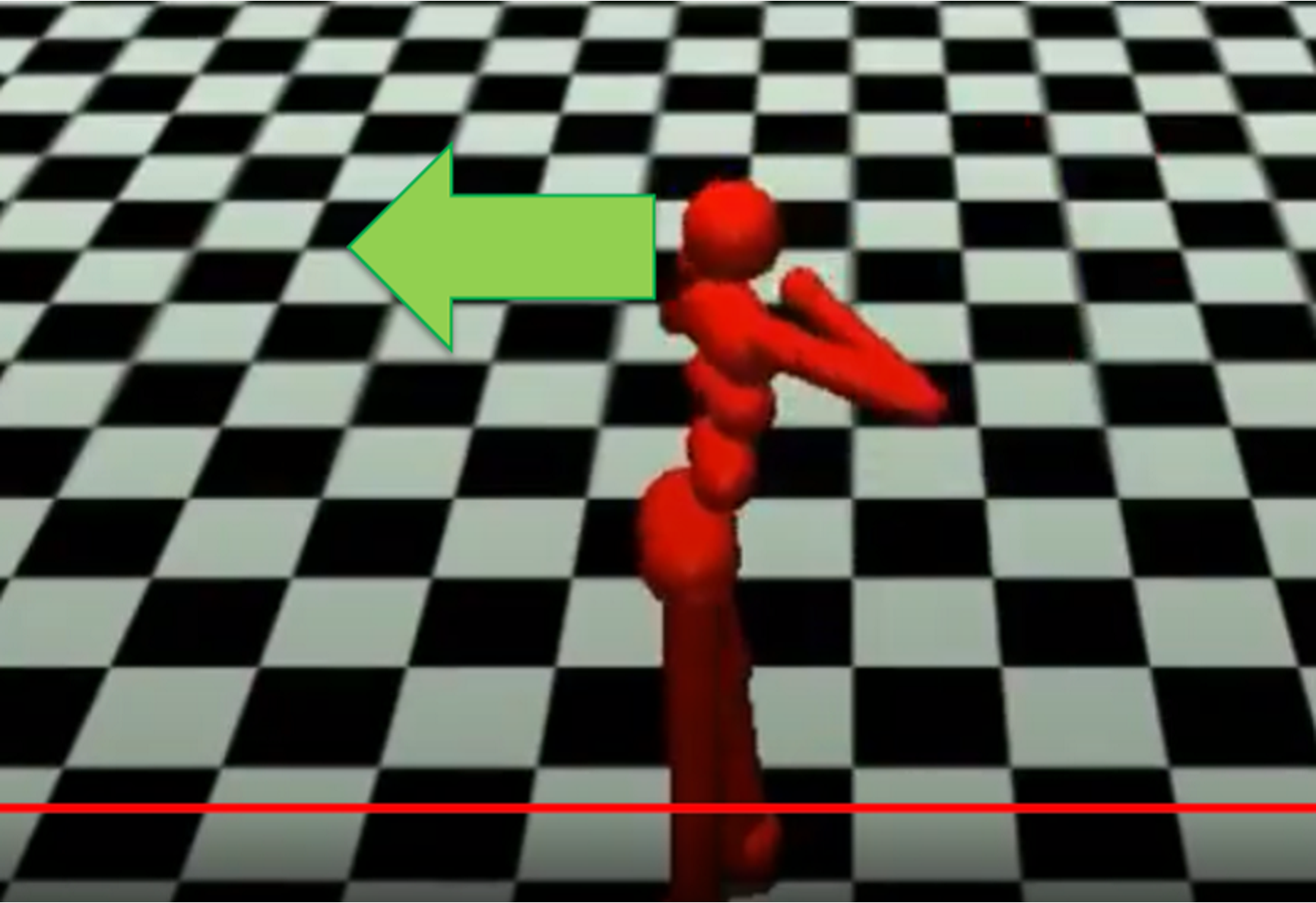}
\includegraphics[clip, width=0.13\hsize]{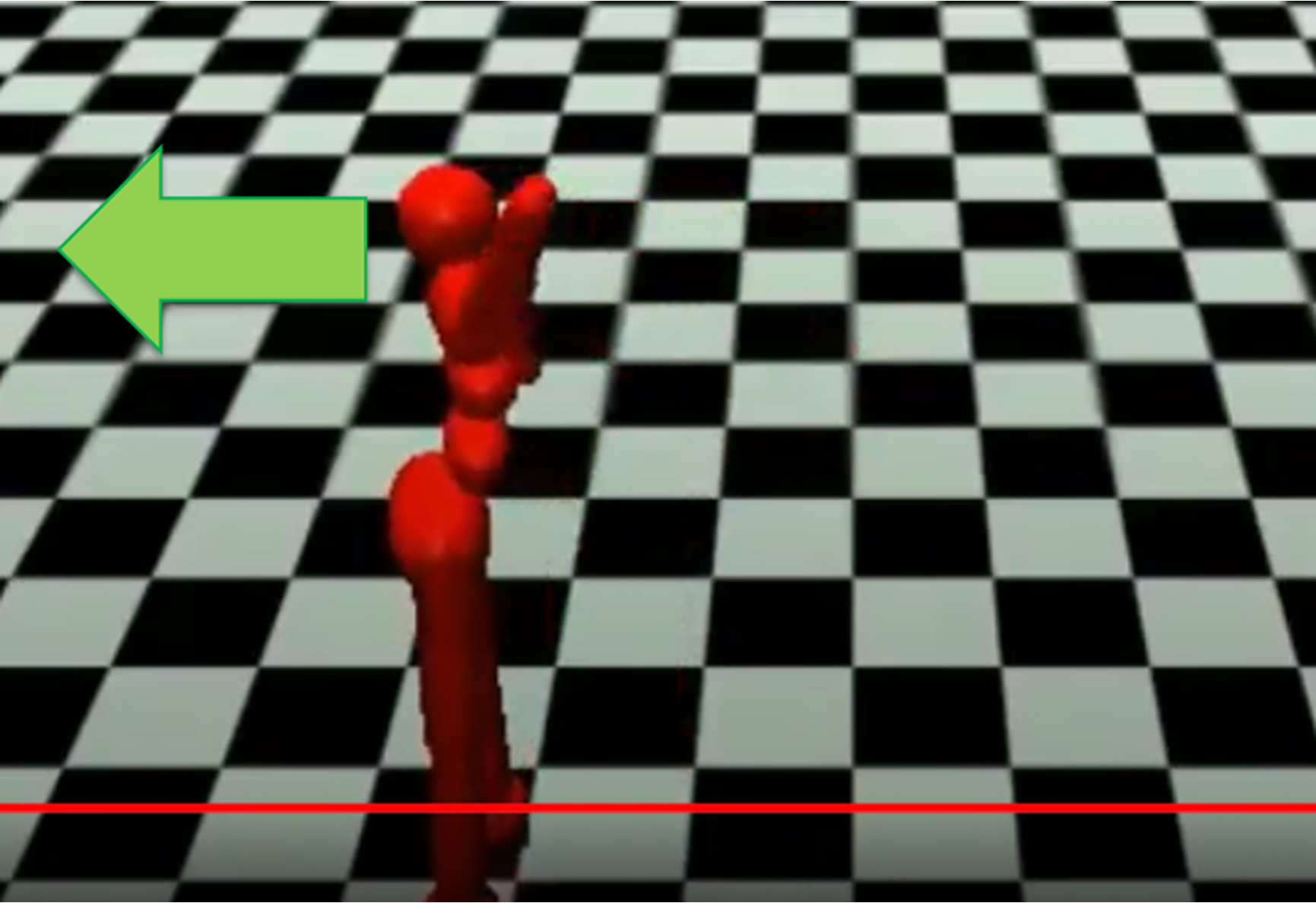}
\includegraphics[clip, width=0.13\hsize]{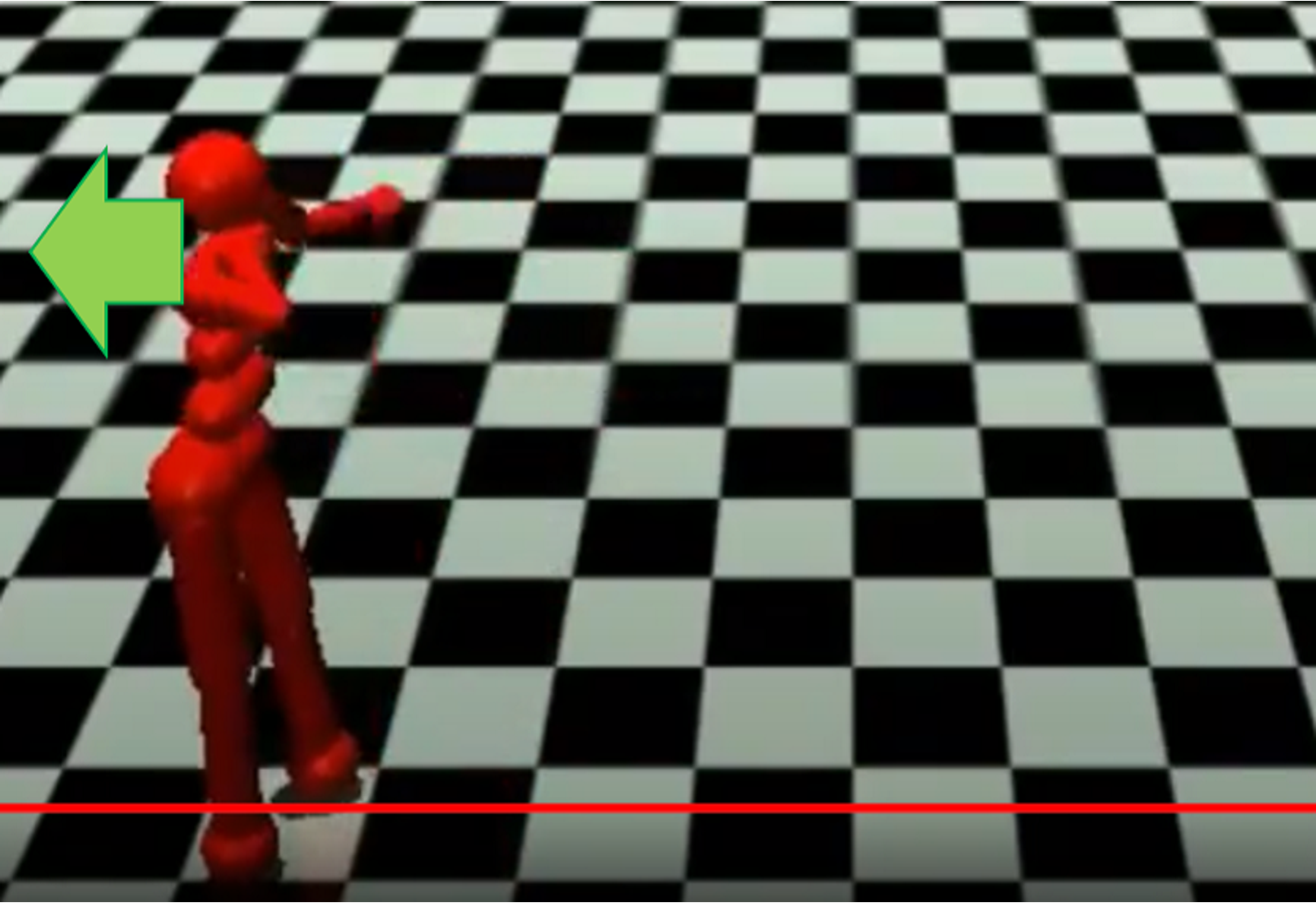}
\includegraphics[clip, width=0.13\hsize]{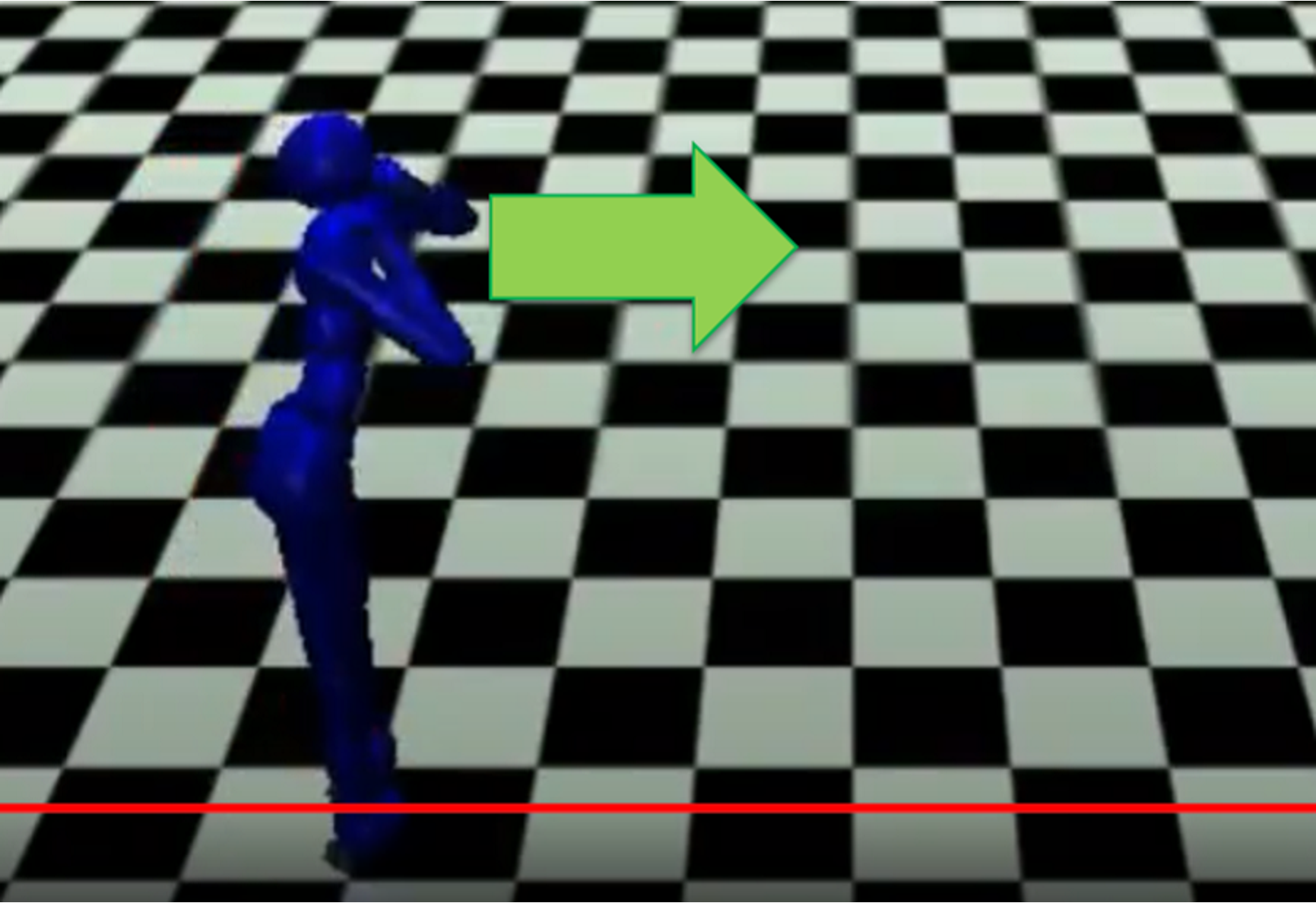}
\includegraphics[clip, width=0.13\hsize]{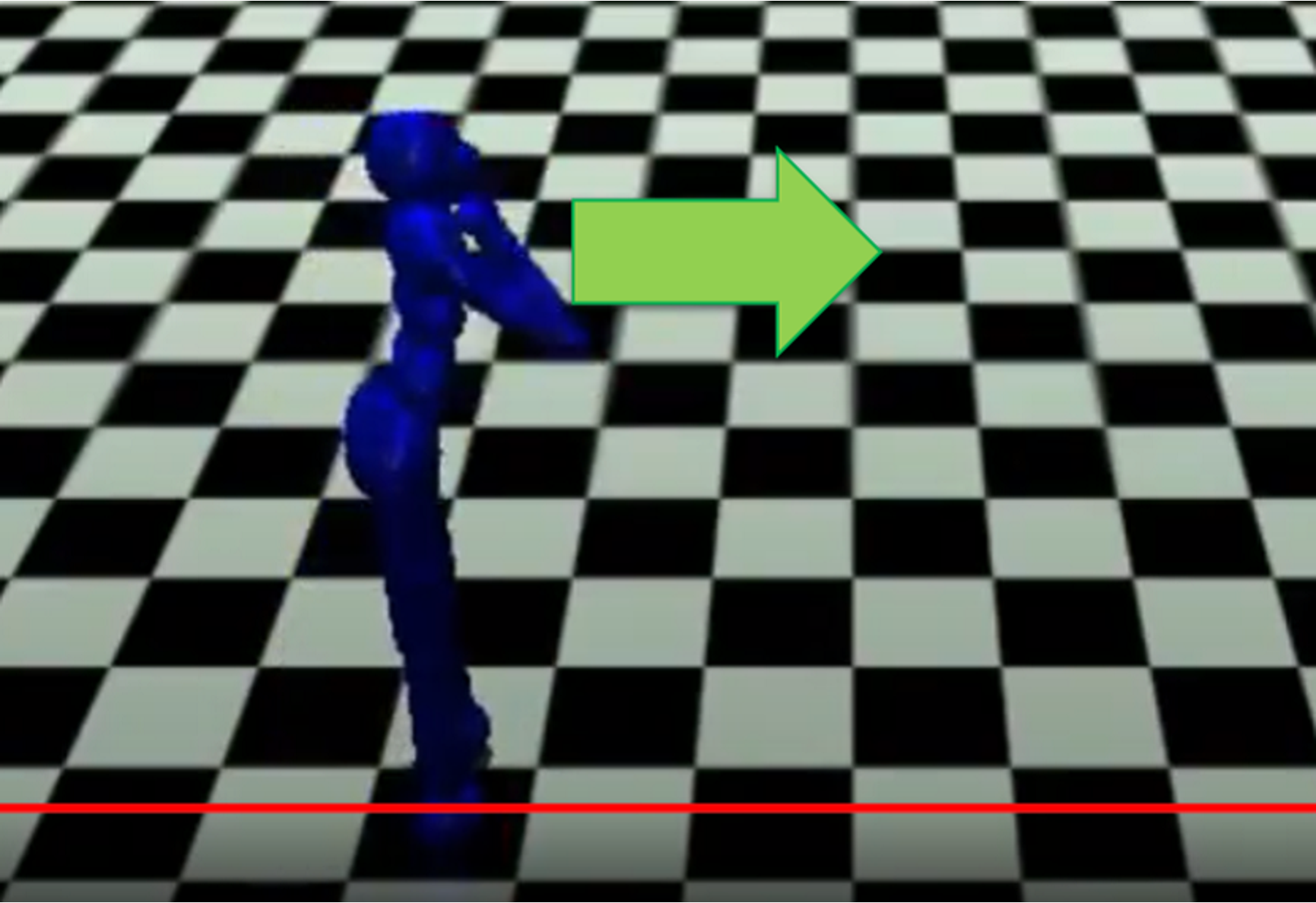}
\includegraphics[clip, width=0.13\hsize]{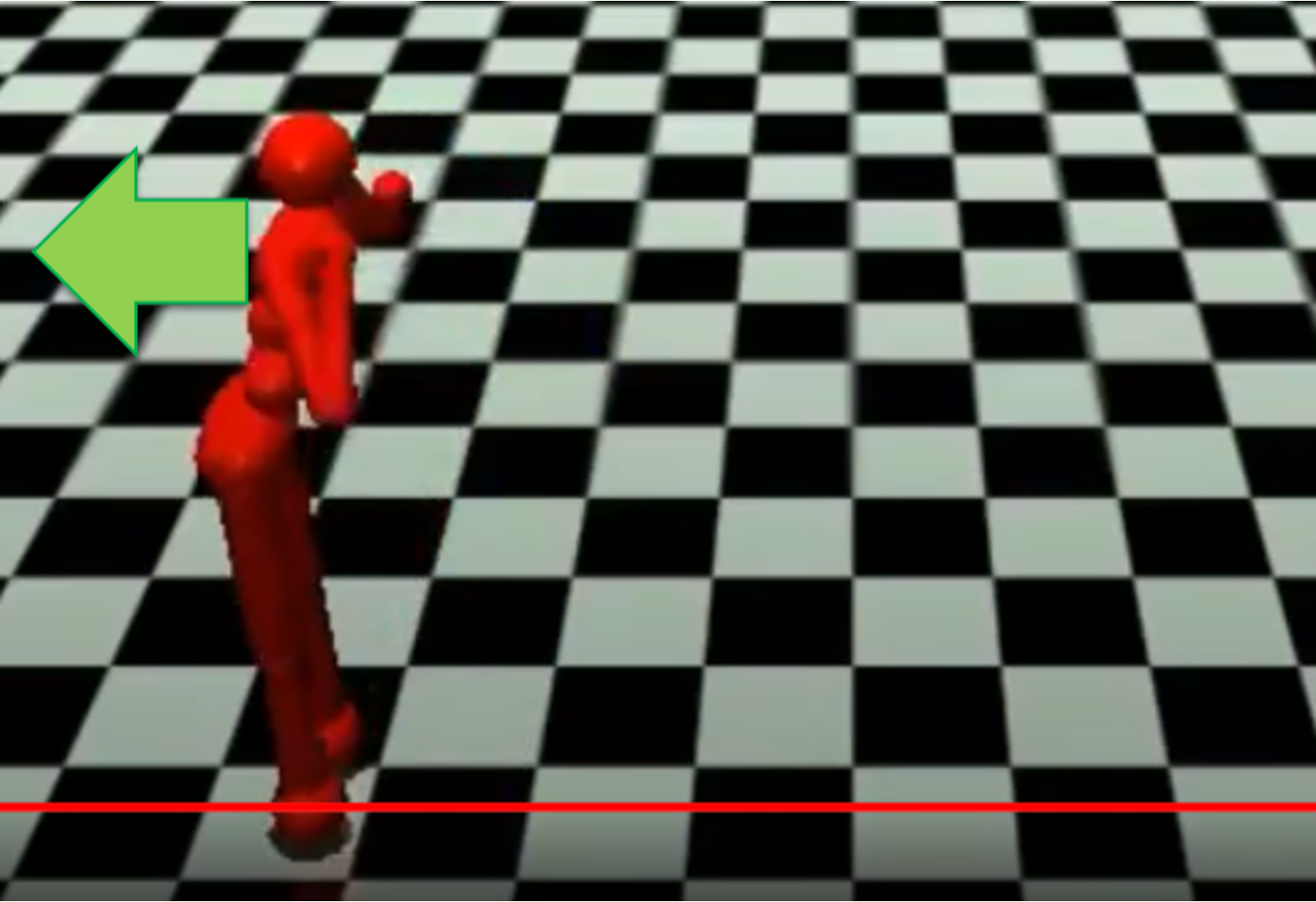}
  \end{center}
%\vspace{-0.9\baselineskip}
  \vspace{-1.8\baselineskip}
\caption{Example of a policy learned by M3PO with 200k training samples, which is equal to around 11 hours of real-world experience, in Humanoid-direc. 
The humanoid is highlighted in accordance with a task (red: move to left, and blue: move to right). 
The figures are time-lapse snapshots, and the first figure on the left is the snapshot at the beginning time. 
The green arrows show the humanoid's movement direction. 
Figures show the policy successfully adapting to the task change.}
%\vspace{-0.5\baselineskip}
\label{fig:example}
  \vspace{-1.0\baselineskip}
\end{figure*}

\subsection{Ablation study}
\begin{figure*}[t!]
\begin{center}
\begin{tabular}{c}
\begin{minipage}{1.0\hsize}
\begin{center}
\includegraphics[clip, width=0.25\hsize]{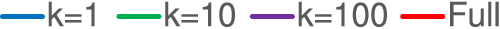}
\end{center}
\end{minipage}\\\hline
\begin{minipage}{1.0\hsize}
      \includegraphics[clip, width=0.333\hsize]{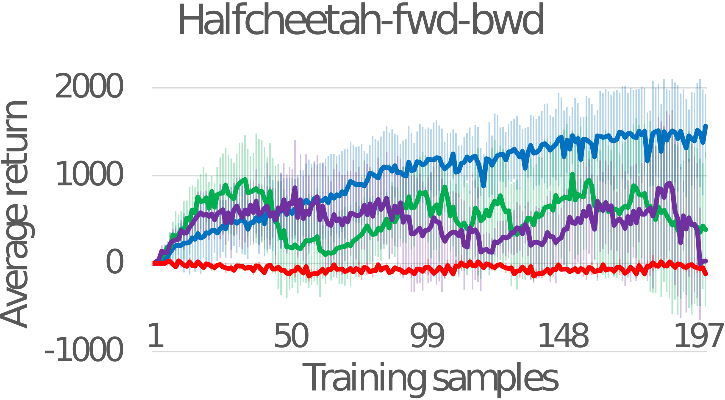}\vline
      \includegraphics[clip, width=0.333\hsize]{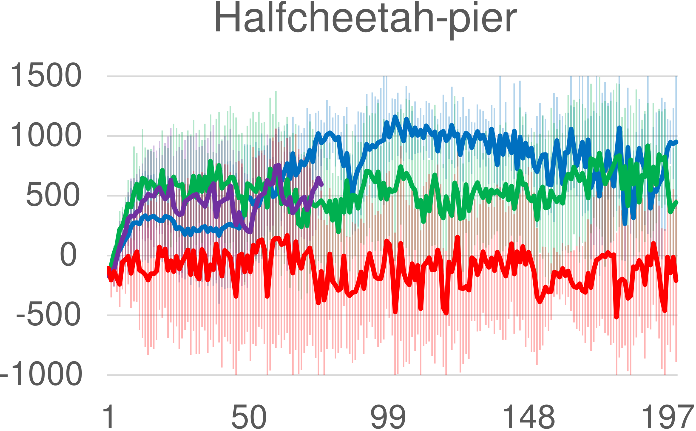}\vline
      \includegraphics[clip, width=0.333\hsize]{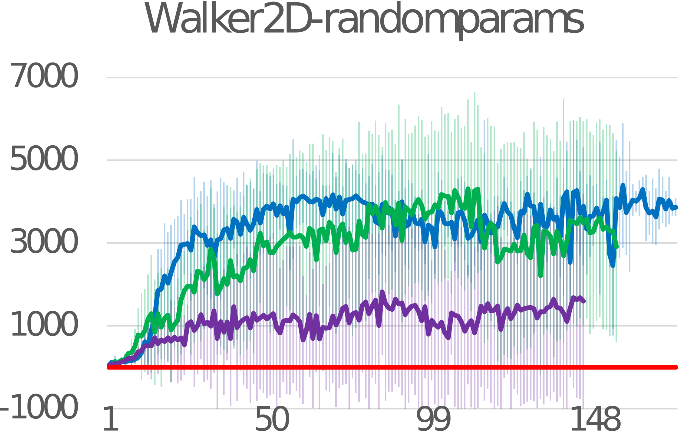}
\end{minipage}\\\hline
\begin{minipage}{1.0\hsize}
      \includegraphics[clip, width=0.333\hsize]{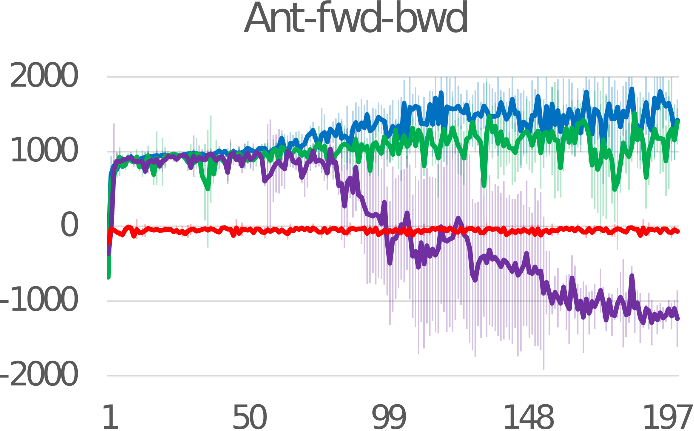}\vline
      \includegraphics[clip, width=0.333\hsize]{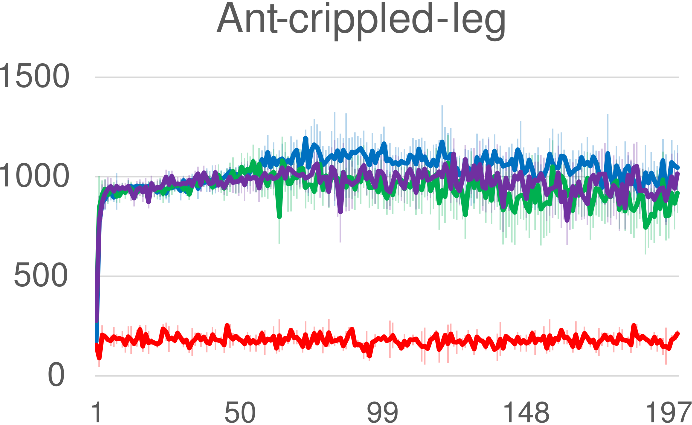}\vline
      \includegraphics[clip, width=0.333\hsize]{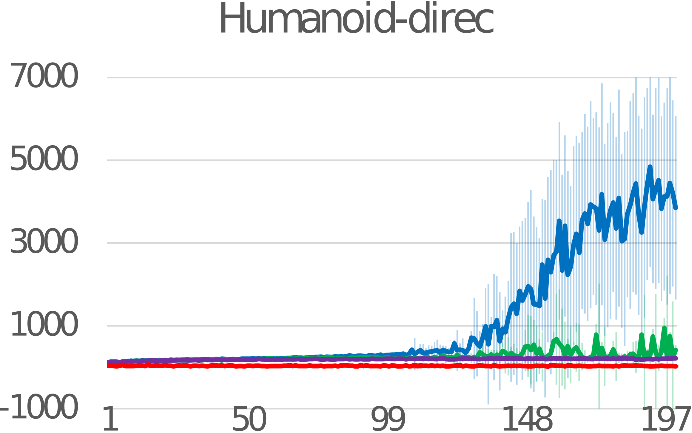}
\end{minipage}
\end{tabular}
\end{center}
%\vspace{-0.9\baselineskip}
  \vspace{-1.8\baselineskip}
\caption{The learning curves of 1) M3PO with different values of model-based rollout length $k$, and 2) a variant of M3PO where we use the full-model-based rollout instead of the branched rollout. 
%Blue lines represent M3PO's learning curves in Figure~\ref{fig:lc}. 
}
\vspace{-0.5\baselineskip}
\label{fig:difk}
\end{figure*}

%-[Q2]
%- [導入]- [異なるkとの比較]
In the next experiment, we perform ablation study to evaluate M3PO with different rollout length $k$ and M3PO with full model-based rollout. 
%deppre First, we perform ablation study to evaluate M3PO with different rollout length $k$ and M3PO with the full model-based rollout. 
%deppre Regarding Q2, we evaluate M3PO by varying its model-based rollout length $k$. 
The evaluation results (Figure~\ref{fig:difk}) show that 1) the performance of M3PO tends to degrade when its model-based rollout length is long, and 2) the performance of M3PO with $k=1$ is mostly the best in all environments. 
These results are consistent with our theoretical result in Section~\ref{subsec:inter-mbrlandmfrl}. 

%-[Q3]
Next, we evaluate a variant of M3PO where we use the full-model-based rollout instead of the branched rollout. 
The learning curves of this instance are plotted as \textbf{Full} in Figure~\ref{fig:difk}. 
We can see that the branched rollouts (M3PO) with $k=1$ is much better than the full-model-based rollout (Full) in all environments. 

\section{Conclusion}
%-[研究内容]
In this paper, we analyzed the performance guarantee of the model-based meta-reinforcement learning (RL) method. 
%--[理論拡張]
We first formulated model-based meta-RL as solving a special case of partially observable Markov decision processes. 
We then theoretically analyzed the performance guarantee of the branched rollout in the meta-RL setting. 
We showed that the branched rollout has a more tightly guaranteed performance than the full model-based rollouts. 
Motivated by the theoretical result, we proposed Meta-Model-Based Meta-Policy Optimization (M3PO), a practical model-based meta-RL method based on branched rollouts. 
%--[実験]
Our experimental results show that M3PO outperforms PEARL and L2A.

%\vspace{50\baselineskip}
%\bibliographystyle{icml2021.bst}
\bibliography{acml21}

\begin{thebibliography}{35}
\providecommand{\natexlab}[1]{#1}
\providecommand{\url}[1]{\texttt{#1}}
\expandafter\ifx\csname urlstyle\endcsname\relax
  \providecommand{\doi}[1]{doi: #1}\else
  \providecommand{\doi}{doi: \begingroup \urlstyle{rm}\Url}\fi

\bibitem[Al-Shedivat et~al.(2018)Al-Shedivat, Bansal, Burda, Sutskever,
  Mordatch, and Abbeel]{al2017continuous}
Maruan Al-Shedivat, Trapit Bansal, Yuri Burda, Ilya Sutskever, Igor Mordatch,
  and Pieter Abbeel.
\newblock Continuous adaptation via meta-learning in nonstationary and
  competitive environments.
\newblock In \emph{Proc. ICLR}, 2018.

\bibitem[Andrychowicz et~al.(2020)Andrychowicz, Baker, Chociej, Józefowicz,
  McGrew, Pachocki, Petron, Plappert, Powell, Ray, Schneider, Sidor, Tobin,
  Welinder, Weng, and Zaremba]{doi:10.1177/0278364919887447}
Marcin Andrychowicz, Bowen Baker, Maciek Chociej, Rafal Józefowicz, Bob
  McGrew, Jakub Pachocki, Arthur Petron, Matthias Plappert, Glenn Powell, Alex
  Ray, Jonas Schneider, Szymon Sidor, Josh Tobin, Peter Welinder, Lilian Weng,
  and Wojciech Zaremba.
\newblock Learning dexterous in-hand manipulation.
\newblock \emph{The International Journal of Robotics Research}, 39\penalty0
  (1):\penalty0 3--20, 2020.
\newblock \doi{10.1177/0278364919887447}.

\bibitem[Cho et~al.(2014)Cho, Van~Merri{\"e}nboer, Gulcehre, Bahdanau,
  Bougares, Schwenk, and Bengio]{cho2014learning}
Kyunghyun Cho, Bart Van~Merri{\"e}nboer, Caglar Gulcehre, Dzmitry Bahdanau,
  Fethi Bougares, Holger Schwenk, and Yoshua Bengio.
\newblock Learning phrase representations using {RNN} encoder-decoder for
  statistical machine translation.
\newblock In \emph{Proc. EMNLP}, 2014.

\bibitem[Chua et~al.(2018)Chua, Calandra, McAllister, and
  Levine]{NEURIPS2018_3de568f8}
Kurtland Chua, Roberto Calandra, Rowan McAllister, and Sergey Levine.
\newblock Deep reinforcement learning in a handful of trials using
  probabilistic dynamics models.
\newblock In \emph{Proc. NeurIPS}, 2018.

\bibitem[Clavera et~al.(2018)Clavera, Rothfuss, Schulman, Fujita, Asfour, and
  Abbeel]{clavera2018model}
Ignasi Clavera, Jonas Rothfuss, John Schulman, Yasuhiro Fujita, Tamim Asfour,
  and Pieter Abbeel.
\newblock Model-based reinforcement learning via meta-policy optimization.
\newblock In \emph{Proc. CoRL}, pages 617--629, 2018.

\bibitem[Duan et~al.(2017)Duan, Schulman, Chen, Bartlett, Sutskever, and
  Abbeel]{duan2016rl}
Yan Duan, John Schulman, Xi~Chen, Peter~L Bartlett, Ilya Sutskever, and Pieter
  Abbeel.
\newblock {RL}$^2$: Fast reinforcement learning via slow reinforcement
  learning.
\newblock In \emph{Proc. ICLR}, 2017.

\bibitem[Feinberg et~al.(2018)Feinberg, Wan, Stoica, Jordan, Gonzalez, and
  Levine]{feinberg2018mbve}
Vladimir Feinberg, Alvin Wan, Ion Stoica, Michael~I. Jordan, Joseph~E.
  Gonzalez, and Sergey Levine.
\newblock Model-based value expansion for efficient model-free reinforcement
  learning.
\newblock In \emph{Proc. ICML}, 2018.

\bibitem[Finn and Levine(2018)]{finn2017meta}
Chelsea Finn and Sergey Levine.
\newblock Meta-learning and universality: Deep representations and gradient
  descent can approximate any learning algorithm.
\newblock In \emph{Proc. ICLR}, 2018.

\bibitem[Finn et~al.(2017)Finn, Abbeel, and Levine]{finn2017model}
Chelsea Finn, Pieter Abbeel, and Sergey Levine.
\newblock Model-agnostic meta-learning for fast adaptation of deep networks.
\newblock In \emph{Proc. ICML}, pages 1126--1135, 2017.

\bibitem[Gupta et~al.(2018)Gupta, Mendonca, Liu, Abbeel, and
  Levine]{gupta2018meta}
Abhishek Gupta, Russell Mendonca, YuXuan Liu, Pieter Abbeel, and Sergey Levine.
\newblock Meta-reinforcement learning of structured exploration strategies.
\newblock In \emph{Proc. NeurIPS}, 2018.

\bibitem[Haarnoja et~al.(2018)Haarnoja, Zhou, Abbeel, and
  Levine]{haarnoja2018soft}
Tuomas Haarnoja, Aurick Zhou, Pieter Abbeel, and Sergey Levine.
\newblock Soft actor-critic: Off-policy maximum entropy deep reinforcement
  learning with a stochastic actor.
\newblock In \emph{Proc. ICML}, pages 1856--1865, 2018.

\bibitem[Henaff(2019)]{henaff2019explicit}
Mikael Henaff.
\newblock Explicit explore-exploit algorithms in continuous state spaces.
\newblock In \emph{Proc. NeurIPS}, 2019.

\bibitem[Humplik et~al.(2019)Humplik, Galashov, Hasenclever, Ortega, Teh, and
  Heess]{humplik2019meta}
Jan Humplik, Alexandre Galashov, Leonard Hasenclever, Pedro~A Ortega, Yee~Whye
  Teh, and Nicolas Heess.
\newblock Meta reinforcement learning as task inference.
\newblock \emph{arXiv preprint arXiv:1905.06424}, 2019.

\bibitem[Janner et~al.(2019)Janner, Fu, Zhang, and Levine]{janner2019trust}
Michael Janner, Justin Fu, Marvin Zhang, and Sergey Levine.
\newblock When to trust your model: Model-based policy optimization.
\newblock In \emph{Proc. NeurIPS}, 2019.

\bibitem[Luo et~al.(2018)Luo, Xu, Li, Tian, Darrell, and
  Ma]{luo2018algorithmic}
Yuping Luo, Huazhe Xu, Yuanzhi Li, Yuandong Tian, Trevor Darrell, and Tengyu
  Ma.
\newblock Algorithmic framework for model-based deep reinforcement learning
  with theoretical guarantees.
\newblock In \emph{Proc. ICLR}, 2018.

\bibitem[Mendonca et~al.(2019)Mendonca, Gupta, Kralev, Abbeel, Levine, and
  Finn]{mendonca2019guided}
Russell Mendonca, Abhishek Gupta, Rosen Kralev, Pieter Abbeel, Sergey Levine,
  and Chelsea Finn.
\newblock Guided meta-policy search.
\newblock In \emph{Proc. NeurIPS}, pages 9653--9664, 2019.

\bibitem[Mishra et~al.(2018)Mishra, Rohaninejad, Chen, and Abbeel]{mishra2018a}
Nikhil Mishra, Mostafa Rohaninejad, Xi~Chen, and Pieter Abbeel.
\newblock A simple neural attentive meta-learner.
\newblock In \emph{Proc. ICLR}, 2018.

\bibitem[Nagabandi et~al.(2019{\natexlab{a}})Nagabandi, Clavera, Liu, Fearing,
  Abbeel, Levine, and Finn]{nagabandilearning}
Anusha Nagabandi, Ignasi Clavera, Simin Liu, Ronald~S Fearing, Pieter Abbeel,
  Sergey Levine, and Chelsea Finn.
\newblock Learning to adapt in dynamic, real-world environments via
  meta-reinforcement learning.
\newblock In \emph{Proc. ICLR}, 2019{\natexlab{a}}.

\bibitem[Nagabandi et~al.(2019{\natexlab{b}})Nagabandi, Finn, and
  Levine]{nagabandi2018deep}
Anusha Nagabandi, Chelsea Finn, and Sergey Levine.
\newblock Deep online learning via meta-learning: Continual adaptation for
  model-based {RL}.
\newblock In \emph{Proc. ICLR}, 2019{\natexlab{b}}.

\bibitem[Perez et~al.(2020)Perez, Such, and Karaletsos]{perez2020generalized}
Christian~F Perez, Felipe~Petroski Such, and Theofanis Karaletsos.
\newblock Generalized hidden parameter {MDP}s transferable model-based {RL} in
  a handful of trials.
\newblock In \emph{Proc. AAAI}, 2020.

\bibitem[Rajeswaran et~al.(2020)Rajeswaran, Mordatch, and
  Kumar]{rajeswaran2020game}
Aravind Rajeswaran, Igor Mordatch, and Vikash Kumar.
\newblock A game theoretic framework for model based reinforcement learning,
  2020.

\bibitem[Rakelly et~al.(2019)Rakelly, Zhou, Finn, Levine, and
  Quillen]{rakelly2019efficient}
Kate Rakelly, Aurick Zhou, Chelsea Finn, Sergey Levine, and Deirdre Quillen.
\newblock Efficient off-policy meta-reinforcement learning via probabilistic
  context variables.
\newblock In \emph{Proc. ICML}, pages 5331--5340, 2019.

\bibitem[Ramachandran et~al.(2017)Ramachandran, Zoph, and
  Le]{ramachandran2017searching}
Prajit Ramachandran, Barret Zoph, and Quoc~V Le.
\newblock Searching for activation functions.
\newblock \emph{arXiv preprint arXiv:1710.05941}, 2017.

\bibitem[Rothfuss et~al.(2019)Rothfuss, Lee, Clavera, Asfour, and
  Abbeel]{rothfuss2018promp}
Jonas Rothfuss, Dennis Lee, Ignasi Clavera, Tamim Asfour, and Pieter Abbeel.
\newblock {ProMP}: Proximal meta-policy search.
\newblock In \emph{Proc. ICLR}, 2019.

\bibitem[S{\ae}mundsson et~al.(2018)S{\ae}mundsson, Hofmann, and
  Deisenroth]{saemundsson2018meta}
Steind{\'o}r S{\ae}mundsson, Katja Hofmann, and Marc~Peter Deisenroth.
\newblock Meta reinforcement learning with latent variable {G}aussian
  processes.
\newblock \emph{arXiv preprint arXiv:1803.07551}, 2018.

\bibitem[Shen et~al.(2020)Shen, Zhao, Zhang, and Yu]{NEURIPS2020_1dc3a89d}
Jian Shen, Han Zhao, Weinan Zhang, and Yong Yu.
\newblock Model-based policy optimization with unsupervised model adaptation.
\newblock In \emph{Proc. NeurIPS}, 2020.

\bibitem[Silver and Veness(2010)]{silver2010monte}
David Silver and Joel Veness.
\newblock Monte-{C}arlo planning in large {POMDP}s.
\newblock In \emph{Proc. NIPS}, pages 2164--2172, 2010.

\bibitem[Stadie et~al.(2018)Stadie, Yang, Houthooft, Chen, Duan, Wu, Abbeel,
  and Sutskever]{stadie2018some}
Bradly~C Stadie, Ge~Yang, Rein Houthooft, Xi~Chen, Yan Duan, Yuhuai Wu, Pieter
  Abbeel, and Ilya Sutskever.
\newblock Some considerations on learning to explore via meta-reinforcement
  learning.
\newblock \emph{arXiv preprint arXiv:1803.01118}, 2018.

\bibitem[Sutton(1991)]{sutton1991dyna}
Richard~S Sutton.
\newblock Dyna, an integrated architecture for learning, planning, and
  reacting.
\newblock \emph{ACM Sigart Bulletin}, 2\penalty0 (4):\penalty0 160--163, 1991.

\bibitem[Todorov et~al.(2012)Todorov, Erez, and Tassa]{todorov2012mujoco}
Emanuel Todorov, Tom Erez, and Yuval Tassa.
\newblock {MuJoCo}: A physics engine for model-based control.
\newblock In \emph{Proc. IROS}, pages 5026--5033. IEEE, 2012.

\bibitem[Wang et~al.(2016)Wang, Kurth-Nelson, Tirumala, Soyer, Leibo, Munos,
  Blundell, Kumaran, and Botvinick]{wang2016learning}
Jane~X Wang, Zeb Kurth-Nelson, Dhruva Tirumala, Hubert Soyer, Joel~Z Leibo,
  Remi Munos, Charles Blundell, Dharshan Kumaran, and Matt Botvinick.
\newblock Learning to reinforcement learn.
\newblock \emph{arXiv preprint arXiv:1611.05763}, 2016.

\bibitem[Williams et~al.(2015)Williams, Aldrich, and
  Theodorou]{williams2015model}
Grady Williams, Andrew Aldrich, and Evangelos Theodorou.
\newblock Model predictive path integral control using covariance variable
  importance sampling.
\newblock \emph{arXiv preprint arXiv:1509.01149}, 2015.

\bibitem[Yu et~al.(2019)Yu, Quillen, He, Julian, Hausman, Finn, and
  Levine]{yu2019meta}
Tianhe Yu, Deirdre Quillen, Zhanpeng He, Ryan Julian, Karol Hausman, Chelsea
  Finn, and Sergey Levine.
\newblock Meta-world: A benchmark and evaluation for multi-task and meta
  reinforcement learning.
\newblock In \emph{Proc. CoRL}, 2019.

\bibitem[Yu et~al.(2020)Yu, Thomas, Yu, Ermon, Zou, Levine, Finn, and
  Ma]{NEURIPS2020_a322852c}
Tianhe Yu, Garrett Thomas, Lantao Yu, Stefano Ermon, James~Y Zou, Sergey
  Levine, Chelsea Finn, and Tengyu Ma.
\newblock {MOPO:} model-based offline policy optimization.
\newblock In \emph{Proc. NeurIPS}, 2020.

\bibitem[Zintgraf et~al.(2020)Zintgraf, Shiarlis, Igl, Schulze, Gal, Hofmann,
  and Whiteson]{zintgraf2020varibad}
Luisa Zintgraf, Kyriacos Shiarlis, Maximilian Igl, Sebastian Schulze, Yarin
  Gal, Katja Hofmann, and Shimon Whiteson.
\newblock {VariBAD}: A very good method for {B}ayes-adaptive deep {RL} via
  meta-learning.
\newblock In \emph{Proc. ICLR}, 2020.

\end{thebibliography}

\clearpage
\appendix
%\onecolumn
\section{}\label{sec:appen}
%\section{Appendices}

\subsection{The relation of returns (performance guarantee) under $k$-step branched rollouts in Markov decision processes}\label{sec:boundinmdp}

%-[MDPs definition]
In this section, we analyze the relation of a true return and a model return in an MDP, which is defined by a tuple $\langle \mathcal{S}, \mathcal{A}, r, \gamma, p_{\text{st}} \rangle$. 
%-[MDP explanation]
Here, $\mathcal{S}$ is a set of states, $\mathcal{A}$ is a set of actions, $p_{\text{st}} = \mathcal{S} \times \mathcal{S} \times \mathcal{A} \rightarrow [0, 1]$ is the state transition probability, $r : \mathcal{S} \times \mathcal{A} \rightarrow \mathbb{R}$ is a reward function and $\gamma \in [0, 1)$ is a discount factor. 
At time step $t$, the state transition probability and reward function are used as $p(s_t |s_{t-1}, a_{t-1})$ and $r(s_t,~ a_t)$, respectively. 
%-[True return]
The true return is defined as $\mathbb{E}_{a \sim \pi, s \sim p}\left[ R = \sum_{t=0}^{\infty} \gamma^t r_t \right]$, where $\pi$ is the agent's current policy. 
%-[Model return]
In addition, the model return is defined by $\mathbb{E}_{(a, s) \sim m_{b}(\pi, p_\theta, \mathcal{D}_{\text{env}})} \left[ R \right]$, where $m(\pi, p_\theta, \mathcal{D}_{\text{env}})$ is the state-action visitation probability based upon an abstract model-based rollout method, which can be calculated on the basis of the $\pi$, $p_\theta$, and $\mathcal{D}_{\text{env}}$. 
$p_{\theta}$ is the predictive model for the next state. 
$\mathcal{D}_{\text{env}}$ is the dataset in which the real trajectories collected by a data collection policy $\pi_{\mathcal{D}}$ is stored. 

%-[relation and discrepancy]
We analyze the relation of the returns, which takes the form of 
\begin{eqnarray}
\mathbb{E}_{a \sim \pi, s \sim p}\left[ R \right] \geq \mathbb{E}_{(a, s) \sim m(\pi, p_\theta, \mathcal{D}_{\text{env}})} \left[ R \right] - C(\epsilon_m, \epsilon_\pi), \nonumber
\end{eqnarray}
where $C(\epsilon_m, \epsilon_\pi)$ is the discrepancy between the returns, which can be expressed as the function of two error quantities $\epsilon_m$ and $\epsilon_\pi$. 
%--[二つのエラーの定義]
Here, $\epsilon_m = \max_{t} \mathbb{E}_{a_t \sim \pi_{\mathcal{D}}, s_t \sim p} \left[ D_{\text{TV}} \left( p(s_{t+1} | s_t, a_t) || p_\theta(s_{t+1} | s_t, a_t) \right) \right]$ and $\epsilon_\pi = \max_{s_t} D_{\text{TV}} \left(\pi(a_t | s_t) || \pi_{\mathcal{D}}(a_t | s_t) \right)$. 
%\vspace{-0.2\baselineskip}
%
%-[Additional notation a d]
In addition, we define the upper bounds of the reward scale as $r_{\text{max}} > \max_{s, a} |r(s, a)|$. 
Note that, in this section, to discuss the MDP case, we are overriding the definition of the variables and functions that were defined for the POMDP case in the main content. %deppre Sections~\ref{sec:preliminaries} and \ref{sec:formulatingMBRL}. 

%-[Janner et al.]
\citet{janner2019trust} analyzed the relations of the returns under the full-model-based rollout and that under the branched rollout in the MDP: \\
\textbf{Theorem 4.1. in \citet{janner2019trust}}. 
\textit{Under the full-model-based rollout in the MDP, the following inequality holds, }
\begin{dmath}
\mathbb{E}_{a \sim \pi, s \sim p} \left[ R \right] \geq \mathbb{E}_{(a, s) \sim m_{f}(\pi, p_\theta, \mathcal{D}_{\text{env}})} \left[ R \right] - 2 r_{\text{max}} \left\{ \frac{\gamma}{(1-\gamma)^2} ( \epsilon_m + 2\epsilon_\pi) + \frac{2}{(1-\gamma)} \epsilon_m \right\}, 
\end{dmath}
\textit{where $\mathbb{E}_{(a, s) \sim m_{f}(\pi, p_\theta, \mathcal{D}_{\text{env}})} \left[ R \right]$ is the model return under the full-model-based rollout. $m_f(\pi_\phi, p_\theta, \mathcal{D}_{\text{env}})$ is the state-action visitation probability under the full model-based rollout method.} \\
\textbf{Theorem 4.2. in \citet{janner2019trust}}. 
\textit{Under the branched  rollout in the MDP, the following inequality holds, }
\begin{dmath}
\mathbb{E}_{a \sim \pi, s \sim p} \left[ R \right] \geq \mathbb{E}_{(a, s) \sim m_{b}(\pi, p_\theta, \mathcal{D}_{\text{env}})} \left[ R \right] - 2 r_{\text{max}} \left\{ \frac{\gamma^{k+1}}{(1-\gamma)^2} \epsilon_\pi + \frac{\gamma^{k} + 2}{(1-\gamma)} \epsilon_\pi + \frac{k}{1-\gamma} (\epsilon_m + 2 \epsilon_{\pi}) \right\}, 
\end{dmath}
\textit{where $\mathbb{E}_{(a, s) \sim m_{b}(\pi, p_\theta, \mathcal{D}_{\text{env}})} \left[ R \right]$ is the model return under the branched rollout. $m_b(\pi_\phi, p_\theta, \mathcal{D}_{\text{env}})$ is the state-action visitation probability under the branched rollout method.} 

\begin{figure}[t]
\begin{center}
\includegraphics[clip, width=0.8\hsize]{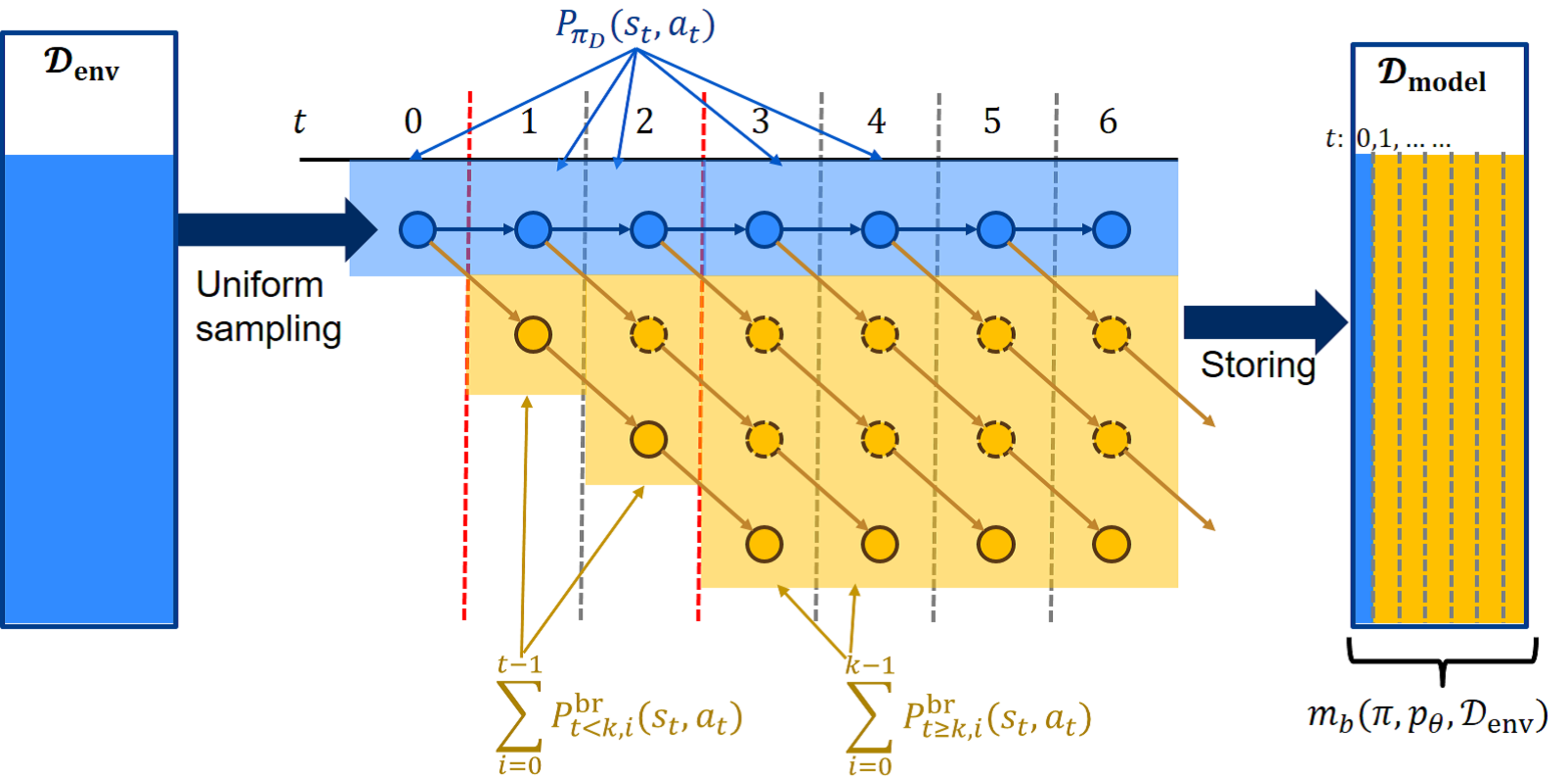}
\end{center}
%\vspace{-1.0\baselineskip}
\caption{The notion of $m_b(\pi_\phi, p_\theta, \mathcal{D}_{\text{env}})$. 
The figure shows the case of the branched rollout method with $k=3$. 
The blue nodes represent real trajectories contained in $\mathcal{D}_{\text{env}}$, and the yellow nodes represent fictitious trajectories generated by $k$-step model-based rollouts with $\pi$ and $p_\theta$. 
The fictitious trajectories are stored in $D_{\text{model}}$. 
The distribution of the trajectories stored in $D_{\text{model}}$ is used as $m_b(\pi, p_\theta, \mathcal{D}_{\text{env}})$. 
%-[Comparison with janner's assummption]
\textbf{In the analysis (derivation of Theorem 4.2) in \citet{janner2019trust}, fictitious trajectories of dashed yellow nodes are not stored in $\mathcal{D}_{\text{model}}$, and thus the state-action visitation probability at each time step is calculated based solely on a single model-based rollout factor. To contrast, in our analysis, these fictitious trajectories are stored in $\mathcal{D}_{\text{model}}$, and thus the state-action visitation probability at each time step is calculated on the basis of multiple model-based rollout factors.}
}
\label{fig:branchrollout}
\end{figure}

%-[branched rollout]
The notion of $m_b(\pi, p_\theta, \mathcal{D}_{\text{env}})$ in the analyses in \citet{janner2019trust} and our case are summarized in Figure~\ref{fig:branchrollout}. 
%-[outline of the branhed rollout]
In the branched rollout method, the real trajectories are uniformly sampled from $\mathcal{D}_{\text{env}}$, and then starting from the sampled trajectories, $k$-step model-based rollouts under $\pi$ and $p_\theta$ are run. 
Then, the fictitious trajectories generated by the branched rollout are stored in a model dataset $D_{\text{model}}$~\footnote{Here, when the trajectories are stored in $D_{\text{model}}$, the states in the trajectories are augmented with time step information to deal with the state transition depending on the time step.}. 
The distribution of the trajectories stored in $D_{\text{model}}$ is used as $m_b(\pi, p_\theta, \mathcal{D}_{\text{env}})$. 
In Figure~\ref{fig:branchrollout}, $p_{\pi_{\mathcal{D}}}(s_t, a_t)$ is the state-action visitation probability under $p_{\text{st}}$ and $\pi_D$. 
This can be used for the initial state distribution for $k$-steps model-based rollouts since the real trajectories are uniformly sampled from $\mathcal{D}_{\text{env}}$. 
In addition, $p_{t < k, i}^{\text{br}}(s_t, a_t)$ and $p_{t \geq k, i}^{\text{br}}(s_t, a_t)$ are the state-action visitation probabilities that the $i$-th yellow fictitious trajectories (nodes) from the bottom at $t$ follow. 

%-[intro Janner]
In the derivation of Theorem 4.2 (more specifically, the proof of Lemma B.4) in \citet{janner2019trust}, important premises are not properly taken into consideration. 
%-[incorrect premise]
In the derivation, state-action visitation probabilities under the branched rollout are affected only by a single model-based rollout factor (see Figure~\ref{fig:branchrollout}). 
For example, a state-action visitation probability at $t$ (s.t. $t>k$) is affected only by the model-based rollout branched from real trajectories at $t-k$ (i.e., $p_{t \geq k, 0}^{\text{br}}(s_t, a_t)$). 
%-[correct premise]
However, state-action visitation probabilities (except for ones at $t=0$ and $t=1$) should be affected by multiple past model-based rollouts. 
For example, a state-action visitation probability at $t$ (s.t. $t>k$) should be affected by the model-based rollout branched from real trajectories at $t-k$ and ones from $t-k+1$ to $t-1$ (i.e., $p_{t \geq k, 0}^{\text{br}}(s_t, a_t) ..., p_{t \geq k, k-1}^{\text{br}}(s_t, a_t)$). 
In addition, in their analysis, they consider that $k$ is an element of the set of non-negative integers. 
However, if $k=0$, the fictitious trajectories for $\mathcal{D}_{\text{model}}$ are not generated, and the distribution of trajectories in $\mathcal{D}_{\text{model}}$ cannot be built. 
Therefore, $k$'s value should not be 0 (it should be an element of the set of non-zero natural numbers $\mathbb N_{> 0}$). 
%-[problem]
These oversights of important premises in their analysis induce a large mismatch between those for their theorem (Theorem 4.2) and those made for the actual implementation of the branched rollout (lines 5--8 in Algorithm 2 in \citet{janner2019trust}).

%[modeiffication ]
Hence, we will newly analyze the relation of the returns under the branched rollout method, considering the aforementioned premises more properly. 
%-[branched rolloutの説明]
Concretely, we consider the multiple model-based rollout factors for $m_{b}(\pi, p_\theta, \mathcal{D}_{\text{env}})$ (See Figure~\ref{fig:branchrollout})~\footnote{Considering the discussion in the last paragraph, we also limit the range of $k$ as $k \in \mathbb{N}_{> 0}$ in our analysis.}. 
%
%-[状態訪問確率の定義]
With this consideration, we define the model-return under the branched rollout $\mathbb{E}_{(a, s) \sim m_{b}(\pi, p_\theta, \mathcal{D}_{\text{env}})} [R]$ as:
\begin{eqnarray}
\mathbb{E}_{(a, s) \sim m_{b}(\pi, p_\theta, \mathcal{D}_{\text{env}})} [R] &=& \sum_{s_0, a_0} p_{\pi_{\mathcal{D}}}(s_0, a_0) r(s_0, a_0) + \sum_{t=1}^{k-1} \sum_{s_t, a_t} \gamma^t p_{t<k}^{\text{br}}(s_{t}, a_{t}) r(s_t, a_t) \nonumber\\&& + \sum_{t=k}^{\infty} \sum_{s_t, a_t} \gamma^t p_{t\geq k}^{\text{br}}(s_{t}, a_{t}) r(s_t, a_t)\\
p_{t < k}^{\text{br}}(s_{t}, a_{t}) &=& \textcolor{red}{\frac{1}{t} \sum_{i=0}^{t-1} p_{t<k, i}^{\text{br}}(s_t, a_t)}\\
p_{t\geq k}^{\text{br}}(s_{t}, a_{t}) &=& \textcolor{red}{\frac{1}{k} \sum_{i=0}^{k-1} p_{t\geq k, i}^{\text{br}}(s_t, a_t)} \\
p_{t < k, i}^{\text{br}}(s_t, a_t) &=& \sum_{s_{i}, ..., s_{t-1}} \sum_{a_{i}, ..., a_{t-1}} p_{\pi_{\mathcal{D}}}(s_{i}) \Pi_{j=i}^{t-1} p_{\theta}(s_{j+1} | s_{j}, a_{j}) \pi(a_{j} | s_{j})\\
p_{t \geq k, i}^{\text{br}}(s_t, a_t) &=& \sum_{s_{t-k+i}, ..., s_{t-1}} \sum_{a_{t-k+i}, ..., a_{t-1}} \nonumber\\&& p_{\pi_{\mathcal{D}}}(s_{t-k+i}) \Pi_{j=t-k+i}^{t-1} p_{\theta}(s_{j+1} | s_{j}, a_{j}) \pi(a_{j} | s_{j})
\end{eqnarray}
Here, $p_{\pi_{\mathcal{D}}}$ is the state visitation probability under $\pi_{\mathcal{D}}$ and $p_{\text{st}}$. 
The ones modified from \citet{janner2019trust} are highlighted in red. 
In the remaining paragraphs in this section, we will derive the new theorems for the relation of the returns under this definition of the model return. 

%-[Derivation]
Before starting the derivation of our theorem, we introduce a useful lemma. 
\begin{lemma1}\label{lemma:boundrollout}
Assume that the rollout process in which the policy and dynamics can be switched to other ones at time step $t_{\text{sw}}$. 
Letting two probabilities be $p_1$ and $p_2$, for $1 \leq t' \leq t_{\text{sw}}$, we assume that the dynamics distributions are bounded as \\ $\epsilon_{m, \text{pre}} = \max_{t'} E_{s \sim p_1} \left[ D_{\text{TV}} \left( p_1(s_{t'} | s_{t'-1}, a_{t'-1}) || p_2(s_{t'} | s_{t'-1}, a_{t'-1}) \right) \right]$. In addition, for $t_{\text{sw}} < t' \leq t$, we assume that the dynamics distributions are bounded as \\ $\epsilon_{m, \text{post}} = \max_{t'} E_{s \sim p_1} \left[ D_{\text{TV}} \left( p_1(s_{t'} | s_{t'-1}, a_{t'-1}) || p_2(s_{t'} | s_{t'-1}, a_{t'-1}) \right) \right]$. 
Likewise, the policy divergence is bounded by $\epsilon_{\pi, \text{pre}}$ and $\epsilon_{\pi, \text{post}}$. 
Then, the following inequality holds
\begin{dmath}
\sum_{s_t, a_t} \left| p_1(s_t, a_t) - p_2(s_t, a_t) \right| \leq 2 (t - t_{\text{sw}}) ( \epsilon_{m, \text{post}} + \epsilon_{\pi, \text{post}}) + 2 t_{\text{sw}} ( \epsilon_{m, \text{pre}} + \epsilon_{\pi, \text{pre}})
\end{dmath}
\end{lemma1}
\begin{proof}
The proof is done in a similar manner to those of Lemma B.1 and B.2 in \citet{janner2019trust}. 

\begin{dmath}
\sum_{s_t, a_t} \left| p_1(s_t, a_t) - p_2(s_t, a_t) \right| = \sum_{s_t, a_t} \left| p_1(s_t)p_1(a_t | s_t) - p_2(s_t) p_2(a_t | s_t) \right| 
= \sum_{s_t, a_t} \left| p_1(s_t)p_1(a_t | s_t) - p_1(s_t)p_2(a_t | s_t) + (p_1(s_t) - p_2(s_t) ) p_2(a_t | s_t) \right| 
\leq \sum_{s_t, a_t} p_1(s_t) \left| p_1(a_t | s_t) - p_2(a_t | s_t) \right| + \sum_{s_t} \left| p_1(s_t) - p_2(s_t) \right|
\leq \sum_{s_t, a_t} p_1(s_t) \left| p_1(a_t | s_t) - p_2(a_t | s_t) \right| + \sum_{s_t, a_{t-1}} \left| p_1(s_t, a_{t-1}) - p_2(s_t, a_{t-1}) \right|
=\sum_{s_t, a_t} p_1(s_t) \left| p_1(a_t | s_t) - p_2(a_t | s_t) \right| \\~+ \sum_{s_t, a_{t-1}} \left| p_1(a_{t-1}) p_1(s_t | a_{t-1}) - p_1(a_{t-1}) p_2(s_t | a_{t-1}) + (p_1(a_{t-1}) - p_2(a_{t-1}) ) p_2(s_t | a_{t-1}) \right| \\
\leq \sum_{s_t, a_t} p_1(s_t) \left| p_1(a_t | s_t) - p_2(a_t | s_t) \right| + \sum_{s_t, a_{t-1}} p_1(a_{t-1}) \left| p_1(s_t | a_{t-1}) - p_2(s_t | a_{t-1}) \right| \\~+ \sum_{a_{t-1}} \left| p_1(a_{t-1}) - p_2(a_{t-1}) \right| 
\leq \sum_{s_t, a_t} p_1(s_t) \left| p_1(a_t | s_t) - p_2(a_t | s_t) \right| + \sum_{s_t, a_{t-1}} p_1(a_{t-1}) \left| p_1(s_t | a_{t-1}) - p_2(s_t | a_{t-1}) \right| \\~+ \sum_{s_{t-1}, a_{t-1}} \left| p_1(s_{t-1}, a_{t-1}) - p_2(s_{t-1}, s_{t-1}) \right|
\leq 2 \epsilon_{m, \text{post}} + 2 \epsilon_{\pi, \text{post}} + \sum_{s_{t-1}, a_{t-1}} \left| p_1(s_{t-1}, a_{t-1}) - p_2(s_{t-1}, s_{t-1}) \right|
\leq 2 (t - t_{\text{sw}}) ( \epsilon_{m, \text{post}} + \epsilon_{\pi, \text{post}}) + \sum_{s_{t_{\text{sw}}}, a_{t_{\text{sw}}}} \left| p_1(s_{t_{\text{sw}}}, a_{t_{\text{sw}}}) - p_2(s_{t_{\text{sw}}}, s_{t_{\text{sw}}}) \right|
\leq 2 (t - t_{\text{sw}}) ( \epsilon_{m, \text{post}} + \epsilon_{\pi, \text{post}}) + 2 t_{\text{sw}} ( \epsilon_{m, \text{pre}} + \epsilon_{\pi, \text{pre}})
\end{dmath}

\end{proof}

%--------------------
%\textcolor{red}(参考：)
% https://mathtrain.jp/ar
% http://hooktail.sub.jp/mathInPhys/infGeoProgres/
% https://www.studyplus.jp/397
Now, we start the derivation of our theorems. 
%-[バウンドの証明]
\begin{theorem1}\label{th:2mdp}
Under the $k \in \mathbb N_{> 0}$ steps branched rollouts in the MDP $\langle \mathcal{S}, \mathcal{A}, r, \gamma, p_{\text{st}} \rangle$, given the bound of the errors $\epsilon_m = \max_{t} \mathbb{E}_{a_t \sim \pi_{\mathcal{D}}, s_t \sim p} \left[ D_{\text{TV}} \left( p(s_{t+1} | s_t, a_t) || p_\theta(s_{t+1} | s_t, a_t) \right) \right]$ and $\epsilon_\pi = \max_{s_t} D_{\text{TV}} \left(\pi(a_t | s_t) || \pi_{\mathcal{D}}(a_t | s_t) \right)$, the following inequality holds, 
\begin{dmath}
\mathbb{E}_{a \sim \pi,s \sim p} \left[ R \right] \geq  \mathbb{E}_{(a, s) \sim m_{b}(\pi, p_\theta, \mathcal{D}_{\text{env}})} \left[ R \right] - r_{\text{max}} \left\{ \frac{1+\gamma^2}{(1-\gamma)^2} 2 \epsilon_\pi + \frac{\gamma -k\gamma^{k} + (k-1) \gamma^{k+1}}{(1-\gamma)^2} \left(\epsilon_\pi + \epsilon_m \right) \\+ \frac{\gamma^{k} - \gamma}{\gamma - 1} (\epsilon_{\pi} + \epsilon_m) + \frac{\gamma^k}{1-\gamma} (k+1) (\epsilon_{\pi} + \epsilon_m) \right\}. 
\end{dmath}
\end{theorem1}
\begin{proof}

\begin{eqnarray}
\left| \mathbb{E}_{a \sim \pi, s \sim p} \left[ R \right] - \mathbb{E}_{(a, s) \sim m_{b}(\pi, p_\theta, \mathcal{D}_{\text{env}})} \left[ R \right] \right| &=& \left|\begin{matrix} \sum_{s_0, a_0} \left\{ p_{\pi}(s_0, a_0) - p_{\pi_{\mathcal{D}}}(s_{0}, a_{0})  \right\} r(s_0, a_0) \\+ \sum_{t=1}^{k-1} \sum_{s_t, a_t} \gamma^t \left\{ p_{\pi}(s_t, a_t) - p_{t<k}^{\text{br}}(s_{t}, a_{t}) \right\} r(s_t, a_t)  \\+ \sum_{t=k}^{\infty} \sum_{s_t, a_t} \gamma^t \left\{ p_{\pi}(s_t, a_t) - p_{t\geq k}^{\text{br}}(s_{t}, a_{t}) \right\} r(s_t, a_t) \end{matrix} \right| \nonumber\\
 &\leq& \left\{ \begin{matrix} \sum_{s_0, a_0} \left| p_{\pi}(s_0, a_0) - p_{\pi_{\mathcal{D}}}(s_{0}, a_{0})  \right| \left| r(s_0, a_0) \right| \\+ \sum_{t=1}^{k-1} \gamma^t \sum_{s_t, a_t} \left| p_{\pi}(s_t, a_t) - p_{t<k}^{\text{br}}(s_{t}, a_{t}) \right| \left| r(s_t, a_t) \right| \\+ \sum_{t=k}^{\infty} \gamma^t \sum_{s_t, a_t} \left| p_{\pi}(s_t, a_t) - p_{t\geq k}^{\text{br}}(s_{t}, a_{t}) \right| \left| r(s_t, a_t) \right| \end{matrix} \right\} \nonumber\\
&\leq& \left\{ \begin{matrix} r_{\text{max}} \underbrace{ \sum_{s_0, a_0} \left| p_{\pi}(s_0, a_0) - p_{\pi_{\mathcal{D}}}(s_{0}, a_{0}) \right| }_{\text{term A}} \\+ r_{\text{max}} \sum_{t=1}^{k-1} \gamma^t \underbrace{ \sum_{s_t, a_t} \left| p_{\pi}(s_t, a_t) - p_{t<k}^{\text{br}}(s_{t}, a_{t}) \right| }_{\text{term B}} \\+ r_{\text{max}} \sum_{t=k}^{\infty} \gamma^t \underbrace{ \sum_{s_t, a_t} \left| p_{\pi}(s_t, a_t) - p_{t\geq k}^{\text{br}}(s_{t}, a_{t}) \right| }_{\text{term C}} \end{matrix} \right\} \label{eq:th1main}
\end{eqnarray}
Here, $p_{\pi}(s_t, a_t)$ is the state-action visitation probability under $p_{\text{st}}$ and $\pi$. 

For \textbf{term A}, we can bound the value in similar manner to the derivation of Lemma~\ref{lemma:boundrollout}: 
\begin{dmath}
\sum_{s_0, a_0} \left| p_{\pi}(s_0, a_0) - p_{\pi_{\mathcal{D}}}(s_{0}, a_{0}) \right| = \sum_{s_0, a_0} \left| p_{\pi}(a_0) p(s_0) -  p_{\pi_{\mathcal{D}}}(a_{0}) p(s_{0}) \right|
= \sum_{s_0, a_0} \left| p_{\pi}(a_0) p(s_0) - p_{\pi}(a_0) p(s_{0}) + \left( p_{\pi}(a_0) -  p_{\pi_{\mathcal{D}}}(a_{0}) \right) p(s_{0}) \right|\\
\leq \underbrace{\sum_{s_0, a_0} p_{\pi}(a_0) \left| p(s_0) - p(s_{0}) \right|}_{=0} + \underbrace{\sum_{a_0} \left| p_{\pi}(a_0) -  p_{\pi_{\mathcal{D}}}(a_{0}) \right| }_{\leq 2 \epsilon_\pi} \\
\leq 2 \epsilon_\pi \label{eq:th1-termA} 
\end{dmath}

For \textbf{term B}, we can apply Lemma \ref{lemma:boundrollout} to bound the value, but it requires the bounded model error under the current policy $\pi$. 
Thus, we need to decompose the distance into two by adding and subtracting $p_{\pi_{\mathcal{D}}}$: 
\begin{eqnarray}
\sum_{s_t, a_t} \left| p_{\pi}(s_t, a_t) - p_{\text{br},t<k}(s_{t}, a_{t}) \right| &=& \sum_{s_t, a_t} \left|\begin{matrix} p_{\pi}(s_t, a_t) - p_{\pi_{\mathcal{D}}}(s_t, a_t) \\+ p_{\pi_{\mathcal{D}}}(s_t, a_t)- p_{t<k}^{\text{br}}(s_{t}, a_{t}) \end{matrix} \right| \nonumber\\
 &\leq& \underbrace{ \sum_{s_t, a_t} \left| p_{\pi}(s_t, a_t) - p_{\pi_{\mathcal{D}}}(s_t, a_t) \right| }_{\leq 2 t \epsilon_\pi} \nonumber\\&& + \sum_{s_t, a_t} \left| p_{\pi_{\mathcal{D}}}(s_t, a_t)- p_{t<k}^{\text{br}}(s_{t}, a_{t}) \right|
\end{eqnarray}
\begin{eqnarray}
\sum_{s_t, a_t} \left| p_{\pi_{\mathcal{D}}}(s_t, a_t) - p_{t<k}^{\text{br}}(s_{t}, a_{t}) \right| &=& \sum_{s_t, a_t} \left| \begin{matrix} \frac{1}{t} \sum_{i=0}^{t-1} p_{\pi_{\mathcal{D}}}(s_t, a_t) - \frac{1}{t} \sum_{i=0}^{t-1} p_{t<k, i}^{\text{br}}(s_t, a_t) \end{matrix} \right| \nonumber\\
 &\leq& \frac{1}{t} \sum_{i=0}^{t-1} \sum_{s_t, a_t} \left| p_{\pi_{\mathcal{D}}}(s_t, a_t)-  p_{t<k, i}^{\text{br}}(s_t, a_t) \right| \nonumber\\
 &\stackrel{(A)}{\leq}& \frac{1}{t} \sum_{i=0}^{t-1} \left\{ 2 \left(t - i \right) \cdot \left( \epsilon_\pi + \epsilon_m \right) \right\} \nonumber\\
 &=& \frac{1}{t} \left\{ t^2 (\epsilon_{\pi} + \epsilon_m) + t (\epsilon_\pi + \epsilon_m) \right\}
\end{eqnarray}
For (A), we apply Lemma~\ref{lemma:boundrollout} with setting $\epsilon_{m, \text{post}}=\epsilon_m$ and $\epsilon_{\pi, \text{post}}=\epsilon_\pi$ for the rollout following $\pi$ and $p_{\theta}$, and $\epsilon_{m, \text{pre}}=0$ and $\epsilon_{\pi, \text{pre}}=0$ for the rollout following $\pi_{\mathcal{D}}$ and $p_{\text{st}}$, respectively. 
To recap \textbf{term B}, the following inequality holds:
\begin{eqnarray}
\sum_{s_t, a_t} \left| p_{\pi}(s_t, a_t) - p_{\text{br},t<k}(s_{t}, a_{t}) \right| &\leq& 2 t \epsilon_\pi + t (\epsilon_{\pi} + \epsilon_m) +  ( \epsilon_{\pi} + \epsilon_m)  \label{eq:th1-termB}
\end{eqnarray}

For \textbf{term C}, we can derive the bound in a similar manner to the term B case:
\begin{eqnarray}
\sum_{s_t, a_t} \left| p_{\pi}(s_t, a_t) - p_{\text{br},t \geq k}(s_{t}, a_{t}) \right| &=& \sum_{s_t, a_t} \left|\begin{matrix} p_{\pi}(s_t, a_t) - p_{\pi_{\mathcal{D}}}(s_t, a_t) \\+ p_{\pi_{\mathcal{D}}}(s_t, a_t)- p_{t \geq k}^{\text{br}}(s_{t}, a_{t}) \end{matrix}\right| \nonumber\\
 &\leq& \underbrace{ \sum_{s_t, a_t} \left| p_{\pi}(s_t, a_t) - p_{\pi_{\mathcal{D}}}(s_t, a_t) \right| }_{\leq 2 t \epsilon_\pi} \nonumber\\&& + \sum_{s_t, a_t} \left| p_{\pi_{\mathcal{D}}}(s_t, a_t)- p_{t\geq k}^{\text{br}}(s_{t}, a_{t}) \right|
\end{eqnarray}
\begin{eqnarray}
\sum_{s_t, a_t} \left| p_{\pi_{\mathcal{D}}}(s_t, a_t) - p_{t\geq k}^{\text{br}}(s_{t}, a_{t}) \right| &=& \sum_{s_t, a_t} \left| \begin{matrix} \frac{1}{k} \sum_{i=0}^{k-1} p_{\pi_{\mathcal{D}}}(s_t, a_t) - \frac{1}{k} \sum_{i=0}^{k-1} p_{t\geq k,i}^{\text{br}}(s_t, a_t) \end{matrix} \right| \nonumber\\
 &\leq& \frac{1}{k} \sum_{i=0}^{k-1} \sum_{s_t, a_t} \left| p_{\pi_{\mathcal{D}}}(s_t, a_t)-  p_{t \geq k,i}^{\text{br}}(s_t, a_t) \right| \nonumber\\
 &\leq& \frac{1}{k} \sum_{i=0}^{k-1} \left\{ 2 \left(k - i \right) \cdot \left( \epsilon_\pi + \epsilon_m \right) \right\} \nonumber\\
 &=& \frac{1}{k} \left\{ k^2 (\epsilon_{\pi} + \epsilon_m) + k (\epsilon_{\pi} + \epsilon_m) \right\} 
\end{eqnarray}
To recap \textbf{term C}, the following equation holds:
\begin{eqnarray}
\sum_{s_t, a_t} \left| p_{\pi}(s_t, a_t) - p_{t \geq k}^{\text{br}}(s_{t}, a_{t}) \right| &\leq& 2 t \epsilon_\pi + k (\epsilon_{\pi} + \epsilon_m) + (\epsilon_{\pi} + \epsilon_m) \label{eq:th1-termC}
\end{eqnarray}

By substituting Eqs.~\ref{eq:th1-termA},~\ref{eq:th1-termB}, and \ref{eq:th1-termC}, into Eq.~\ref{eq:th1main}, we obtain the result: 
\begin{eqnarray}
\left| \mathbb{E}_{a \sim \pi, s \sim p} \left[ R \right] - \mathbb{E}_{(a, s) \sim m_{b}(\pi, p_\theta, \mathcal{D}_{\text{env}})} \left[ R \right] \right| &\leq& \left\{ \begin{matrix} r_{\text{max}} 2 \epsilon_\pi \\+ r_{\text{max}} \sum_{t=1}^{k-1} \gamma^t \left\{ 2 t \epsilon_\pi + t (\epsilon_{\pi} + \epsilon_m) + (\epsilon_{\pi} + \epsilon_m) \right\} \\+ r_{\text{max}} \sum_{t=k}^{\infty} \gamma^t \left\{ 2 t \epsilon_\pi + k (\epsilon_{\pi} + \epsilon_m) + (\epsilon_{\pi} + \epsilon_m) \right\} \end{matrix} \right\} \nonumber\\ 
&=& r_{\text{max}} \left\{ \begin{matrix} 2 \epsilon_\pi +  \frac{1-k\gamma^{(k-1)} + (k-1)\gamma^{k}}{(1-\gamma)^2} \gamma \left( 3 \epsilon_\pi + \epsilon_m \right) + \frac{\gamma^{k} - \gamma}{\gamma - 1} (\epsilon_{\pi} + \epsilon_m) \\+ \sum_{t=k}^{\infty} \gamma^t \left\{ 2 t \epsilon_\pi + k (\epsilon_{\pi} + \epsilon_m) + (\epsilon_{\pi} + \epsilon_m) \right\} \end{matrix} \right\} \nonumber\\ 
&=& r_{\text{max}} \left\{ \begin{matrix} 2 \epsilon_\pi + \frac{1-k\gamma^{(k-1)} + (k-1) \gamma^{k}}{(1-\gamma)^2} \gamma \left( 3 \epsilon_\pi + \epsilon_m \right) + \frac{\gamma^{k} - \gamma}{\gamma - 1} (\epsilon_{\pi} + \epsilon_m) \\+ \sum_{t=1}^{\infty} \gamma^t \left\{ 2 t \epsilon_\pi + k (\epsilon_{\pi} + \epsilon_m) + (\epsilon_{\pi} + \epsilon_m) \right\} \\ - \sum_{t=1}^{k-1} \gamma^t \left\{ 2 t \epsilon_\pi + k (\epsilon_{\pi} + \epsilon_m) + (\epsilon_{\pi} + \epsilon_m) \right\} \end{matrix} \right\} \nonumber\\
&=& r_{\text{max}} \left\{ \begin{matrix} 2 \epsilon_\pi + \frac{1-k\gamma^{(k-1)} + (k-1) \gamma^{k}}{(1-\gamma)^2} \gamma \left( 3 \epsilon_\pi + \epsilon_m \right) + \frac{\gamma^{k} - \gamma}{\gamma - 1} (\epsilon_{\pi} + \epsilon_m) \\+ \frac{2}{(1-\gamma)^2} \gamma \epsilon_\pi + \frac{\gamma}{1-\gamma} \left\{ k (\epsilon_{\pi} + \epsilon_m) + (\epsilon_{\pi} + \epsilon_m) \right\} \\ - \sum_{t=1}^{k-1} \gamma^t \left\{ 2 t \epsilon_\pi + k (\epsilon_{\pi} + \epsilon_m) + (\epsilon_{\pi} + \epsilon_m) \right\} \end{matrix} \right\} \nonumber\\
&=& r_{\text{max}} \left\{ \begin{matrix} 2 \epsilon_\pi + \frac{1-k\gamma^{(k-1)} + (k-1) \gamma^{k}}{(1-\gamma)^2} \gamma \left( 3 \epsilon_\pi + \epsilon_m \right) + \frac{\gamma^{k} - \gamma}{\gamma - 1} (\epsilon_{\pi} + \epsilon_m) \\+ \frac{2}{(1-\gamma)^2} \gamma \epsilon_\pi + \frac{\gamma}{1-\gamma} \left\{ k (\epsilon_{\pi} + \epsilon_m) + (\epsilon_{\pi} + \epsilon_m) \right\} \\- \frac{1-k\gamma^{(k-1)} + (k-1) \gamma^{k}}{(1-\gamma)^2} 2 \gamma \epsilon_\pi \\- \frac{\gamma^{k} - \gamma}{\gamma - 1} \left\{ k (\epsilon_{\pi} + \epsilon_m) + (\epsilon_{\pi} + \epsilon_m) \right\} \end{matrix} \right\} \nonumber\\
&=& r_{\text{max}} \left\{ \begin{matrix} \frac{1+\gamma^2}{(1-\gamma)^2} 2 \epsilon_\pi + \frac{\gamma-k\gamma^{k} + (k-1) \gamma^{k+1}}{(1-\gamma)^2} \left(\epsilon_\pi + \epsilon_m \right) \\+ \frac{\gamma^{k} - \gamma}{\gamma - 1} (\epsilon_{\pi} + \epsilon_m) + \left( \frac{\gamma}{1-\gamma} - \frac{\gamma^{k} - \gamma}{\gamma - 1} \right) (k+1) (\epsilon_{\pi} + \epsilon_m) \end{matrix} \right\} \nonumber\\
&=& r_{\text{max}} \left\{ \begin{matrix} \frac{1+\gamma^2}{(1-\gamma)^2} 2 \epsilon_\pi + \frac{\gamma-k\gamma^{k} + (k-1) \gamma^{k+1}}{(1-\gamma)^2} \left(\epsilon_\pi + \epsilon_m \right) \\+ \frac{\gamma^{k} - \gamma}{\gamma - 1} (\epsilon_{\pi} + \epsilon_m) + \frac{\gamma^k}{1-\gamma} (k+1) (\epsilon_{\pi} + \epsilon_m) \end{matrix} \right\} \label{ieq:resm}
\end{eqnarray}
\end{proof}

\subsection{Proofs of theorems for performance guarantee in the model-based meta-RL setting}\label{sec:proofs}

% proof of main theorems
%-[補題の導出]
Before starting the derivation of the main theorems, we first introduce a lemma useful for bridging POMDPs, our meta-RL setting, and MDPs. 
\begin{lemma1}[\citet{silver2010monte}]
Given a POMDP $\langle \mathcal{O}, \mathcal{S}, \mathcal{A}, p_{\text{ob}}, r, \gamma, p_{\text{st}} \rangle$, consider the derived MDP with histories as states, $\langle \mathcal{H}, \mathcal{A}, \gamma, \bar{r}, p_{\text{hi}} \rangle$, where $p_{\text{hi}} = p(h_{t+1} | a_{t}, h_t) = \sum_{s_t} \sum_{s_{t+1}} p(s_t | h_t) p(s_{t+1} | s_t, a_t) p(o_{t+1} |s_{t+1} , a_t)$ and $\bar{r}(h_t, a_t) = \sum_{s_t} p(s_t | h_t) r(s_t, a_t)$. 
Then, the value function of the derived MDP is equal to that of the POMDP. 
\label{lemma:pomdp2augmdp}
\end{lemma1}
\begin{proof}
The statement can be derived by backward induction on the value functions. 
See the proof of Lemma 1 in \citet{silver2010monte} for details. 
\end{proof}

\begin{lemma1}
Given meta-RL setting in Section~\ref{sec:formulatingMBRL}, consider the derived MDP with histories as states, $\langle \mathcal{H}, \mathcal{A}, \gamma, \bar{r}, p_{\text{hi}} \rangle$, where $p_{\text{hi}} = p(h_{t+1} | a_{t}, h_t) = \sum_{\tau_t} p(\tau_t | h_t) p(o_{t+1} | \tau_t, o_t, a_t)$ and $\bar{r}(h_t, a_t) = \sum_{\tau_t} p(\tau_t | h_t) r(\tau_t, o_t, a_t)$. 
Then, the value function of the derived MDP is equal to that in the meta-RL setting in Section~\ref{sec:formulatingMBRL}. 
\label{lemma:metarl2augmdp}
\end{lemma1}
\begin{proof}

%-[1.POMDPからMDPに等価であることを示す]
By Lemma~\ref{lemma:pomdp2augmdp}, the value function of the POMDP can be mapped into that of the derived MDP with histories as states, $\langle \mathcal{H}, \mathcal{A}, \gamma, \bar{r}, p_{\text{hi}} \rangle$, where \\ $p_{\text{hi}} = p(h_{t+1} | a_{t}, h_t) = \sum_{s_t} \sum_{s_{t+1}} p(s_t | h_t) p(s_{t+1} | s_t, a_t) p(o_{t+1} |s_{t+1} , a_t)$ and $\bar{r}(h_t, a_t) = \sum_{s_t} p(s_t | h_t) r(s_t, a_t)$. 

%-[2.SectionでのMeta-RL settingはPOMDPの特殊ケースであることを示す]
By considering that the hidden state is defined as $\mathcal{S}=\mathcal{T} \times \mathcal{O}$ in the meta-RL setting in Section~\ref{sec:formulatingMBRL}, 
$p(h_{t+1} | a_{t}, h_t)$ and $\bar{r}(h_t, a_t)$ can be transformed as: 
\begin{eqnarray}
p(h_{t+1} | a_{t}, h_t) &=& \sum_{s_t} \sum_{s_{t+1}} p(s_t | h_t) p(s_{t+1} | s_t, a_t) p(o_{t+1} |s_{t+1} , a_t) \nonumber\\
&=& \sum_{s_t} p(s_t | h_t) p(o_{t+1} |s_{t} , a_t) \nonumber\\
&=& \sum_{\tau_t, o_{t}}p(\tau_t, o_{t} | h_t) p(o_{t+1} | \tau_t, o_{t}, a_t) \nonumber\\
&=& \sum_{\tau_t} p(\tau_t | h_t) p(o_{t+1} | \tau_t, o_t, a_t) 
\end{eqnarray}
\begin{eqnarray}
\bar{r}(h_t, a_t) = \sum_{s_t} p(s_t | h_t) r(s_t, a_t) = \sum_{\tau_t, o_t} p(\tau_t, o_t | h_t) r(\tau_t, o_t, a_t) = \sum_{\tau_t} p(\tau_t | h_t) r(\tau_t, o_t, a_t)
\end{eqnarray}

\end{proof}

%-[start main proof]
Now, we provide the proof of Theorems~\ref{th:1pomdp} and \ref{th:2pomdp} in Section~\ref{sec:monotonic}. 

\subsubsection*{Proof of Theorem~\ref{th:1pomdp}:}
\begin{proof}
%-[convert problem]
By Lemma~\ref{lemma:metarl2augmdp} our problem of meta-RL can be mapped into the problem in the derived MDP $\langle \mathcal{H}, \mathcal{A}, \gamma, \bar{r}, p_{\text{hi}} \rangle$. 

%-[conlusion]
By applying Theorem 4.1 in \citet{janner2019trust} to the derived MDP $\langle \mathcal{H}, \mathcal{A}, \gamma, \bar{r}, p_{\text{hi}} \rangle$ with defining $\epsilon_m$, and $\epsilon_\pi$ by Definitions~\ref{def:epsm} and \ref{def:epspi} respectively, 
we obtain the following inequality in the derived MDP: 
\begin{dmath}
\mathbb{E}_{a \sim \pi, s \sim p} \left[ R \right] \geq \mathbb{E}_{(a, s) \sim m_{f}(\pi, p_\theta, \mathcal{D}_{\text{env}})} \left[ R \right] - 2 r_{\text{max}} \left\{ \frac{\gamma}{(1-\gamma)^2} ( \epsilon_m + 2\epsilon_\pi) + \frac{2}{(1-\gamma)} \epsilon_m \right\}, 
\end{dmath}

\end{proof}

\subsubsection*{Proof of Theorem~\ref{th:2pomdp}:}
\begin{proof}
%-[convert problem]
By Lemma~\ref{lemma:metarl2augmdp} our problem of meta-RL can be mapped into that in the derived MDP $\langle \mathcal{H}, \mathcal{A}, \gamma, \bar{r}, p_{\text{hi}} \rangle$. 

%-[conlusion]
By applying Theorem~\ref{th:2mdp} in Appendix~\ref{sec:boundinmdp} to the derived MDP $\langle \mathcal{H}, \mathcal{A}, \gamma, \bar{r}, p_{\text{hi}} \rangle$ with defining $\epsilon_m$, and $\epsilon_\pi$ by Definitions~\ref{def:epsm} and \ref{def:epspi} respectively, 
we obtain the following inequality in the derived MDP: 
\begin{dmath}
\mathbb{E}_{a \sim \pi, s \sim p} \left[ R \right] \geq  \mathbb{E}_{(a, h) \sim m_{b}(\pi, p_\theta, \mathcal{D}_{\text{env}})} \left[ R \right] - r_{\text{max}} \left\{ \frac{1+\gamma^2}{(1-\gamma)^2} 2 \epsilon_\pi + \frac{\gamma -k\gamma^{k} + (k-1) \gamma^{k+1}}{(1-\gamma)^2} \left(\epsilon_\pi + \epsilon_m \right) \\+ \frac{\gamma^{k} - \gamma}{\gamma - 1} (\epsilon_{\pi} + \epsilon_m) + \frac{\gamma^k}{1-\gamma} (k+1) (\epsilon_{\pi} + \epsilon_m) \right\}. 
\end{dmath}
\end{proof}
%
%deprpe Similarly, Theorem~\ref{th:1pomdp} is con be proven by mapping our meta-RL setting into the problem in the derived MDP by Lemma~\ref{lemma:metarl2augmdp} and leveraging theoretical results on the MDP. 

\subsection{The discrepancy factors relying on the model error $\epsilon_m$ in Theorem~\ref{th:1pomdp} and those in Theorem~\ref{th:2pomdp}}\label{sec:disc1anddisc2}
\subsubsection*{Proof of Corollary~\ref{cor:disc1anddisc2}:}
\begin{proof}
Let the terms relying on $\epsilon_m$ in $C_{\text{{\normalfont Th}1}}$ as $C_{\text{{\normalfont Th}1}, m}$. 
Let the terms relying on $\epsilon_m$ at $k=1$ in $C_{\text{{\normalfont Th}2}}$ as $C_{\text{{\normalfont Th}2}, m}$. 

By Theorems~\ref{th:1pomdp} and \ref{th:2pomdp}, 
\begin{dmath}
C_{\text{{\normalfont Th}1}, m} = r_{\text{max}} \frac{2 \gamma \epsilon_m}{(1-\gamma)^{2}}. 
\end{dmath}
\begin{dmath}
C_{\text{{\normalfont Th}2}, m} = r_{\text{max}} \frac{\gamma}{1-\gamma} 2 \epsilon_m. 
\end{dmath}

Given that $\gamma \in [0, 1)$, $r_{\text{max}} > 0$ and $\epsilon_m \geq 0$, 
\begin{dmath}
C_{\text{{\normalfont Th}2}, m} - C_{\text{{\normalfont Th}1}, m} = r_{\text{max}} \frac{- 2\gamma^2\epsilon_m }{(1-\gamma)^{2}} \leq 0. 
\end{dmath}
\end{proof}

\subsection{Baseline methods for our experiment}\label{sec:baselines}
\textbf{PEARL:} The model-free meta-RL method proposed in \citet{rakelly2019efficient}. 
This is an off-policy method and implemented by extending Soft Actor-Critic~\citep{haarnoja2018soft}. 
By leveraging experience replay, this method shows high sample efficiency. 
We reimplemented the PEARL method on TensorFlow, referring to the original implementation on PyTorch (\url{https://github.com/katerakelly/oyster}). \\
%deppre In addition, we make a modification on PEARL implementation in order to adapt it to our meta-RL setting. 
%deppre In the meta-RL setting in \citet{rakelly2019efficient}, the task is assumed to be invariant during an episode. 
%deppre However, in our meta-RL setting, the task changes during an episode. 
%deppre To adapt the PEARL to our meta-RL setting, we replace the context encoder and context-conditioned policy with the policy for history-action space that we implemented in Section~\ref{sec:practicalMBPO}. 
%deppre The resulting implementation is the same as that of M3PO (Algorithm~\ref{alg1:Meta-MBPO} ) except that the model-based rollout is not used and that the real trajectory is used for a policy optimization.  \\
%
\textbf{Learning to adapt (L2A):} The model-based meta-RL proposed in \citet{nagabandilearning}. 
In this method, the model is implemented with MAML~\citep{finn2017model} and the optimal action is found by the model predictive path integral control~\citep{williams2015model}. 
We adapt the following implementation of L2A to our experiment: \url{https://github.com/iclavera/learning_to_adapt} 

\subsection{Environments for our experiments}\label{sec:environment_desc}
For our experiments in Section \ref{sec:experiments}, we prepare simulated robot environments using the MuJoCo physics engine~\citep{todorov2012mujoco}: \\
\textbf{Halfcheetah-fwd-bwd:} In this environment, policies are used to control the half-cheetah, which is a planar biped robot with eight rigid links, including two legs and a torso, along with six actuated joints. Here, the half-cheetah's moving direction is randomly selected from ``forward'' and ``backward'' around every 15 seconds (in simulation time).
If the half-cheetah moves in the correct direction, a positive reward is fed to the half-cheetah in accordance with the magnitude of movement, otherwise, a negative reward is fed. \\
\textbf{Halfcheetah-pier:} In this environment, the half-cheetah runs over a series of blocks that are floating on water. Each block moves up and down when stepped on, and the changes in the dynamics are rapidly changing due to each block having different damping and friction properties. These properties are randomly determined at the beginning of each episode. \\
\textbf{Ant-fwd-bwd:} Same as Halfcheetah-fwd-bwd except that the policies are used for controlling the ant, which is a quadruped robot with nine rigid links, including four legs and a torso, along with eight actuated joints. \\
\textbf{Ant-crippled-leg:} In this environment, we randomly sample a leg on the ant to cripple. 
The crippling of the leg causes unexpected and drastic changes to the underlying dynamics. 
One of the four legs is randomly crippled every 15 seconds. \\
\textbf{Walker2D-randomparams:} In this environment, the policies are used to control the walker, which is a planar biped robot consisting of seven links, including two legs and a torso, along with six actuated joints. The walker's torso mass and ground friction is randomly determined every 15 seconds. \\
\textbf{Humanoid-direc:} In this environment, the policies are used to control the humanoid, which is a biped robot with 13 rigid links, including two legs, two arms and a torso, along with 17 actuated joints. 
In this task, the humanoid moving direction is randomly selected from two different directions around every 15 seconds. 
If the humanoid moves in the correct direction, a positive reward is fed to the humanoid in accordance with the magnitude of its movement, otherwise, a negative reward is fed.

\clearpage
\subsection{Complementary experimental results}\label{sec:comp_analysis}
\begin{figure}[h!]
\begin{center}
\begin{tabular}{c}
\begin{minipage}{1.0\hsize}
      \includegraphics[clip, width=0.329\hsize]{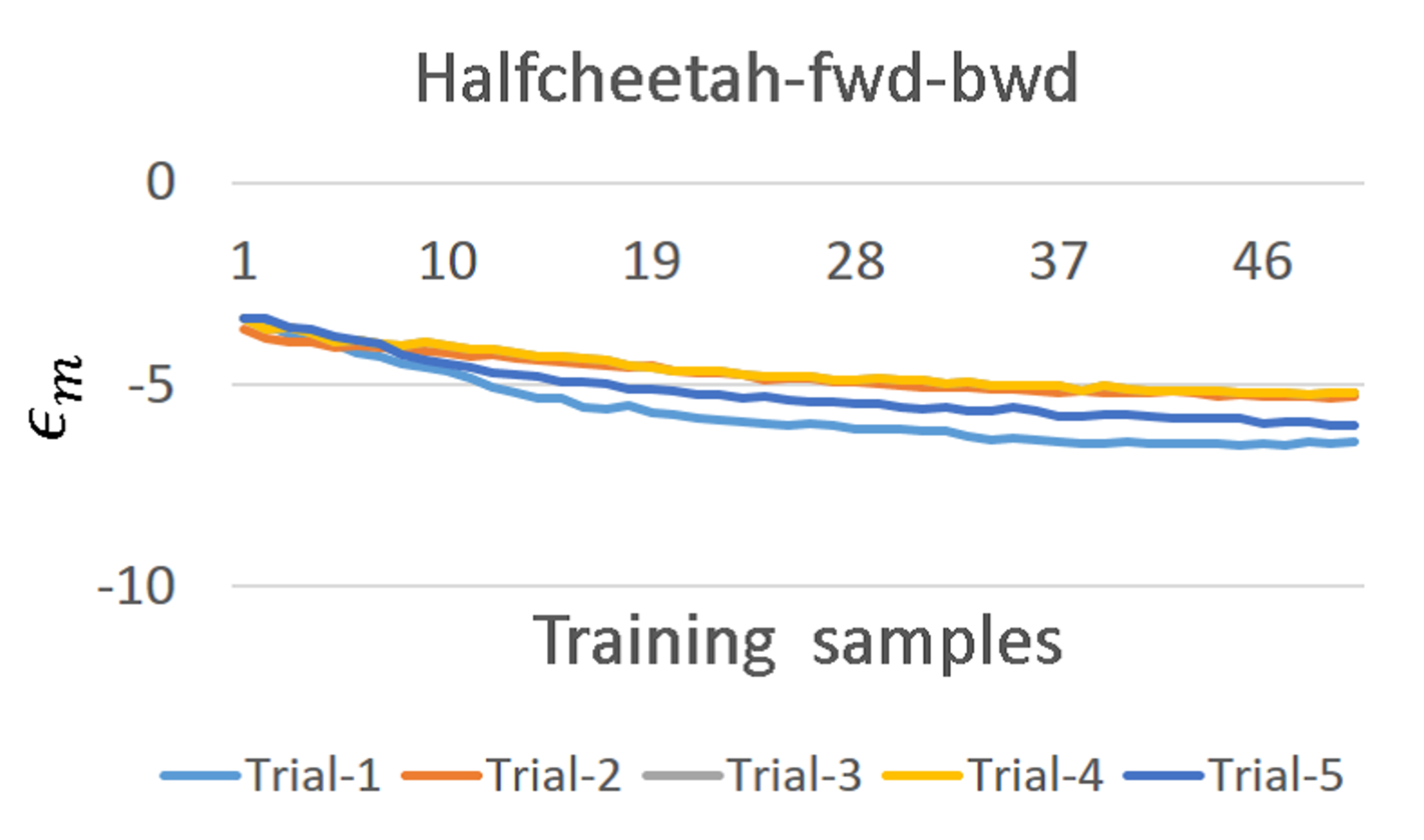}\vline
      \includegraphics[clip, width=0.329\hsize]{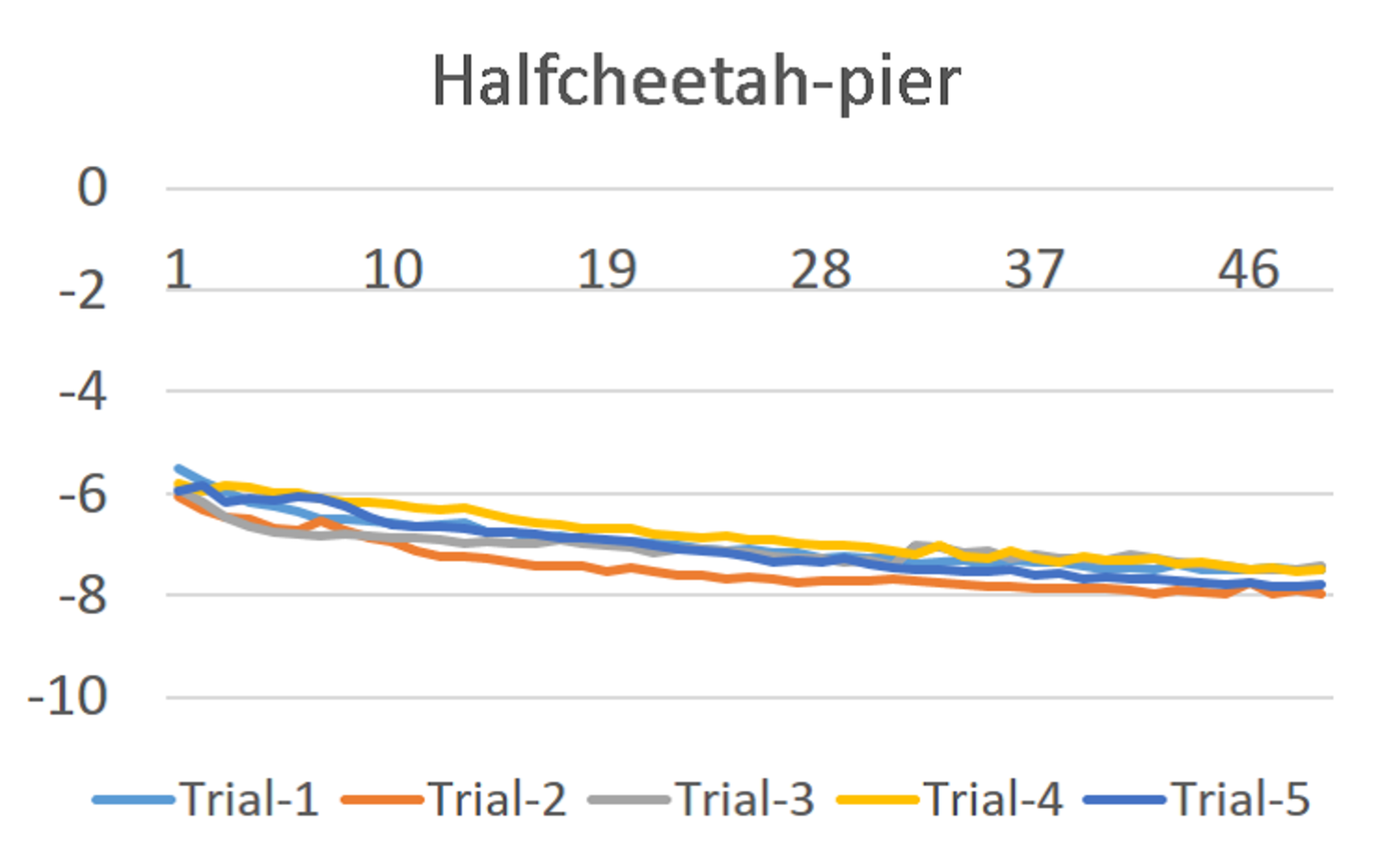}\vline
      \includegraphics[clip, width=0.329\hsize]{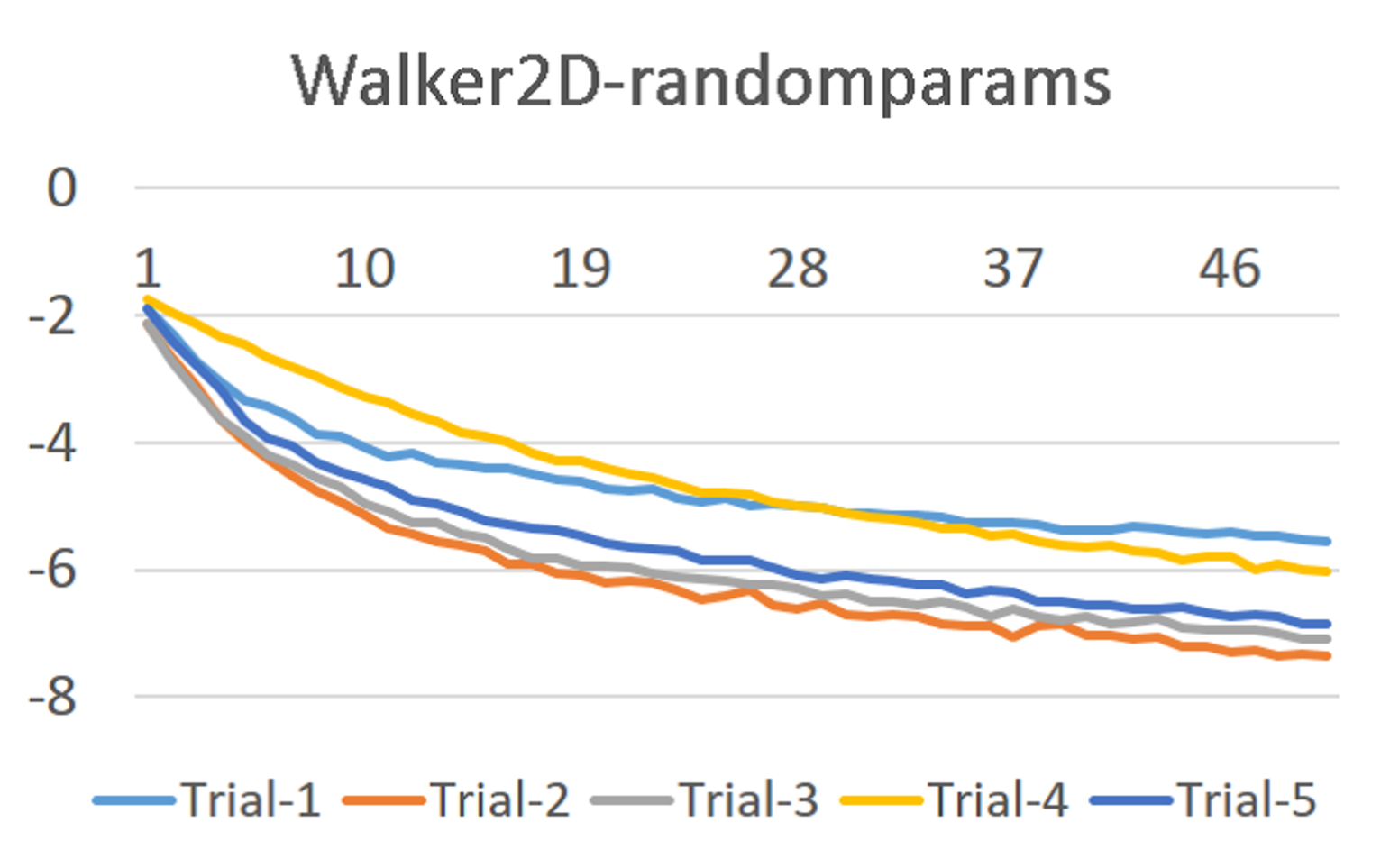}
\end{minipage}\\\hline
\begin{minipage}{1.0\hsize}
      \includegraphics[clip, width=0.329\hsize]{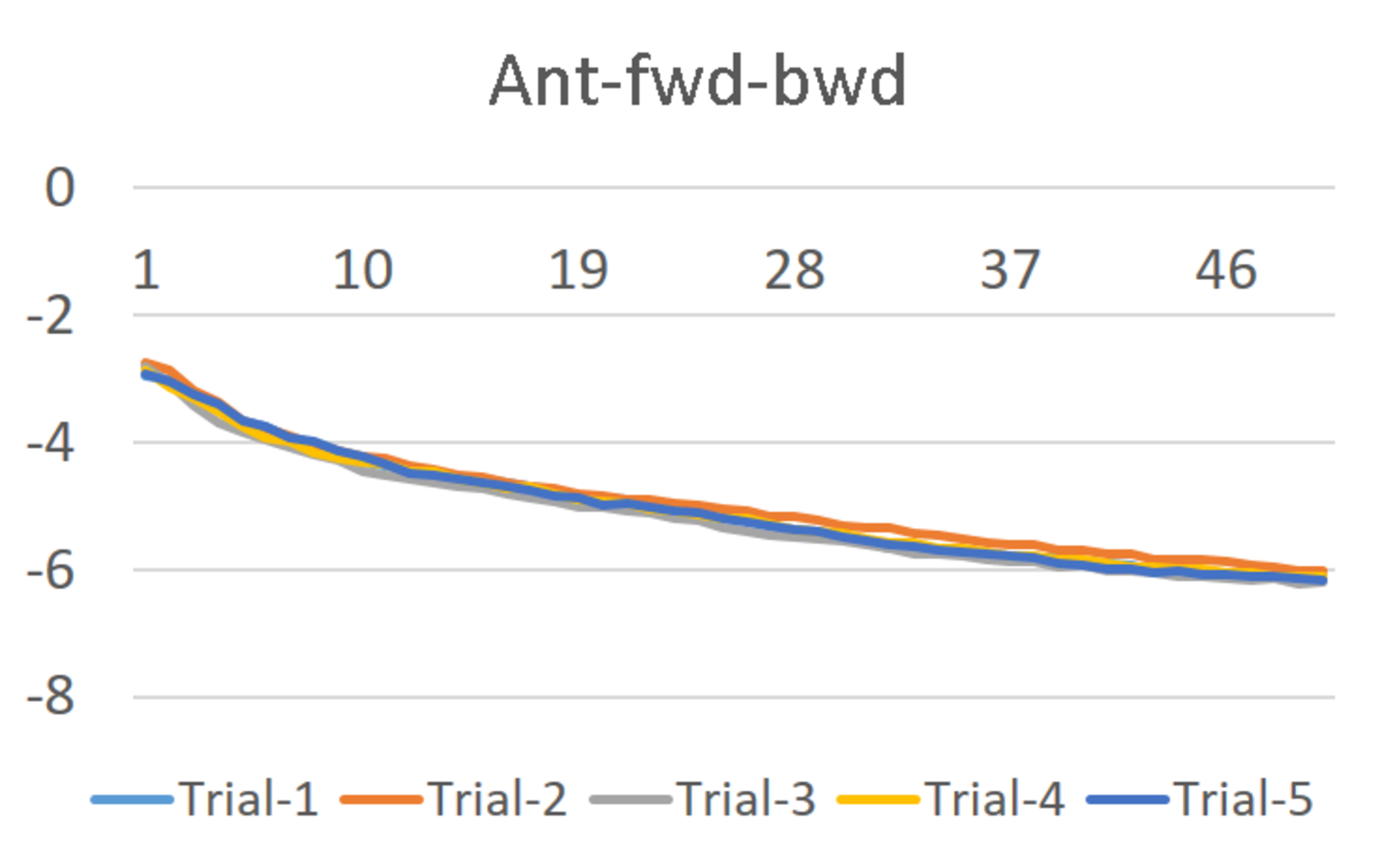}\vline
      \includegraphics[clip, width=0.329\hsize]{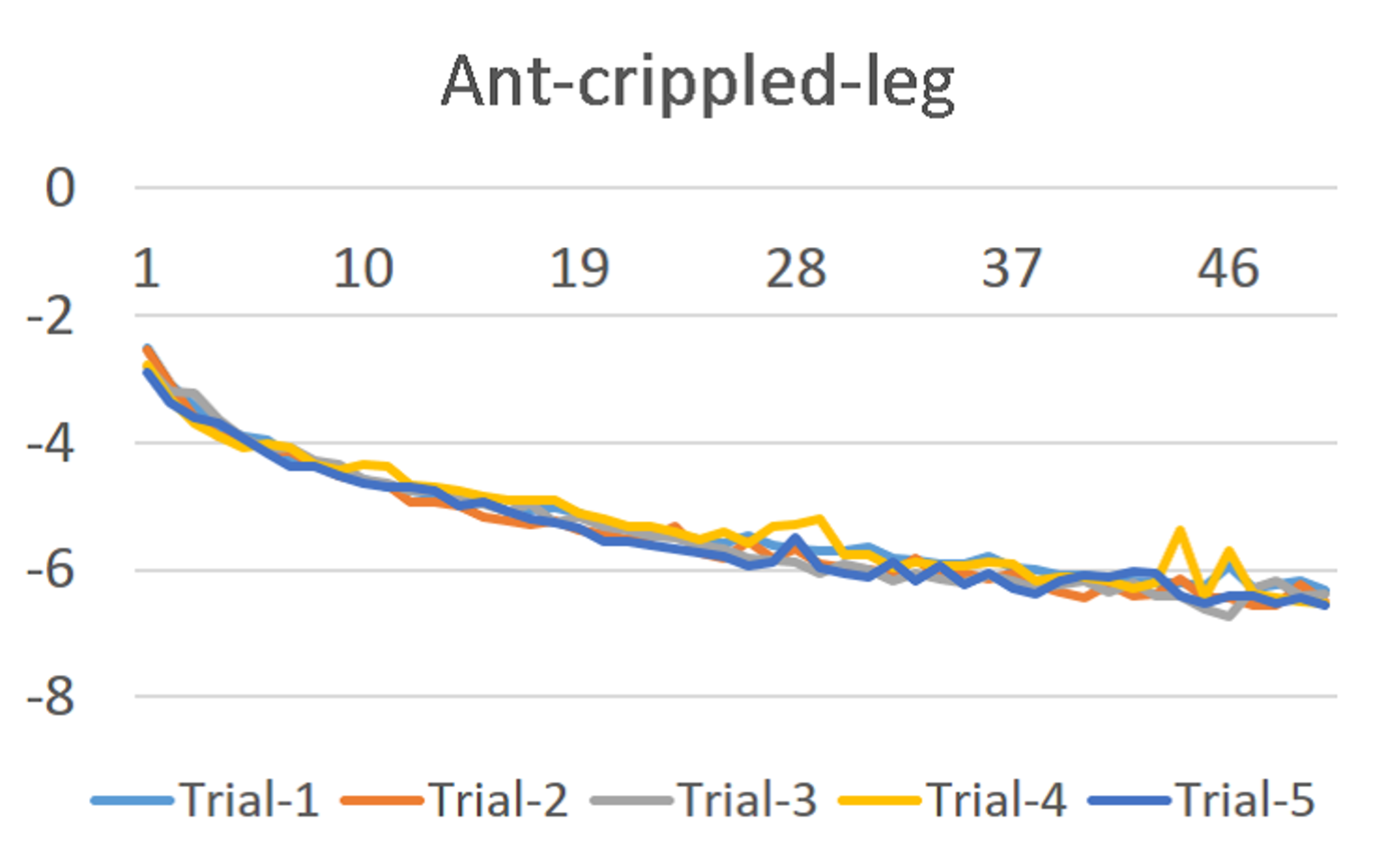}\vline
      \includegraphics[clip, width=0.329\hsize]{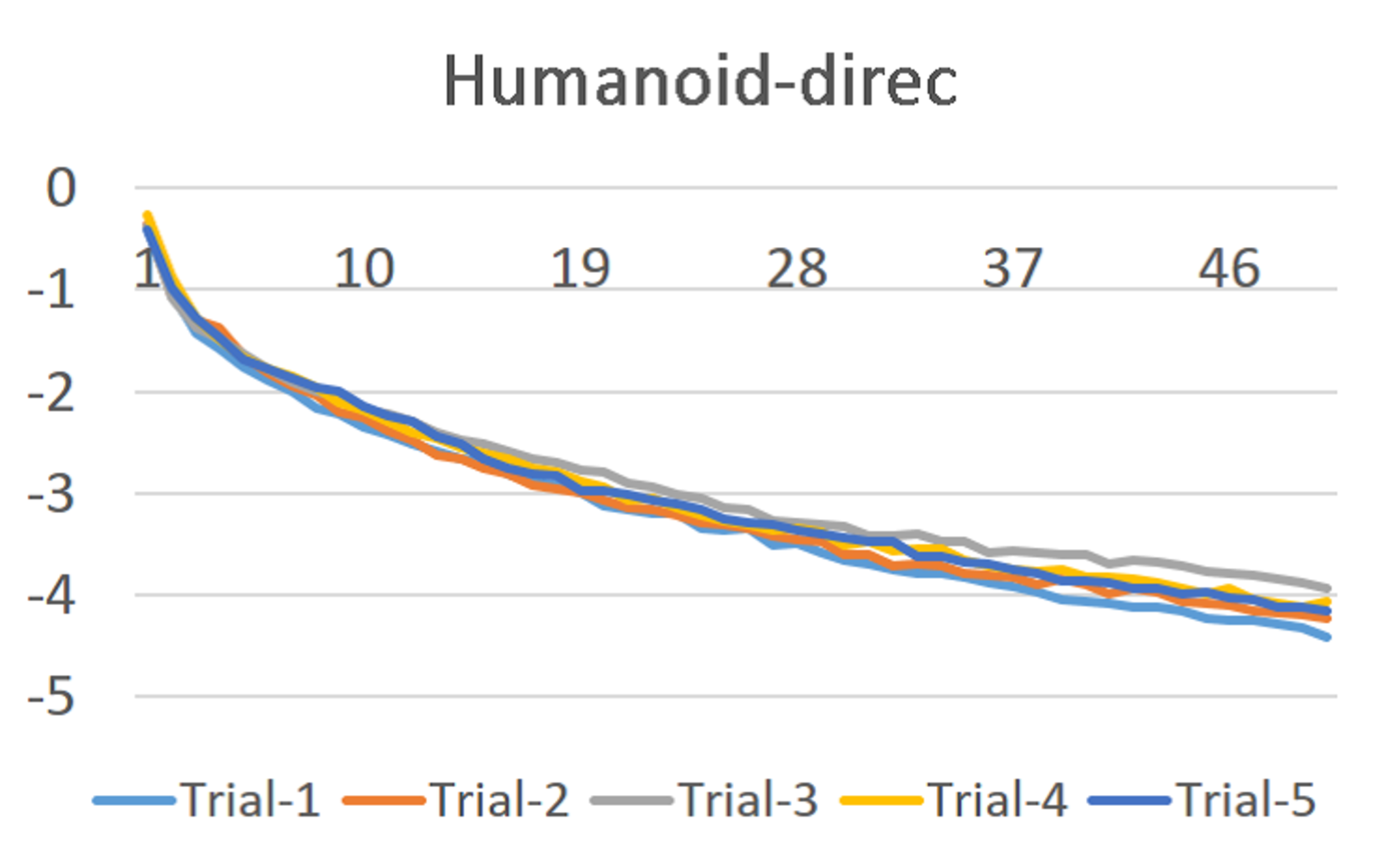}
\end{minipage}
\end{tabular}
\end{center}
\caption{Transition of model errors on training. In each figure, the vertical axis represents empirical values of $\epsilon_m$ and the horizontal axis represents the number of training samples (x1000). We ran five trials with different random seeds. The result of the $x$-th trial is denoted by Trial-$x$. We used the negative of log-likelihood of the model on validation samples as the approximation of $\epsilon_m$. The figures show that the model error tends to decrease as epochs elapse.}
\label{fig:modelerror}
\end{figure}
\begin{figure}[h!]
\begin{center}
\begin{tabular}{c}
\begin{minipage}{1.0\hsize}
      \includegraphics[clip, width=0.329\hsize]{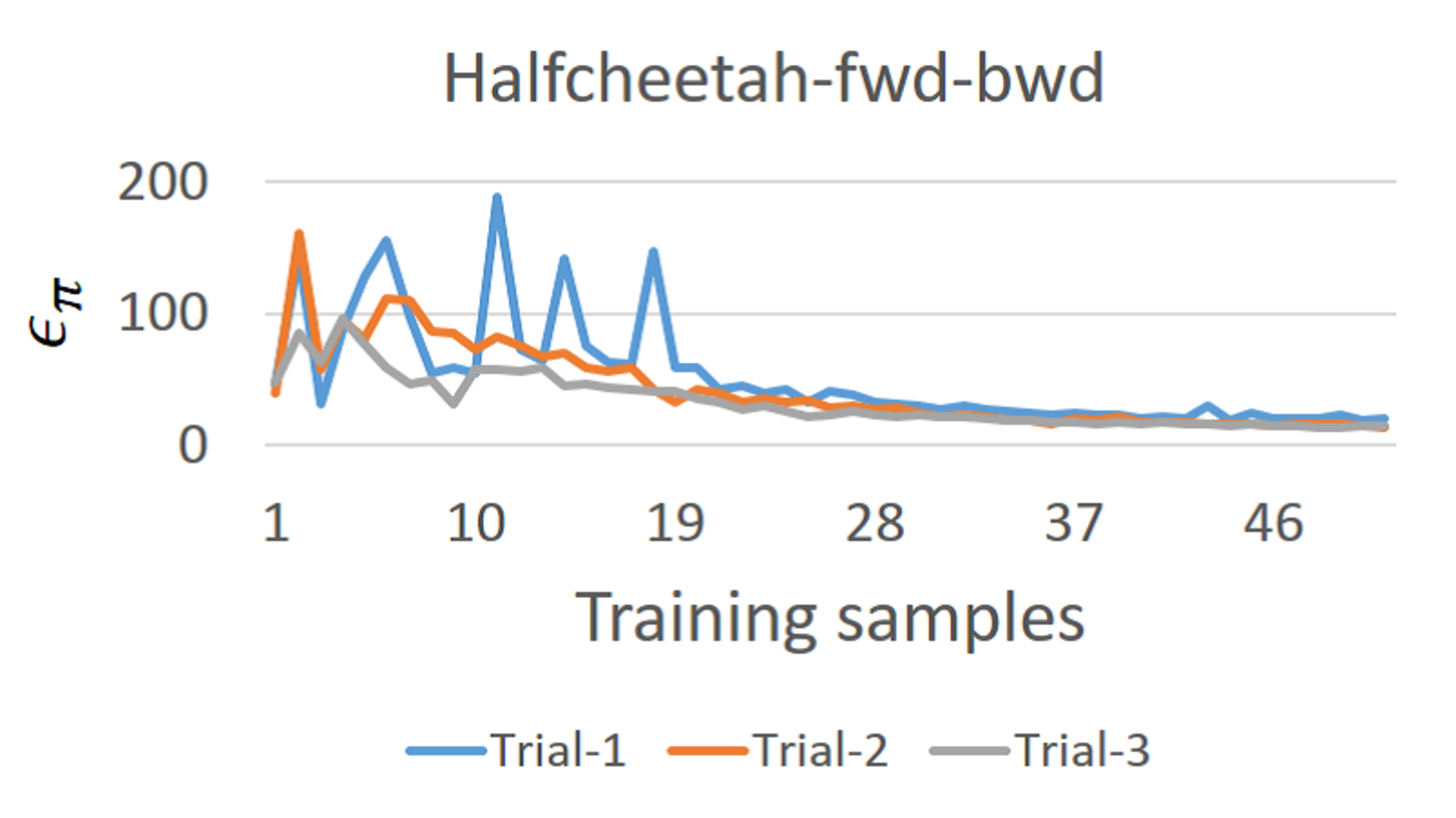}\vline
      \includegraphics[clip, width=0.329\hsize]{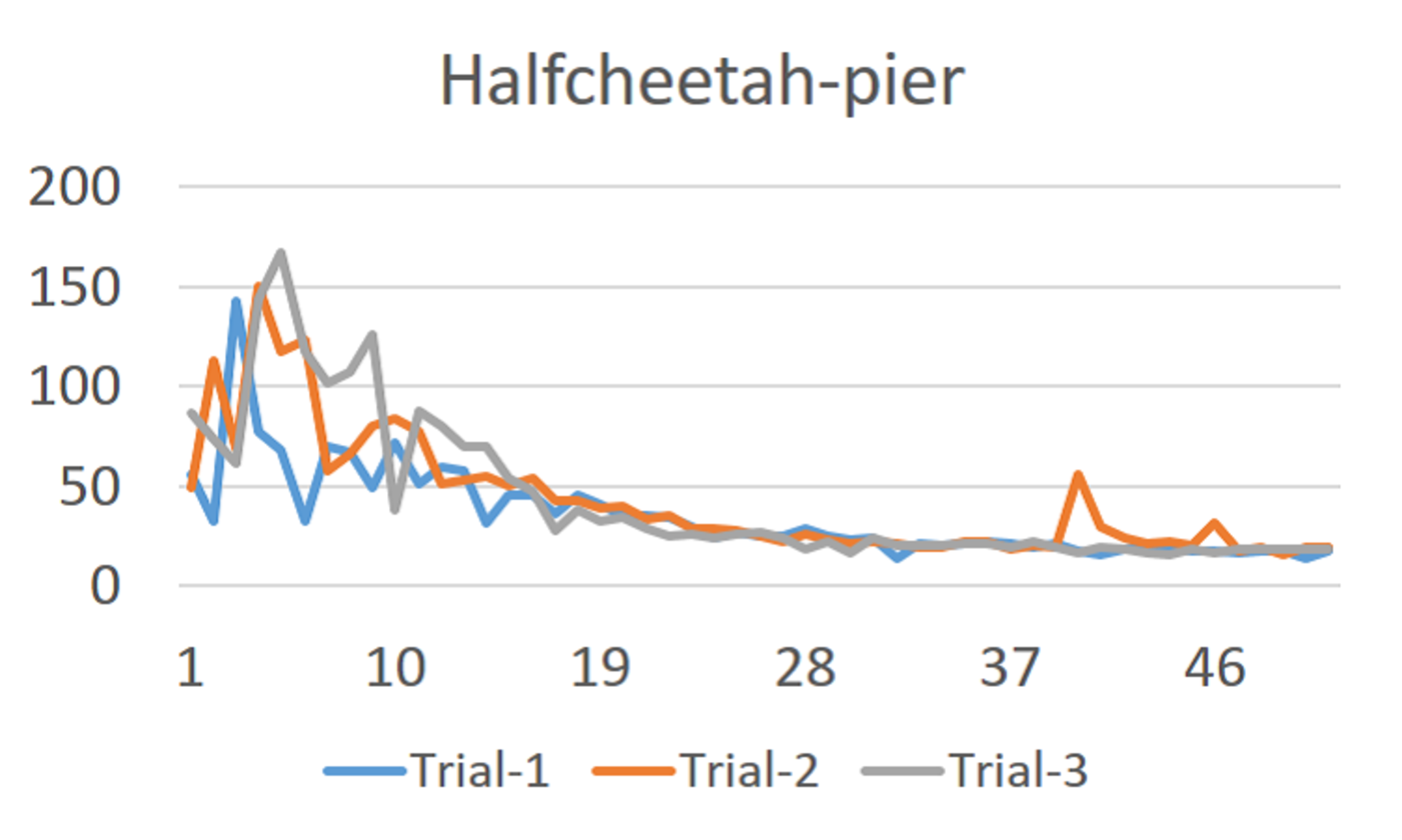}\vline
      \includegraphics[clip, width=0.329\hsize]{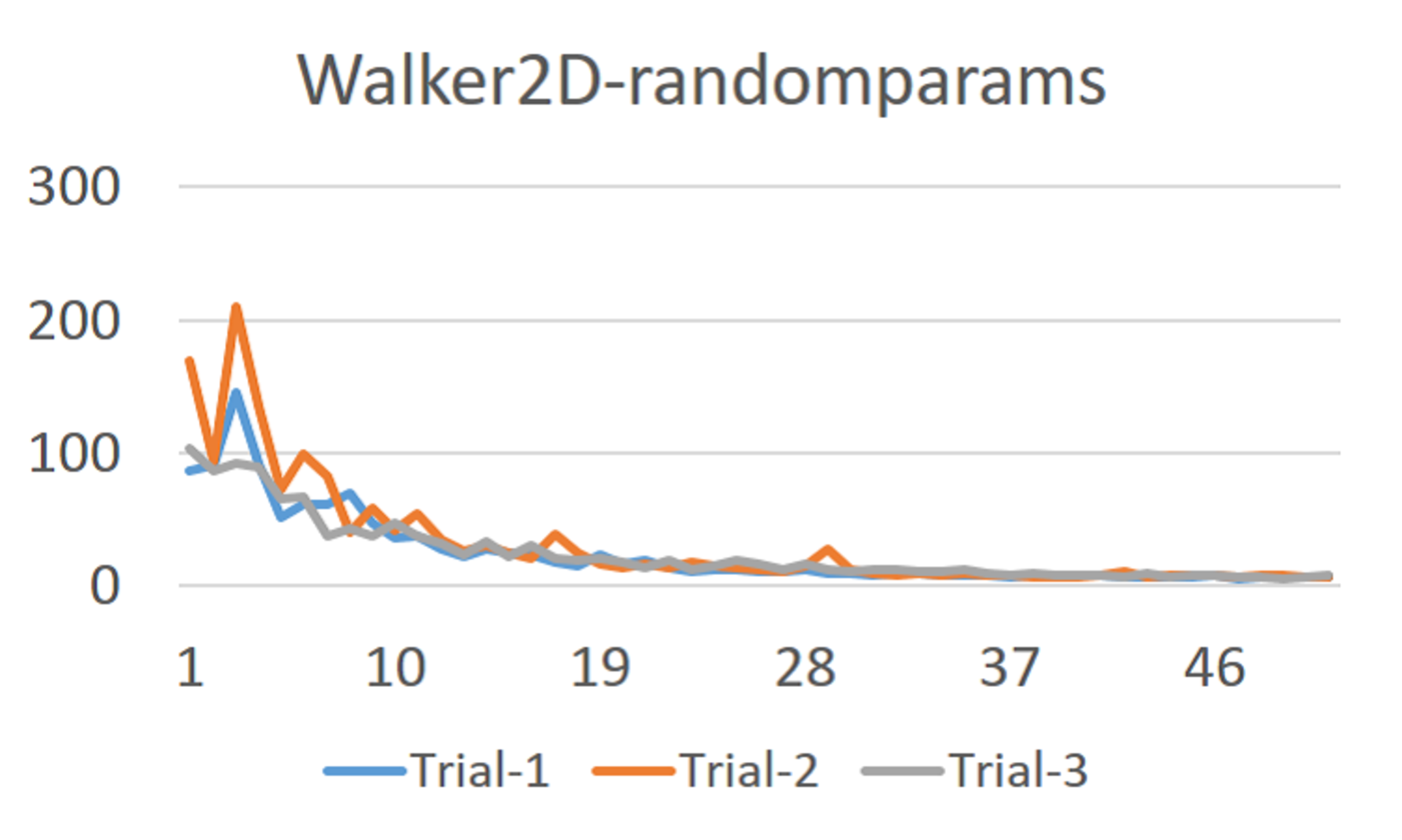}
\end{minipage}\\\hline
\begin{minipage}{1.0\hsize}
      \includegraphics[clip, width=0.329\hsize]{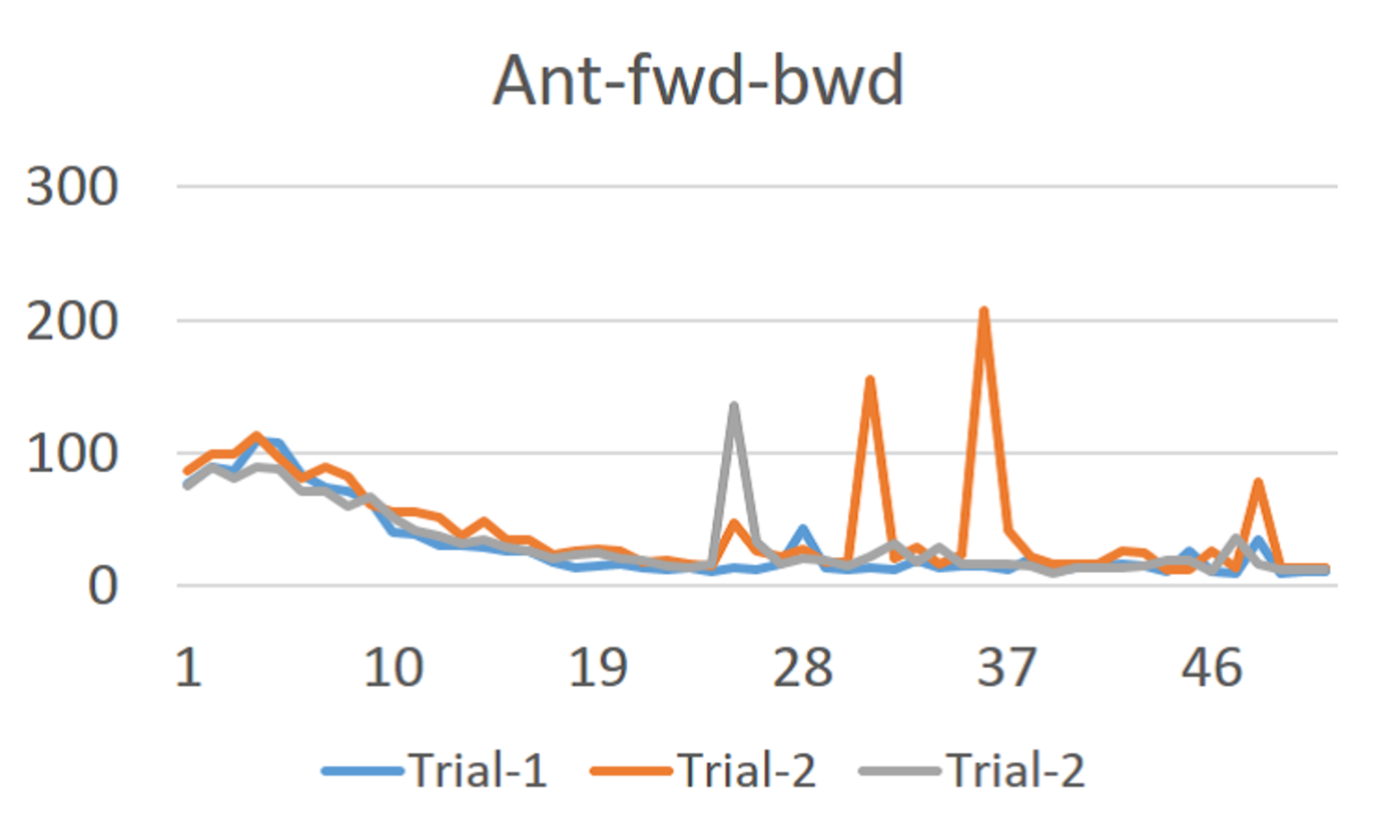}\vline
      \includegraphics[clip, width=0.329\hsize]{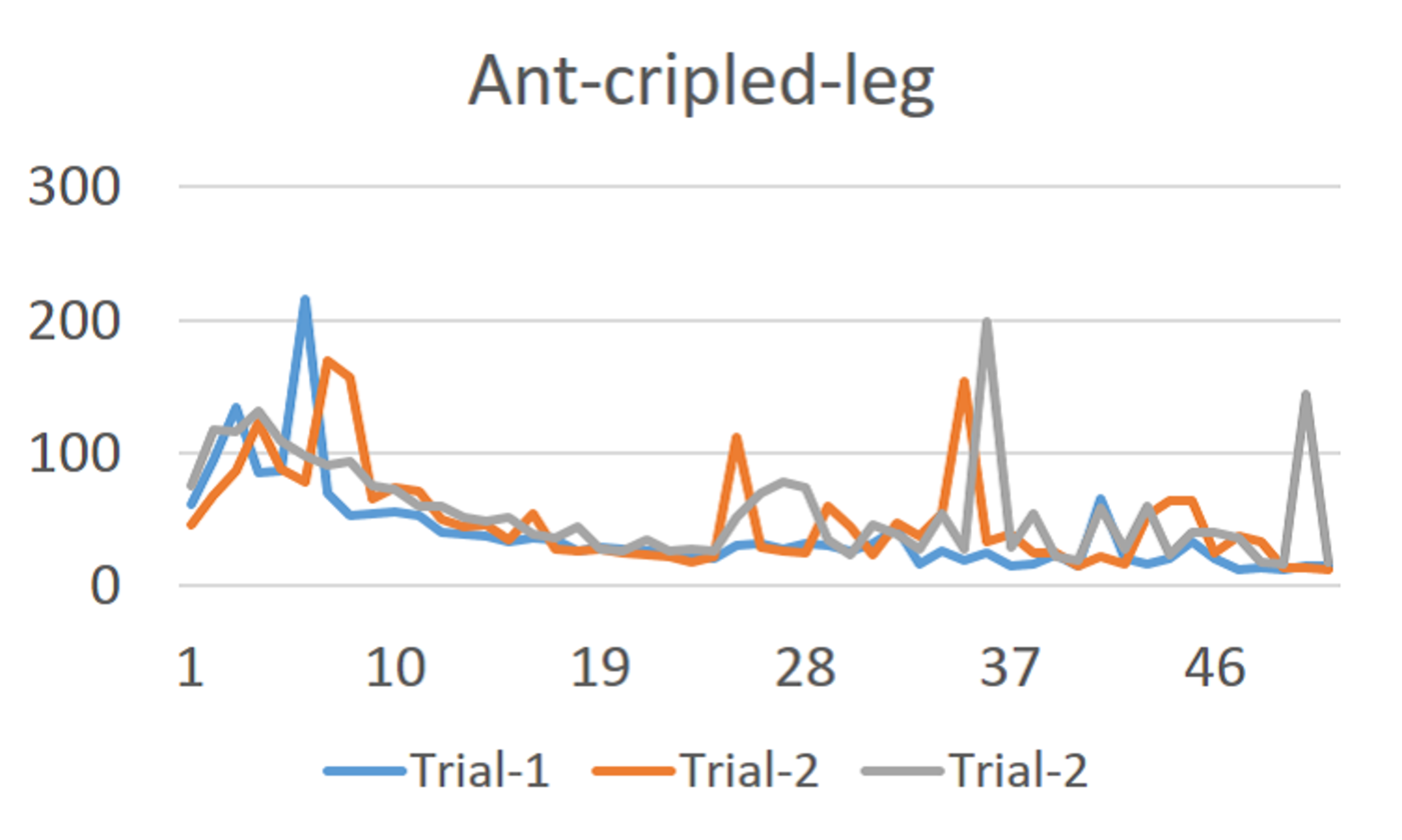}\vline
      \includegraphics[clip, width=0.329\hsize]{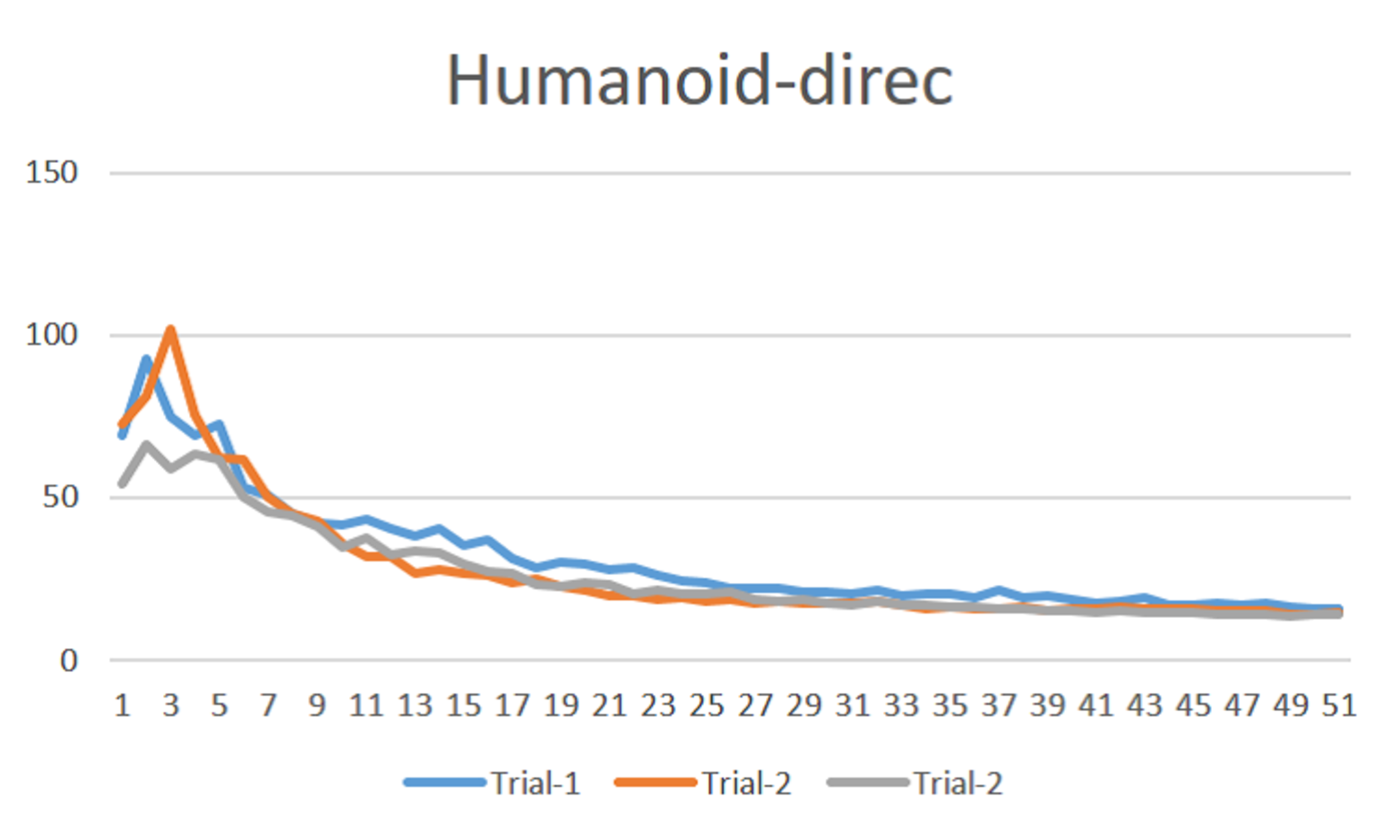}
\end{minipage}
\end{tabular}
\end{center}
\caption{Transition of policy divergence on training. In each figure, the vertical axis represents empirical values of $\epsilon_\pi$ and the horizontal axis represents the number of training samples (x1000). We ran three trials with different random seeds. The result of the $x$-th trial is denoted by Trial-$x$. For $\epsilon_\pi$, we used the empirical Kullback-Leibler divergence of $\pi_\theta$ and $\pi_{\mathcal{D}}$. Here, $\pi_{\mathcal{D}}$ has the same policy network architecture with $\pi_\theta$ and is learned by maximum likelihood estimation with the trajectories stored in $\mathcal{D}_{\text{env}}$. }
\label{fig:policy_shift}
\end{figure}
\begin{figure}[h!]
\begin{center}
\begin{tabular}{c}
\begin{minipage}{1.0\hsize}
      \includegraphics[clip, width=0.329\hsize]{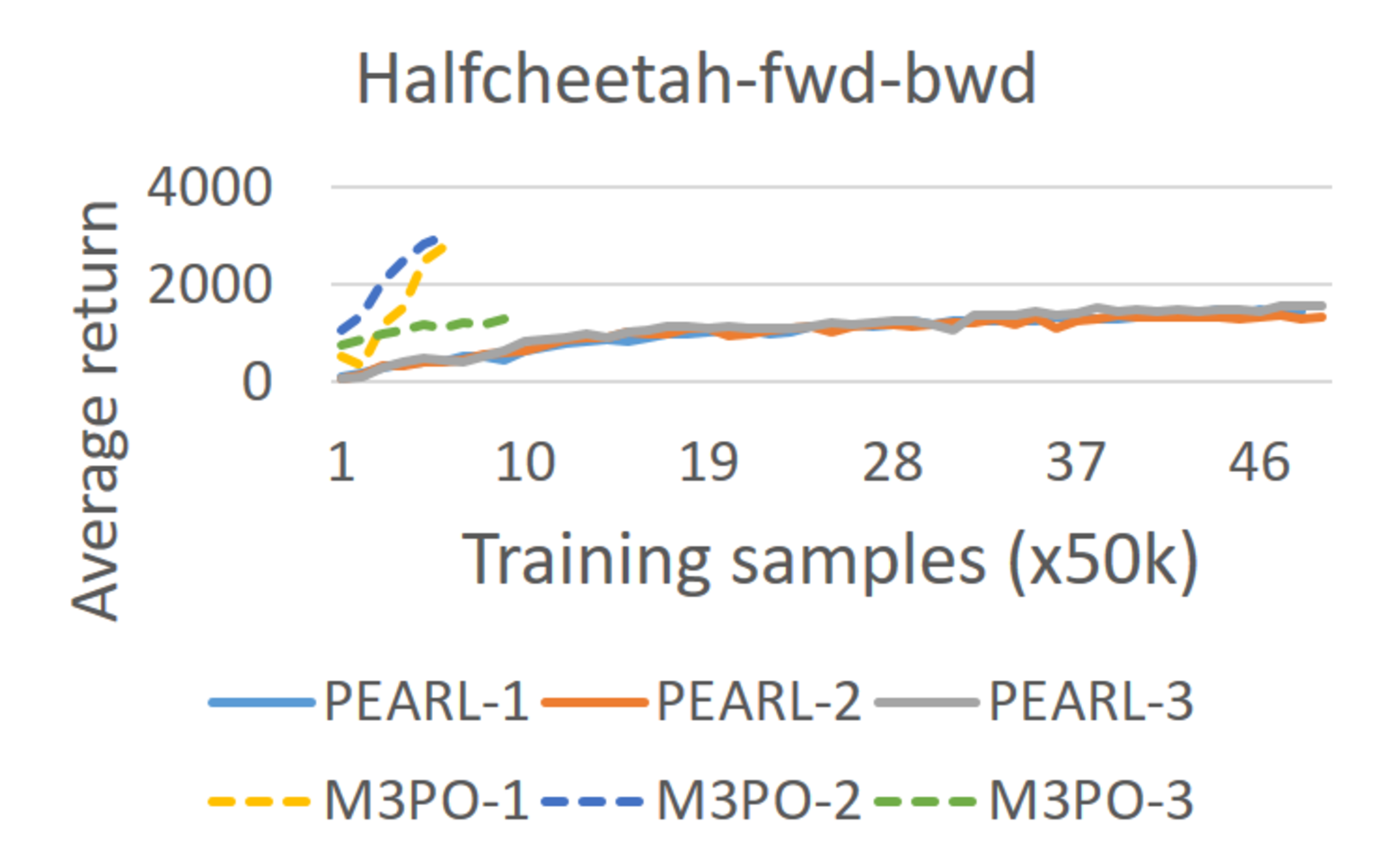}\vline
      \includegraphics[clip, width=0.329\hsize]{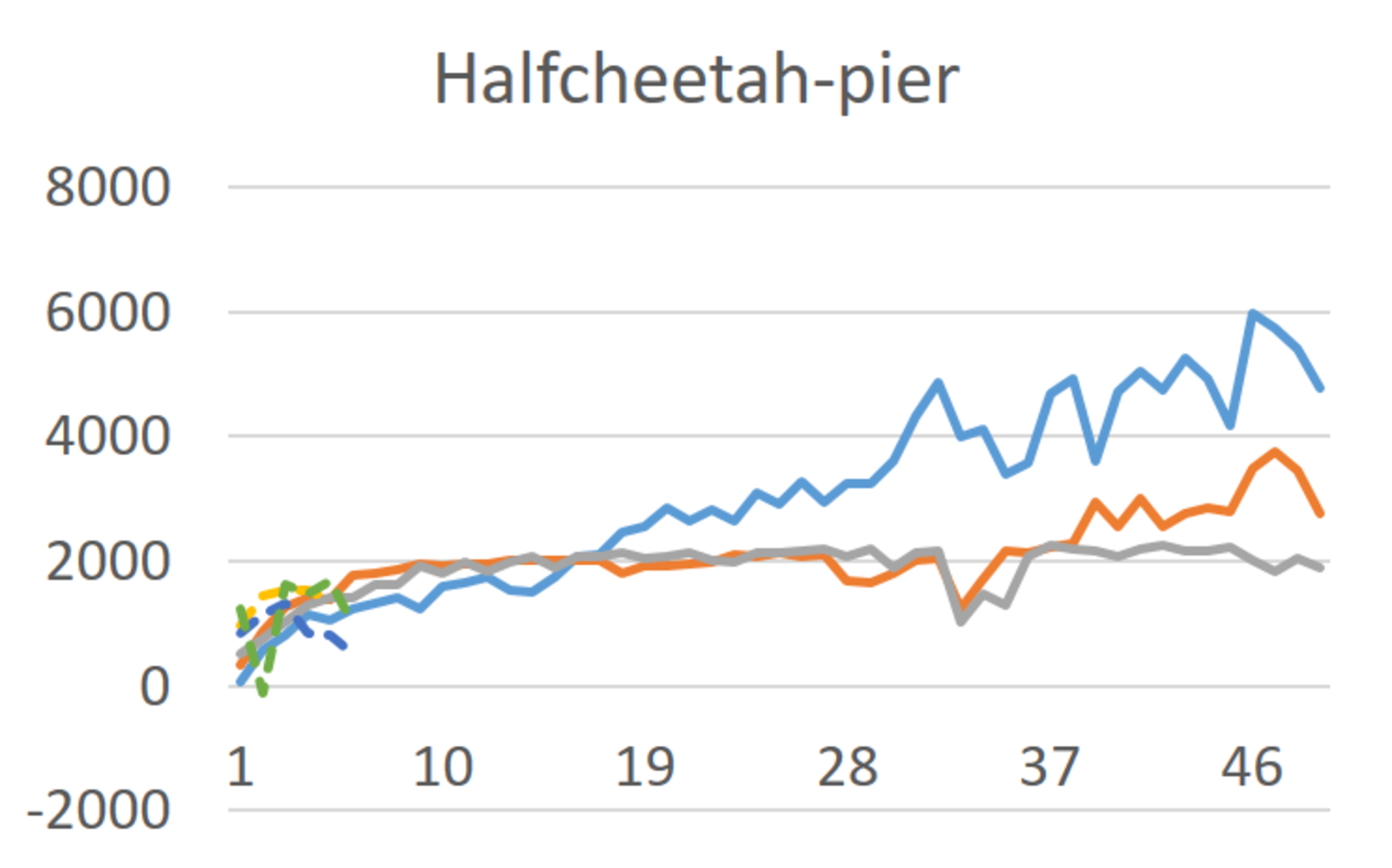}\vline
      \includegraphics[clip, width=0.329\hsize]{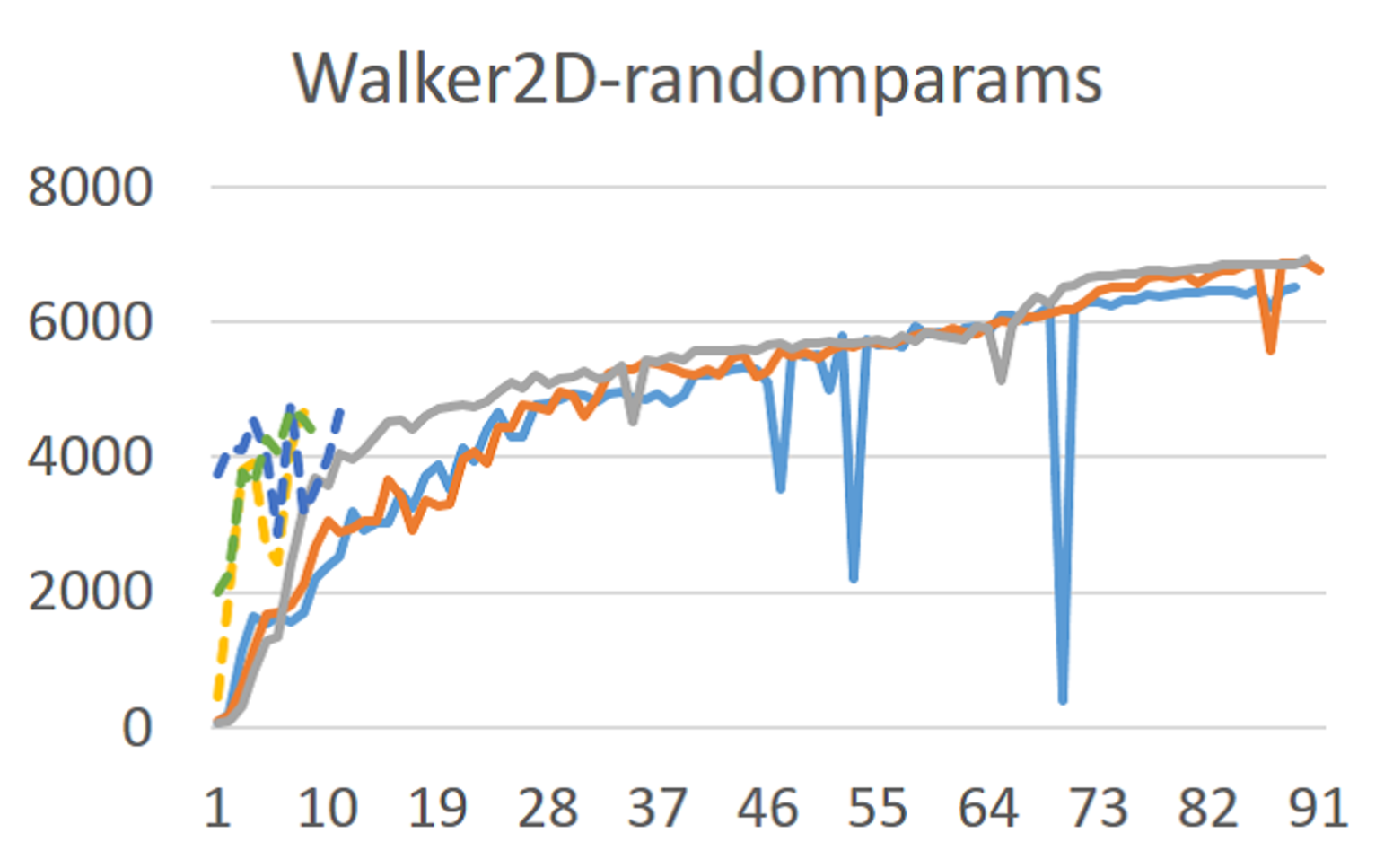}
\end{minipage}\\\hline
\begin{minipage}{1.0\hsize}
      \includegraphics[clip, width=0.329\hsize]{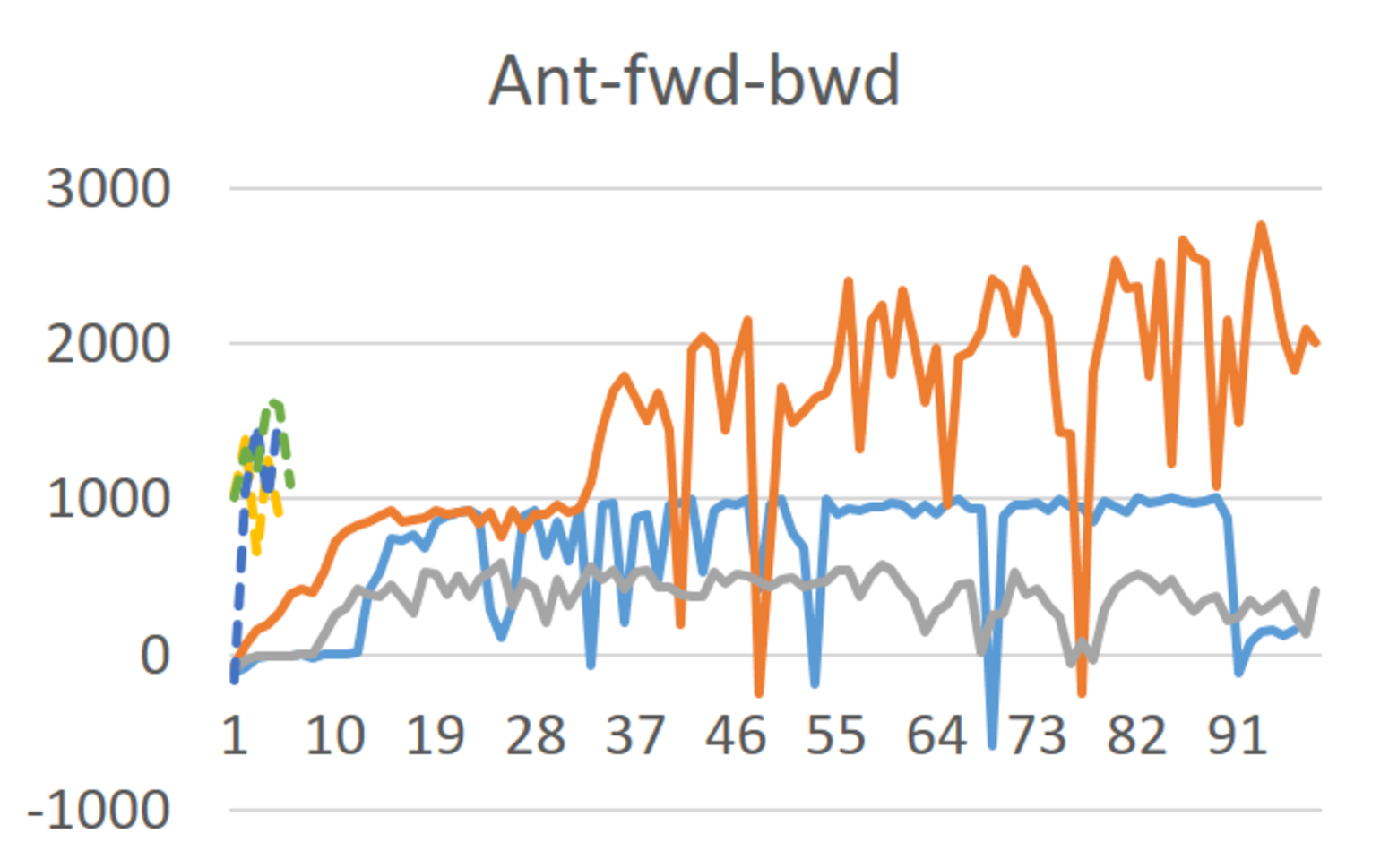}\vline
      \includegraphics[clip, width=0.329\hsize]{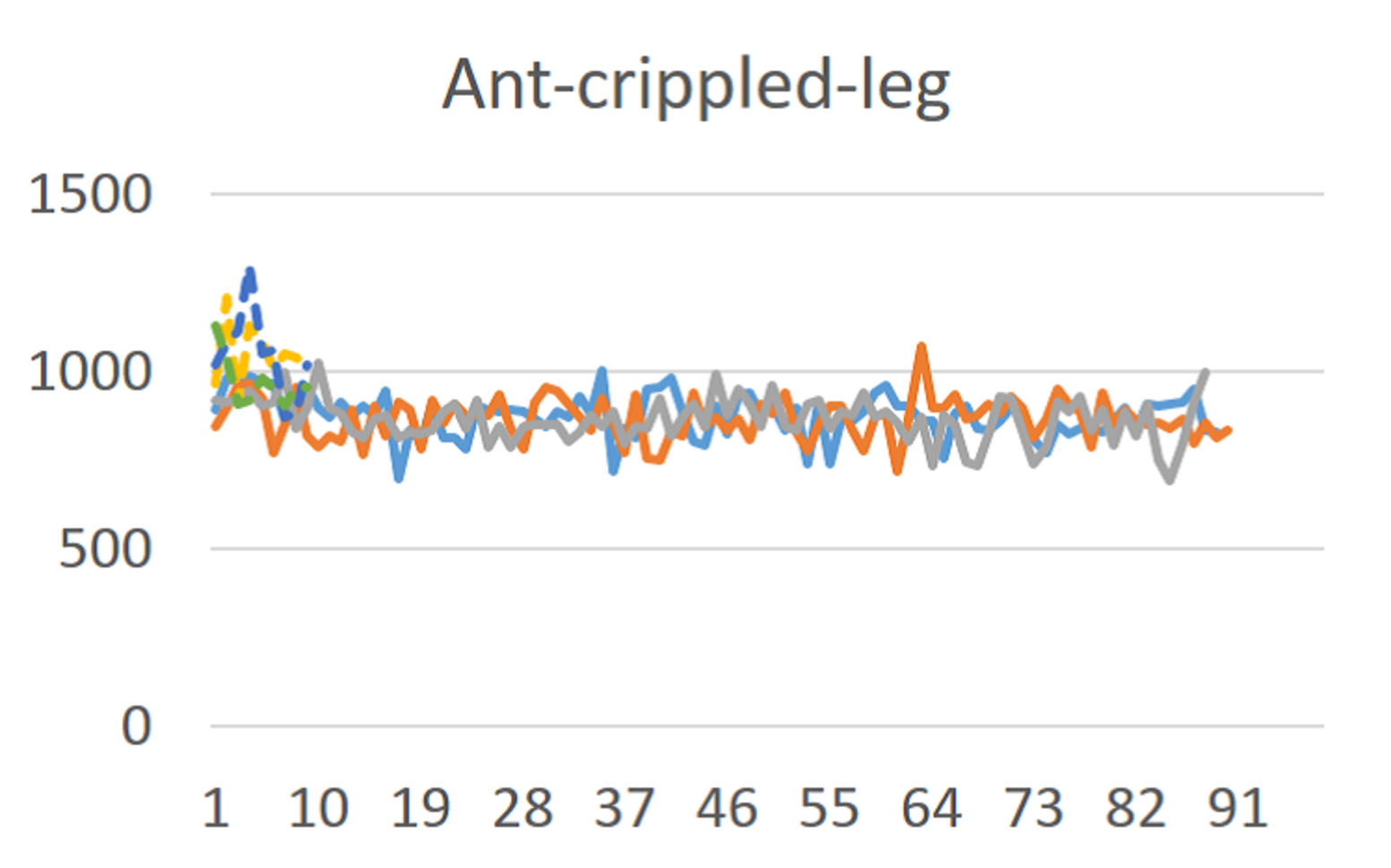}\vline
      \includegraphics[clip, width=0.329\hsize]{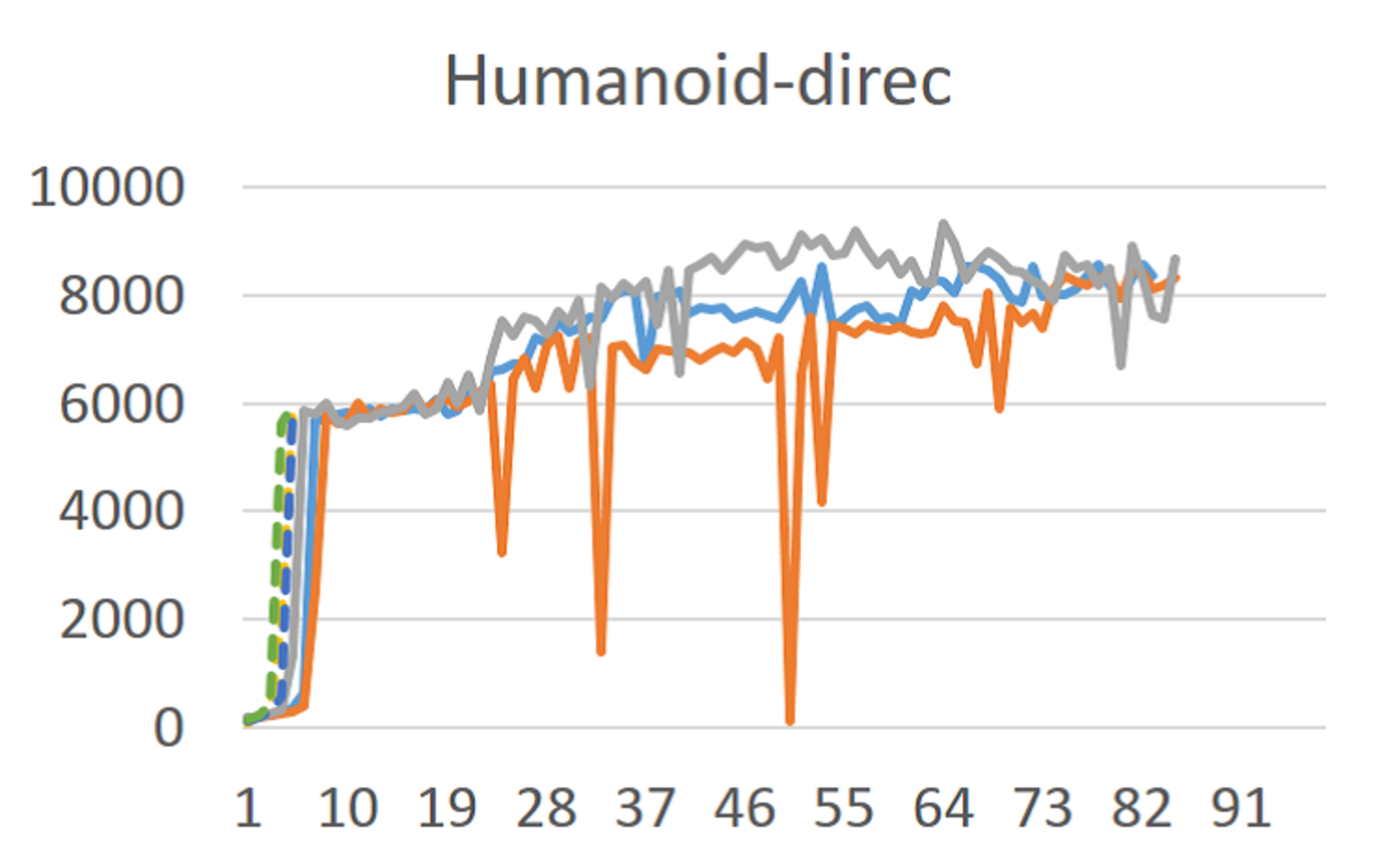}
\end{minipage}
\end{tabular}
\end{center}
\caption{Learning curve of PEARL and M3PO in a long-term training. In each figure, the vertical axis represents expected returns and the horizontal axis represents the number of training samples (\textbf{x50000}). The policy and model were fixed and their expected returns were evaluated on 50 test episodes at every 50,000 training samples. Each method was evaluated in three trials, and the result of the $x$-th trial is denoted by method-$x$. \textbf{Note that the scale of the horizontal axis is larger than that in Figure~\ref{fig:lc} by 50 times (i.e., 4 in this figure is equal to 200 in Figure~\ref{fig:lc}).}}
\label{fig:unlim}
\end{figure}
\begin{figure}[h!]
\begin{center}
\begin{tabular}{c}
\begin{minipage}{1.0\hsize}
      \includegraphics[clip, width=0.329\hsize]{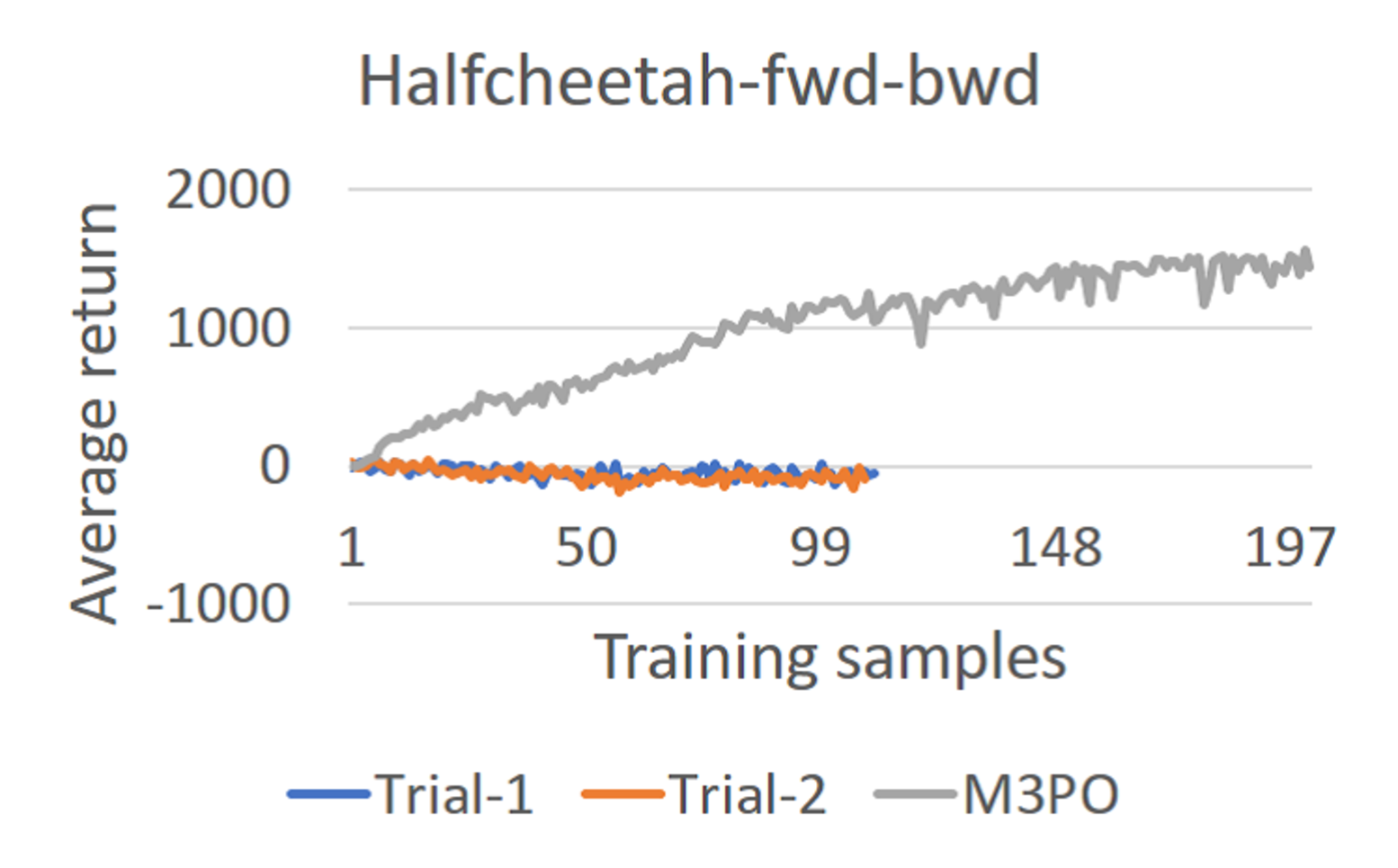}\vline
      \includegraphics[clip, width=0.329\hsize]{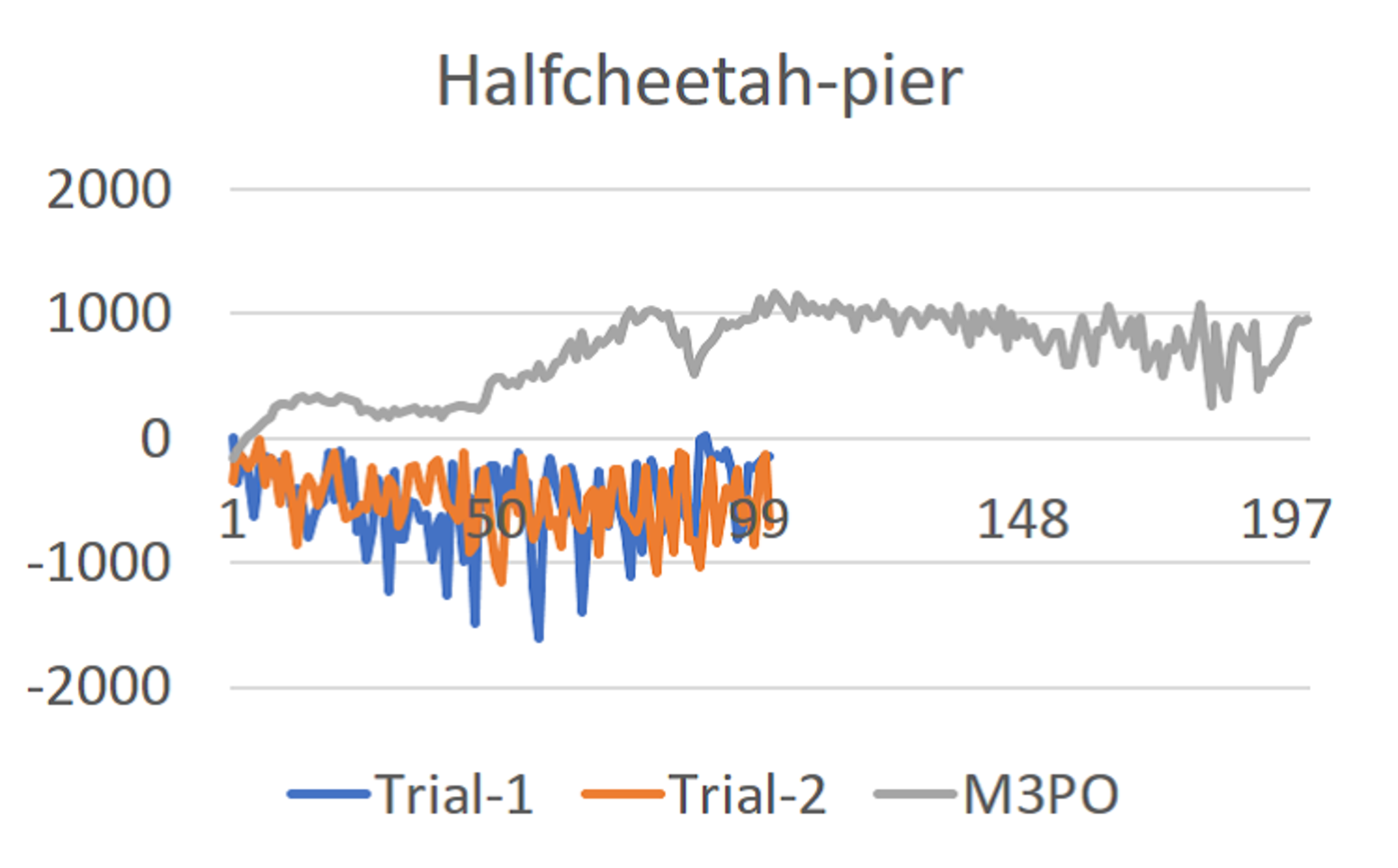}\vline
      \includegraphics[clip, width=0.329\hsize]{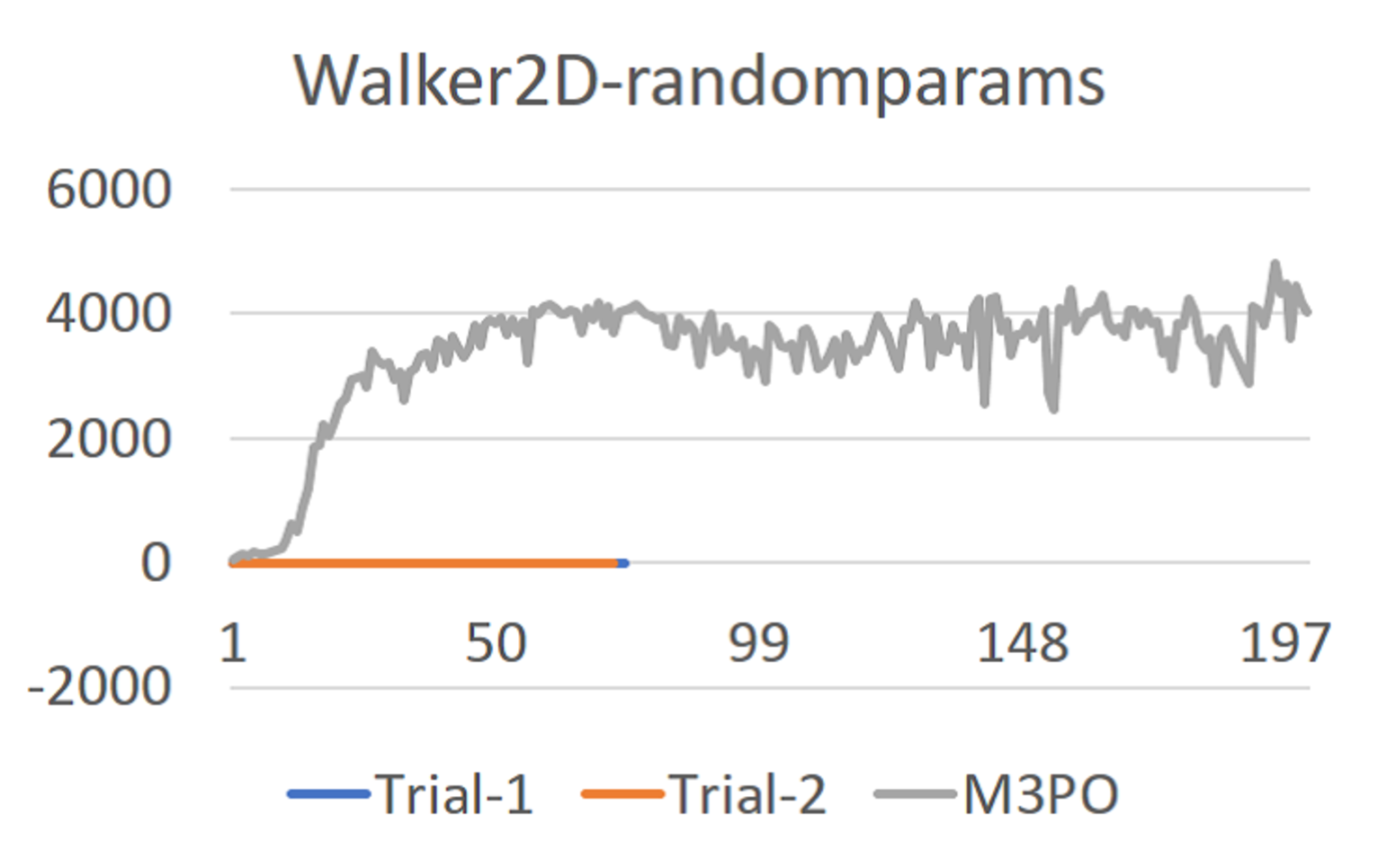}
\end{minipage}\\\hline
\begin{minipage}{1.0\hsize}
      \includegraphics[clip, width=0.329\hsize]{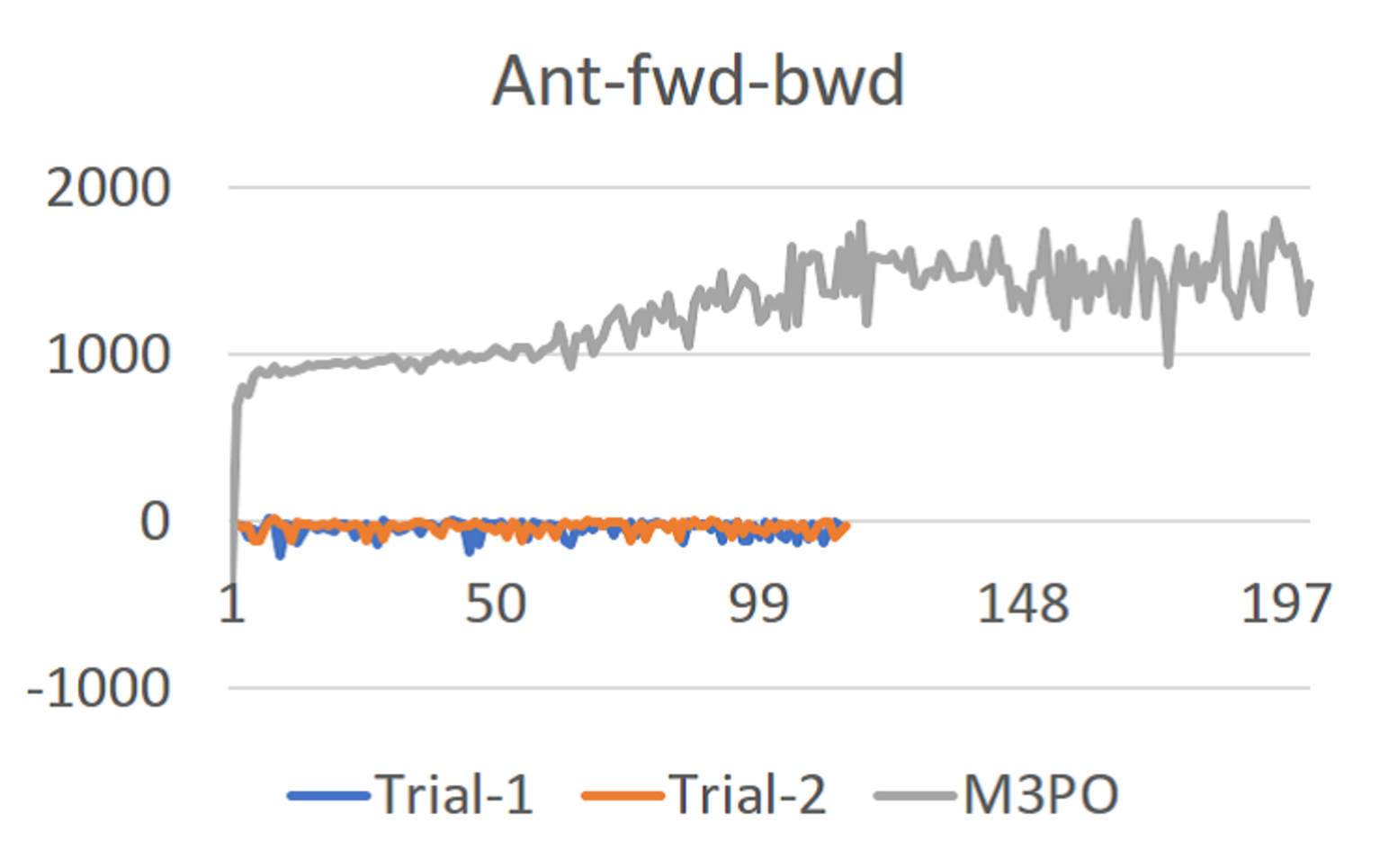}\vline
      \includegraphics[clip, width=0.329\hsize]{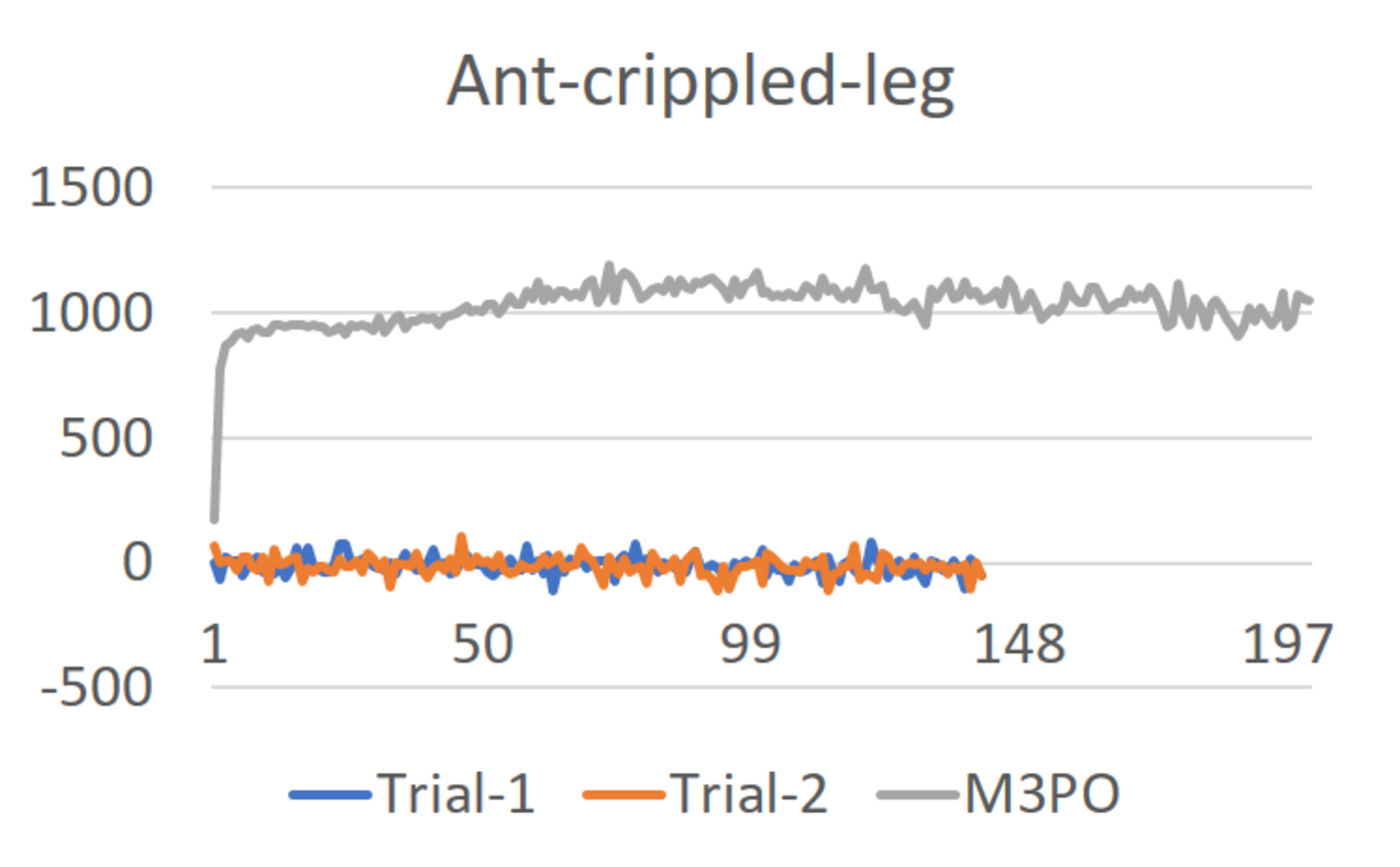}\vline
      \includegraphics[clip, width=0.329\hsize]{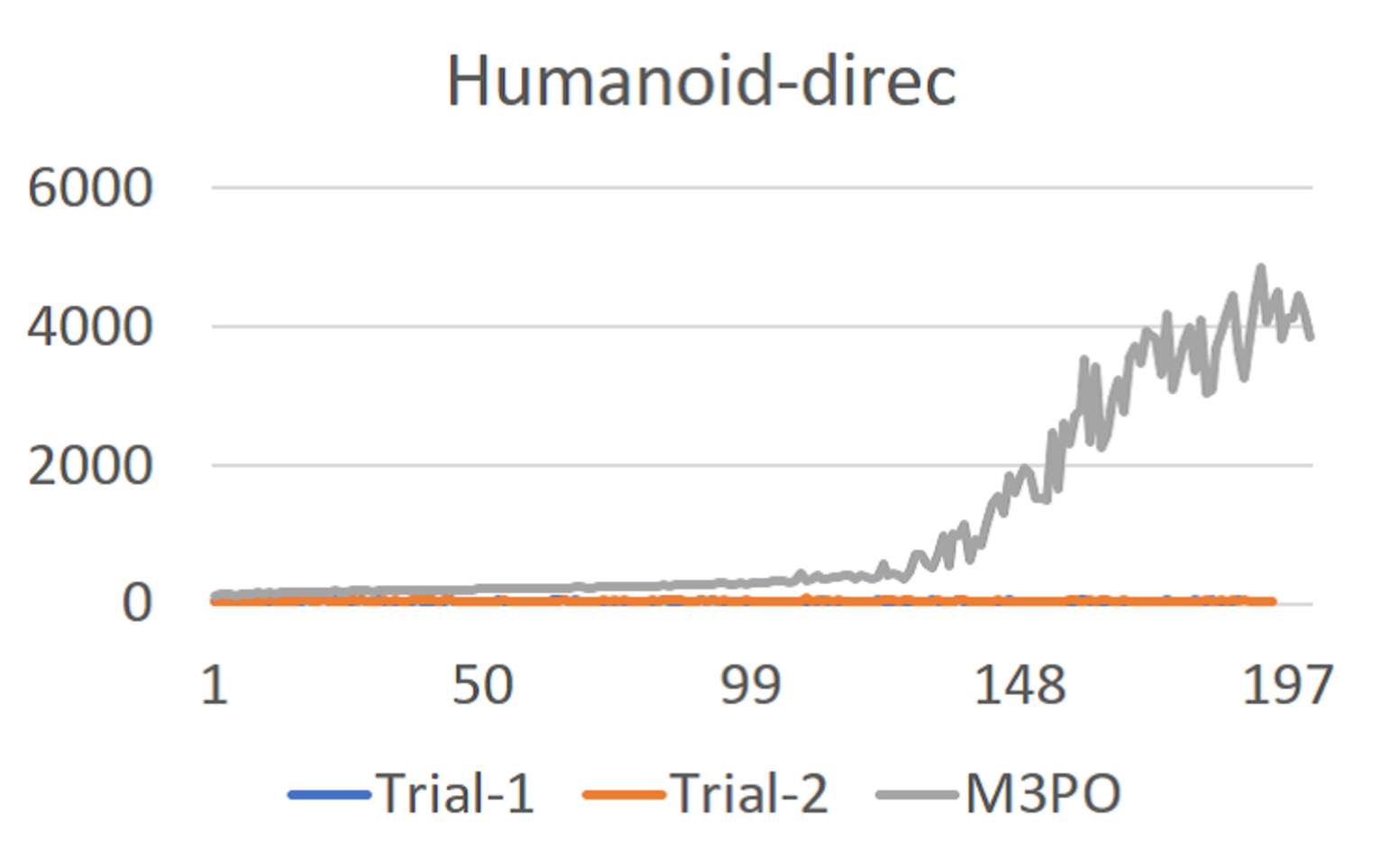}
\end{minipage}
\end{tabular}
\end{center}
\caption{Comparison of GHP-MDP (Algorithm 1 in \citet{perez2020generalized}) and M3PO. The figures show learning curves of GHP-MDP and M3PO. In each figure, the vertical axis represents expected returns and the horizontal axis represents the number of training samples (x1000). GHP-MDP was evaluated in two trials, and each trial was run for three days in real-times. Due to the limitation of computational resources, we could not run this experiment as many days and trials as other experiments. The expected returns of GHP-MDP in each trial (denoted by ``Trial-1'' and ``Trial-2'') are plotted in the figure. The results of M3PO is referred to those in Figure~\ref{fig:lc}. From the comparison result, we can see that M3PO achieves better sample efficiency than GHP-MDP.}
\label{fig:ghpmdp}
\end{figure}

\clearpage

\subsection{Hyperparameter setting}\label{sec:hypara}

\begin{table}[h!]
    \caption{Hyperparameter settings for M3PO results shown in Figure~\ref{fig:lc}.}\label{tab:hparams}
        \begin{center}
            \begin{tabular}{c|c|c|c|c|c|c|c|}\cline{3-8}
                 \multicolumn{2}{c|}{} & \rotatebox{85}{Halfcheetah-fwd-bwd} & \rotatebox{85}{Halfcheetah-pier} & \rotatebox{85}{Ant-fwd-bwd} & \rotatebox{85}{Ant-crippled-leg} & \rotatebox{85}{Walker2D-randomparams} & \rotatebox{85}{Humanoid-direc} \\\cline{2-8}
             $N$ & epoch & \multicolumn{6}{c|}{200} \\\cline{2-8}
             $E$ & environment step per epoch &\multicolumn{6}{c|}{1000} \\\cline{2-8}
             $M$ & model-based rollouts  &  \multicolumn{3}{c|}{1e3} & 5e2 & 1e3 & 5e2 \\%\cline{2-8}
             & per environment step &  \multicolumn{3}{c|}{} & &  &  \\\cline{2-8}
             $B$ & ensemble size  & \multicolumn{6}{c|}{3} \\\cline{2-8}
             $G$ & policy update & \multicolumn{2}{c|}{40} & \multicolumn{4}{c|}{20} \\%\cline{2-8}
            & per environment step &  \multicolumn{2}{c|}{} & \multicolumn{4}{c|}{}  \\\cline{2-8}
             $k$ & model-based rollout length  & \multicolumn{6}{c|}{1} \\\cline{2-8}
             $L$ & history length  &  \multicolumn{6}{c|}{$10$} \\\cline{2-8}
              & network architecture  & \multicolumn{6}{c|}{GRU~\citep{cho2014learning} of five units for RNN. }  \\
              &  & \multicolumn{6}{c|}{MLP of two hidden layers of} \\
              &  & \multicolumn{6}{c|}{400 swish~\citep{ramachandran2017searching} units for FFNN} \\\cline{2-8}
        \end{tabular}
    \end{center}
\end{table}

\clearpage
\subsection{M3PO with hybrid dataset}\label{sec:m3po-h}
%-[introduction]
In Figures \ref{fig:lc} and \ref{fig:unlim}, we can see that, in a number of the environments (Halcheetah-pier, Walker2D-randomparams, and Humanoid-direc), the long-term performance of M3PO is worse than that of PEARL. 
This indicates that a gradual transition from M3PO to PEARL (or other model-free approaches) needs to be considered to improve overall performance. 
In this section, we propose to introduce such a gradual transition approach to M3PO and evaluate it on the environments where the long-term performance of M3PO is worse than that of PEARL. 

%-[proposing approach]
For the gradual transition, we introduce a hybrid dataset $\mathcal{D}_{\text{hyb}}$. 
%-[hybrid dataset is nani?]
This contains a mixture of the real trajectories in $\mathcal{D}_{\text{env}}$ and the fictitious trajectories in $\mathcal{D}_{\text{model}}$, on the basis of mixture ratio $\beta \in [0, 1]$. 
Formally, $\mathcal{D}_{\text{mix}}$ is defined as 
\vspace{-0.2\baselineskip}\begin{equation}
\mathcal{D}_{\text{hyb}} = \beta \cdot \mathcal{D}_{\text{model}} + (1 - \beta) \cdot \mathcal{D}_{\text{env}}. 
\end{equation}
%-[how to use the hybrid dataset]
We replace $\mathcal{D}_{\text{model}}$ in line 11 in Algorithm~\ref{alg1:Meta-MBPO} with $\mathcal{D}_{\text{hyb}}$. 
We linearly reduce the value of $\beta$ from 1 to 0 in accordance with the training epoch. 
With this value reduction, the M3PO gradually becomes less dependent on the model and is transitioned to the model-free approach. 
\begin{table}[t]
    \caption{$\beta$ settings for results shown in Figure~\ref{fig:unlim-hybrid}. $a \rightarrow b$ denotes a thresholded linear function, i.e., at epoch $e$, $f(e)=\min(\max(1  - \frac{e-a}{b-a}, 1), 0)$. }\label{tab:expk}
        \begin{center}
            \begin{tabular}{c|c|c|c|c|}\cline{3-5}
                 \multicolumn{2}{c|}{} & \rotatebox{85}{Halfcheetah-pier} & \rotatebox{85}{Walker2D-randomparams} & \rotatebox{85}{Humanoid-direc} \\\cline{2-5}
             $\beta$ & mixture ratio  & $80\rightarrow 130$ & $50\rightarrow 100$ & $150\rightarrow 250$ \\\cline{2-5}
        \end{tabular}
    \end{center}
\end{table}

\begin{figure}[h!]
\begin{tabular}{c}
\begin{minipage}{0.329\hsize}
\begin{center}
\includegraphics[clip, width=1.0\hsize]{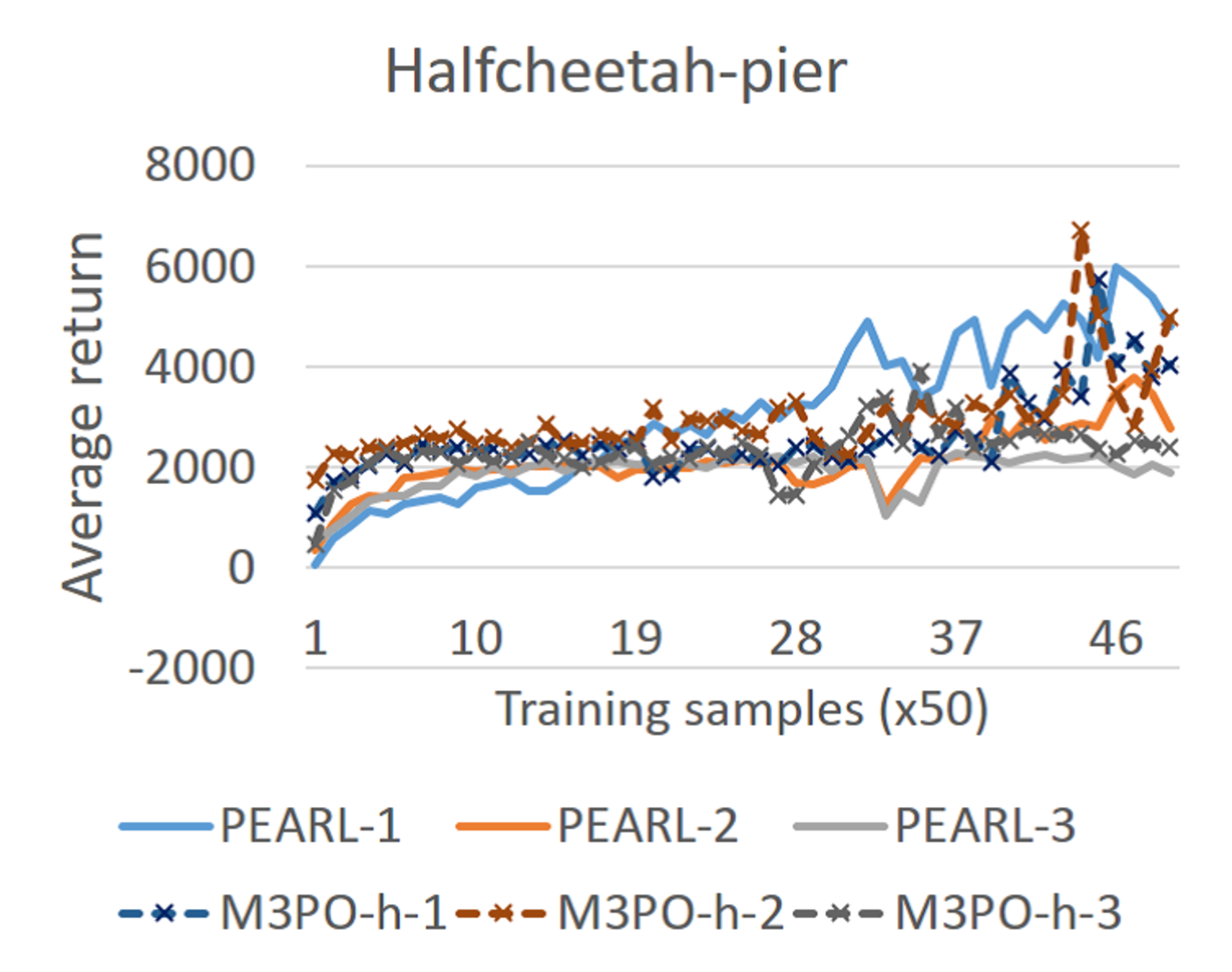}
\end{center}
\end{minipage}
\begin{minipage}{0.329\hsize}
\begin{center}
\includegraphics[clip, width=1.0\hsize]{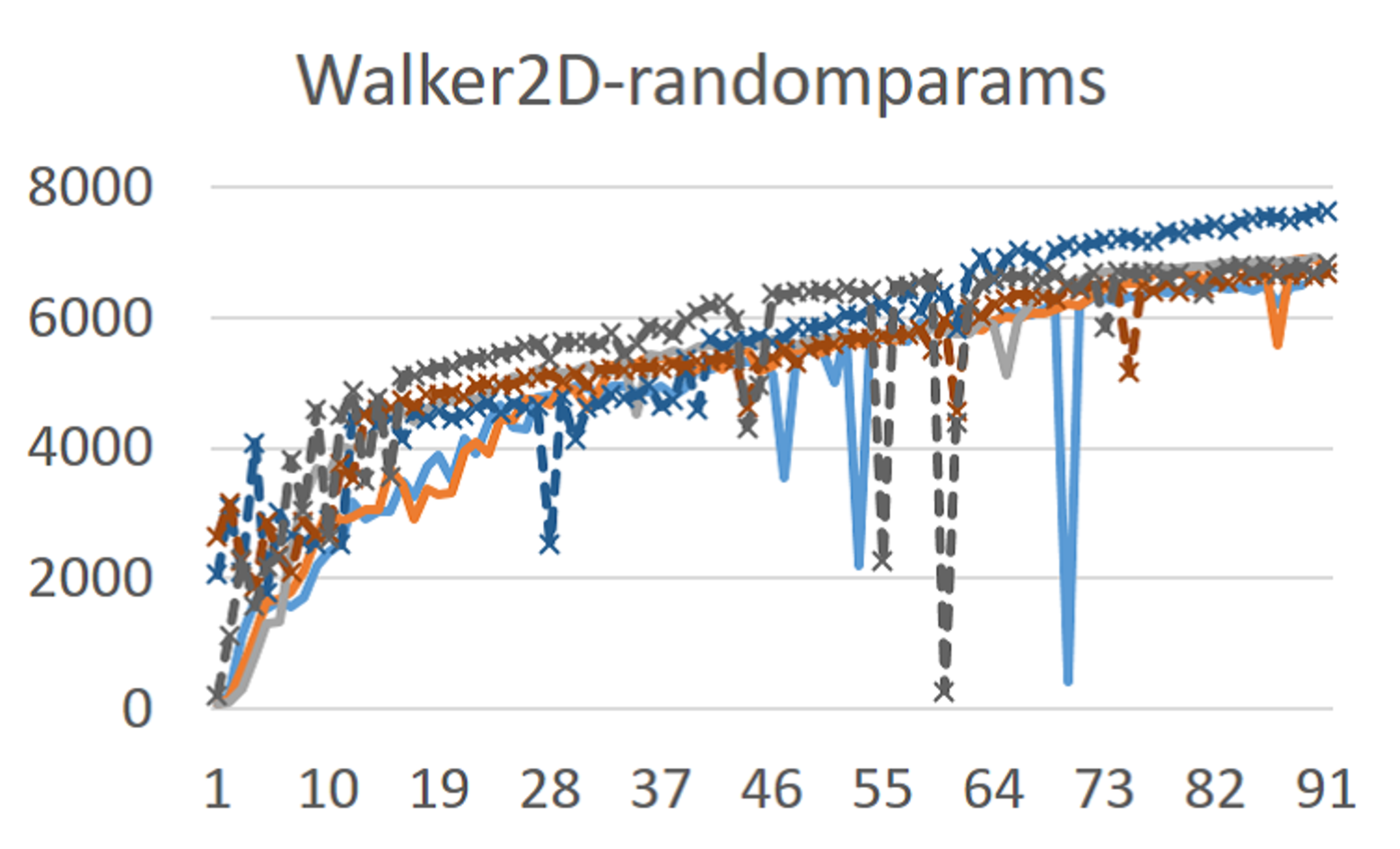}
\end{center}
\end{minipage}
\begin{minipage}{0.329\hsize}
\begin{center}
\includegraphics[clip, width=1.0\hsize]{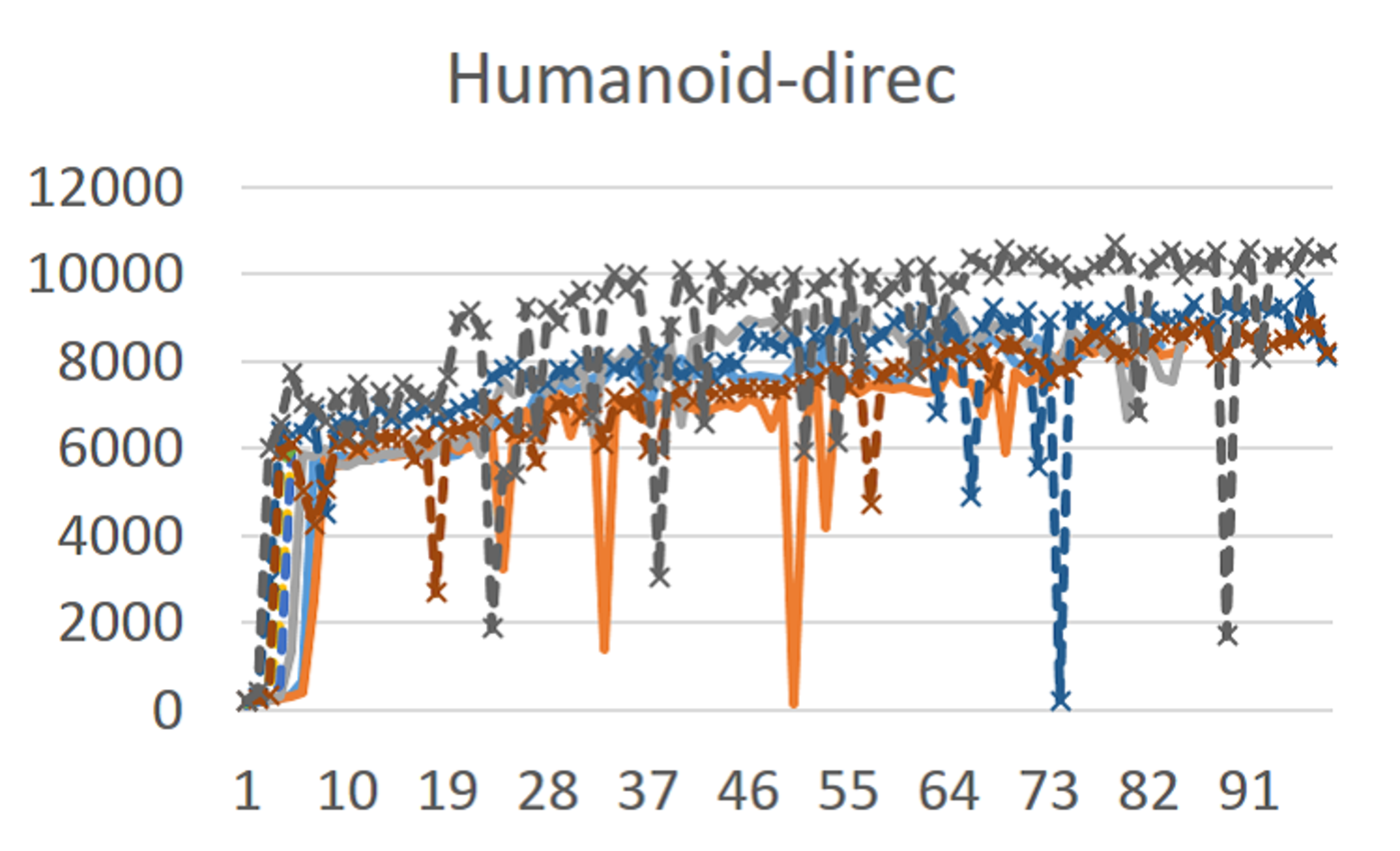}
\end{center}
\end{minipage}
\end{tabular}

\caption{Learning curve of PEARL, M3PO, and M3PO-h in a long-term training. In each figure, the vertical axis represents expected returns and the horizontal axis represents the number of training samples (\textbf{x50000}). The policy and model were fixed and their expected returns were evaluated on 50 episodes at every 50,000 training samples. Each method was evaluated in three trials, and the result of the $x$-th trial is denoted by method-$x$. \textbf{Note that the scale of the horizontal axis is larger than that in Figure~\ref{fig:lc} by 50 times (i.e., 4 in this figure is equal to 200 in Figure~\ref{fig:lc}).} }
\label{fig:unlim-hybrid}
\end{figure}

%-[experiment setting]
We evaluate M3PO with the gradual transition (M3PO-h) in three environments (Halfcheetah-pier, Walker2D-randomparams and Humanoid-direc), in which the long-term performance of M3PO is worse than that of PEARL in Figures~\ref{fig:lc} and \ref{fig:unlim}. 
The hyperparameter setting (except for the setting schedule for the value of $\beta$) for the experiment is the same as that for Figures~\ref{fig:lc} and \ref{fig:unlim} (i.e., the one shown in Table~\ref{tab:hparams}). 
Regarding the setting schedule for the value of $\beta$, we reduce it in accordance with Table~\ref{tab:expk}. 
%-[result]
Evaluation results are shown in Figure~\ref{fig:unlim-hybrid}. 
We can see that M3PO-h achieves the same or better scores with the long-term performances of PEARL in all environments. 

\clearpage
\subsection{The effect of model adaptation}
%-[intro]
In this section, we present a complementary analysis to answer the question ``Does model adaptation (i.e., the use of a meta-model) in M3PO contribute to the improvement of the meta-policy?''. 

%-[method]
We compare M3PO with \textbf{Model-based Meta-Policy Optimization (M2PO)}. 
M2PO is a variant of M3PO in which a non-adaptive transition model is used instead of the meta-model. 
The model architecture is the same as that in the MBPO algorithm~\citep{janner2019trust} (i.e., the ensemble of Gaussian distributions based on four-layer feed-forward neural networks). 

\begin{figure}[h!]
\begin{center}
\begin{tabular}{c}
\begin{minipage}{1.0\hsize}
      \includegraphics[clip, width=0.329\hsize]{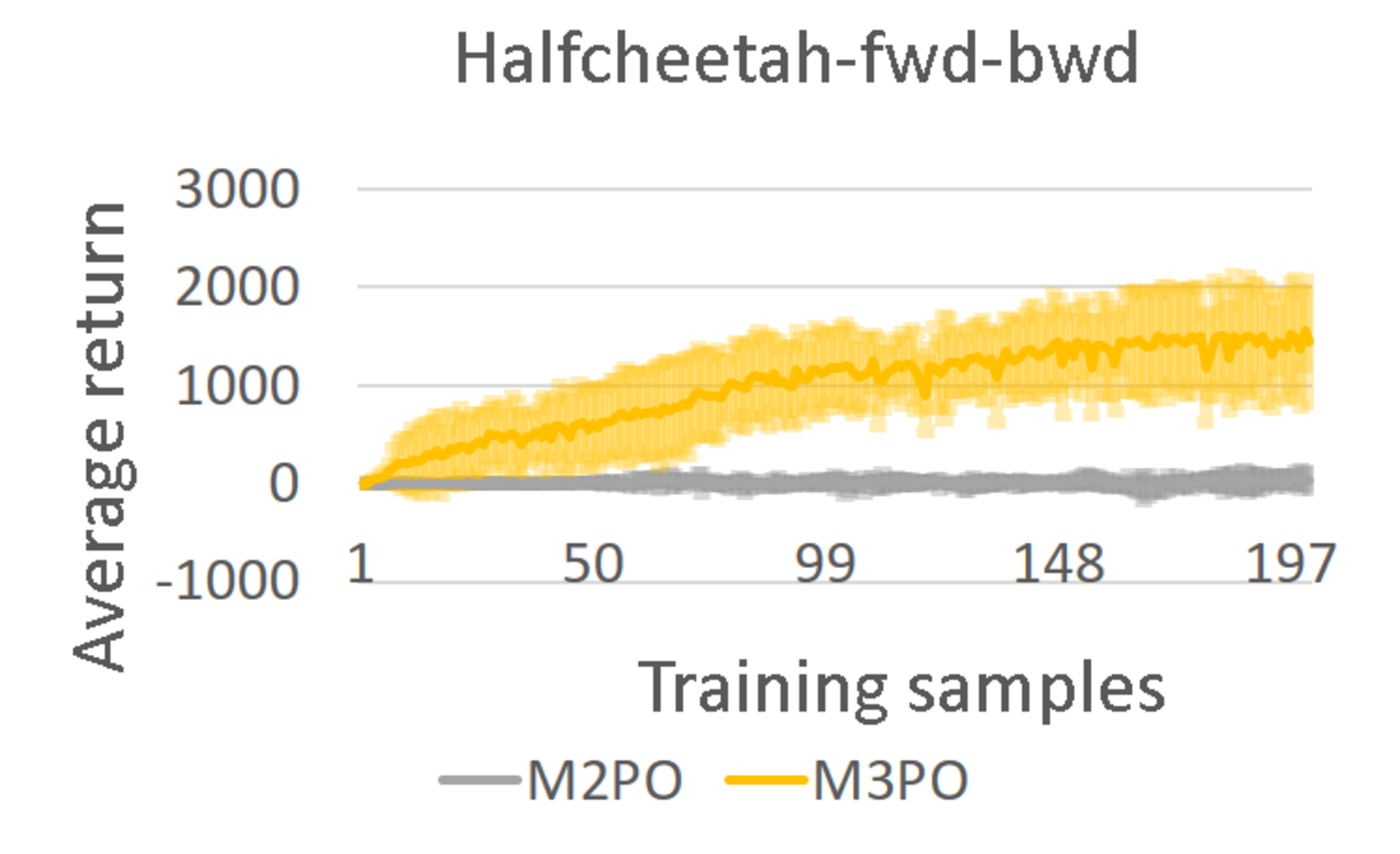}\vline
      \includegraphics[clip, width=0.329\hsize]{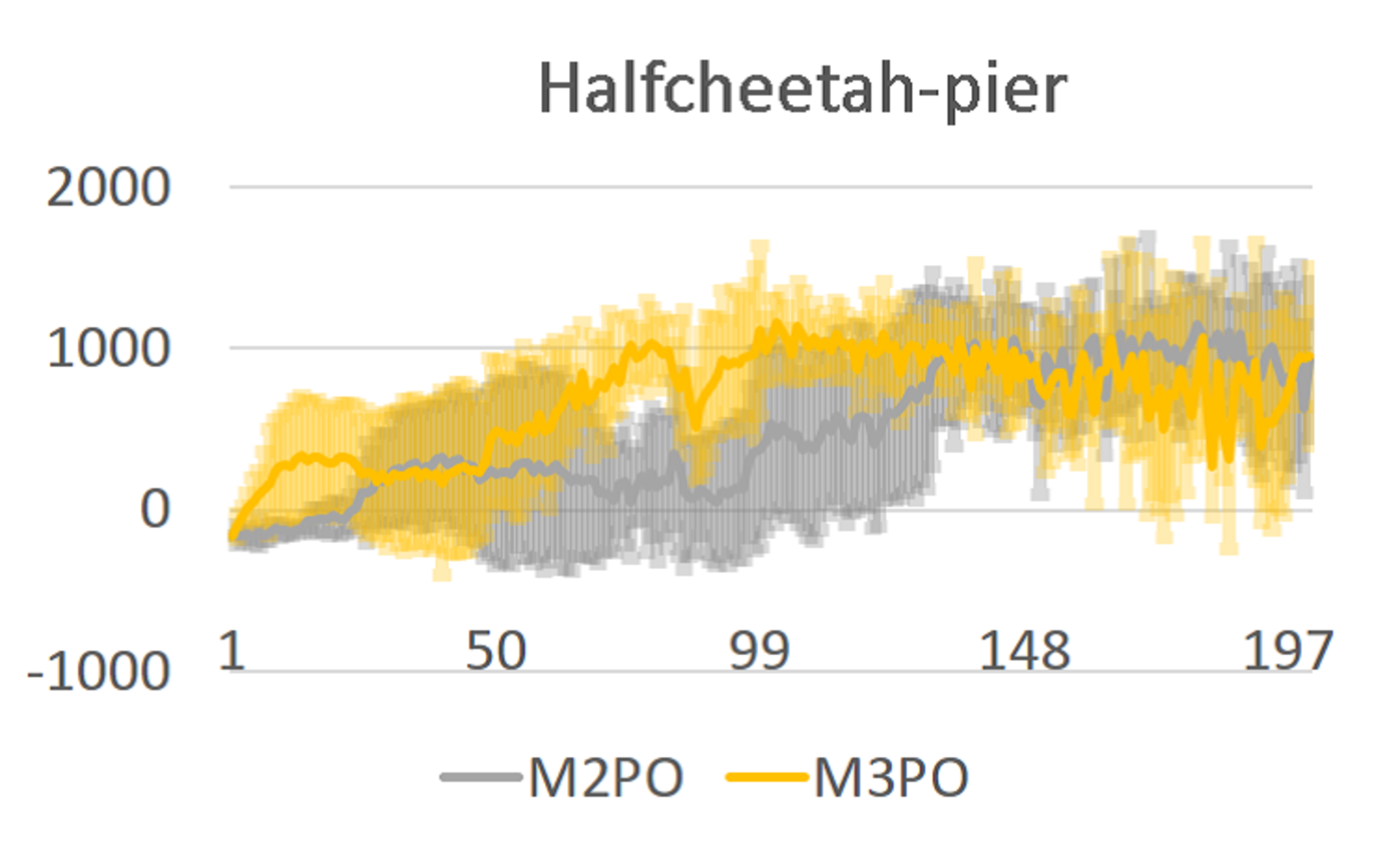}\vline
      \includegraphics[clip, width=0.329\hsize]{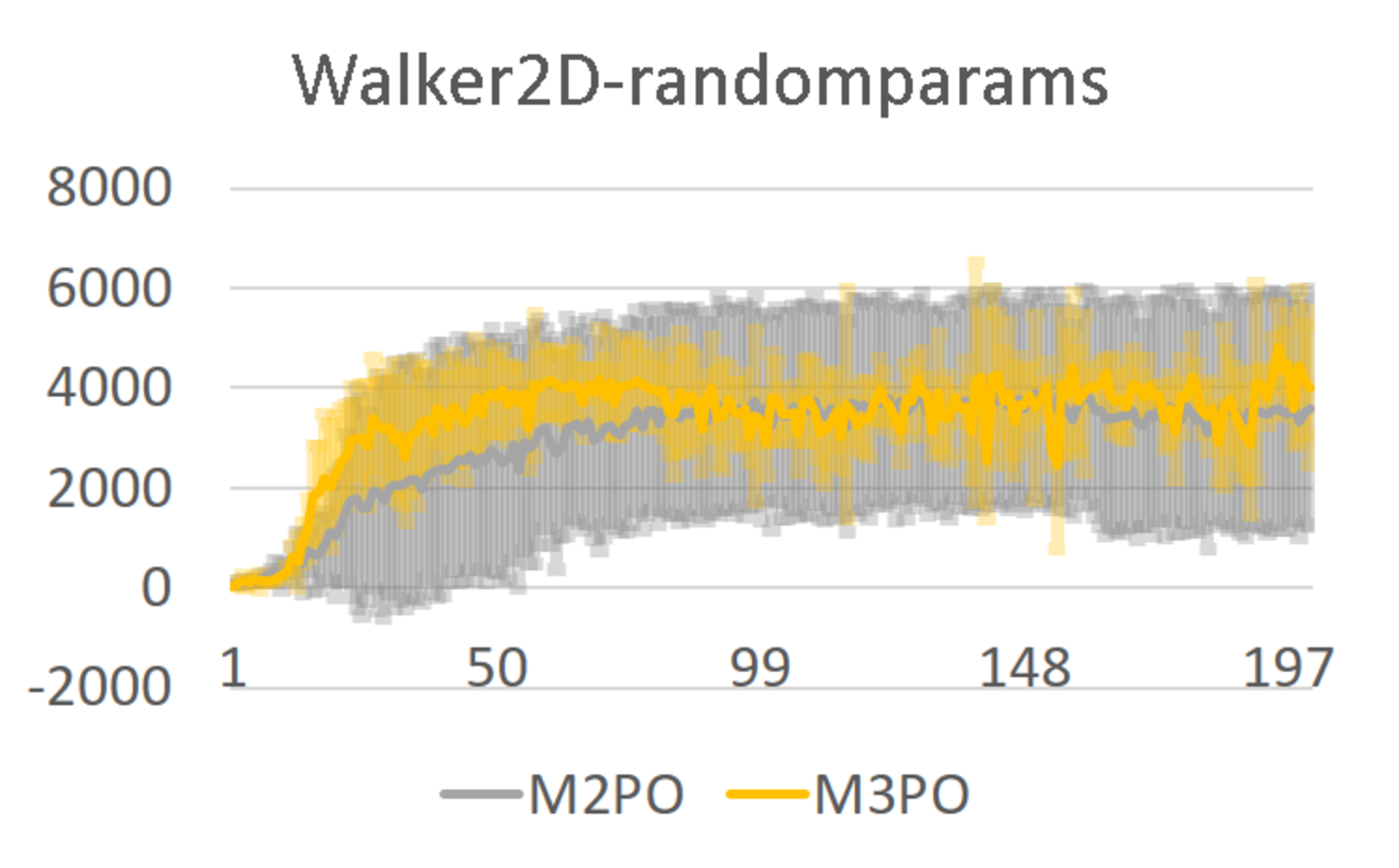}
\end{minipage}\\\hline
\begin{minipage}{1.0\hsize}
      \includegraphics[clip, width=0.329\hsize]{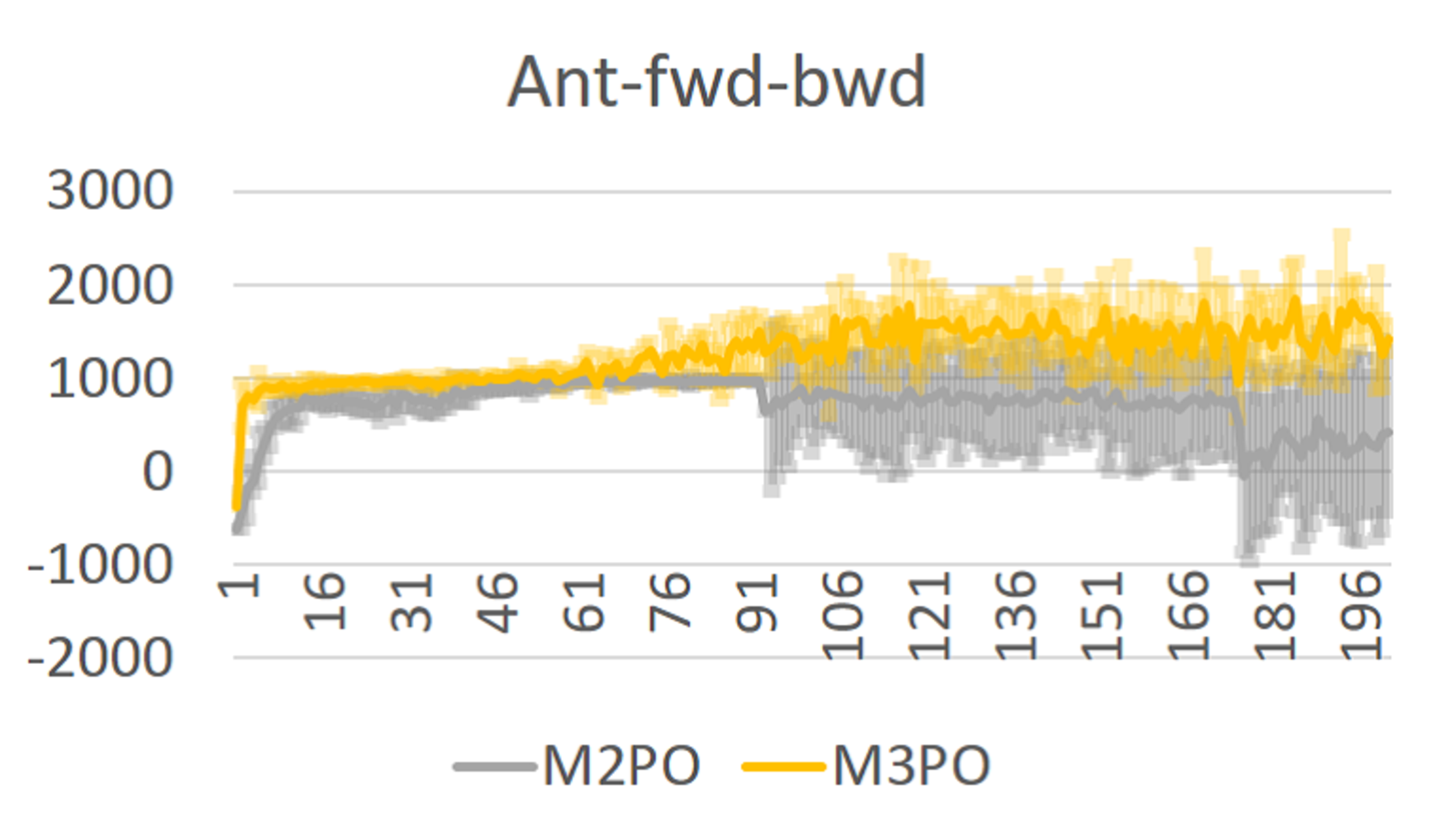}\vline
      \includegraphics[clip, width=0.329\hsize]{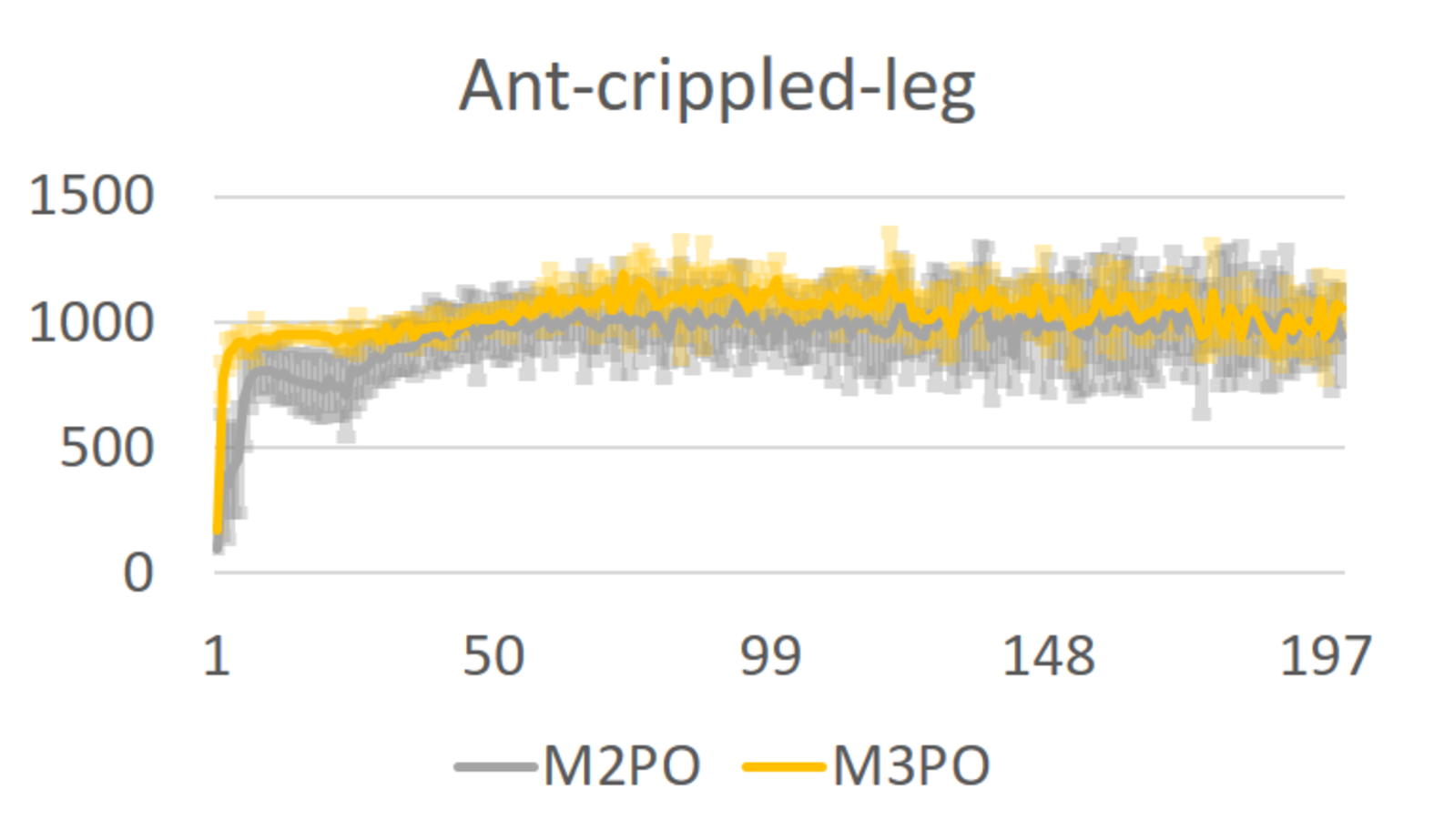}\vline
      \includegraphics[clip, width=0.329\hsize]{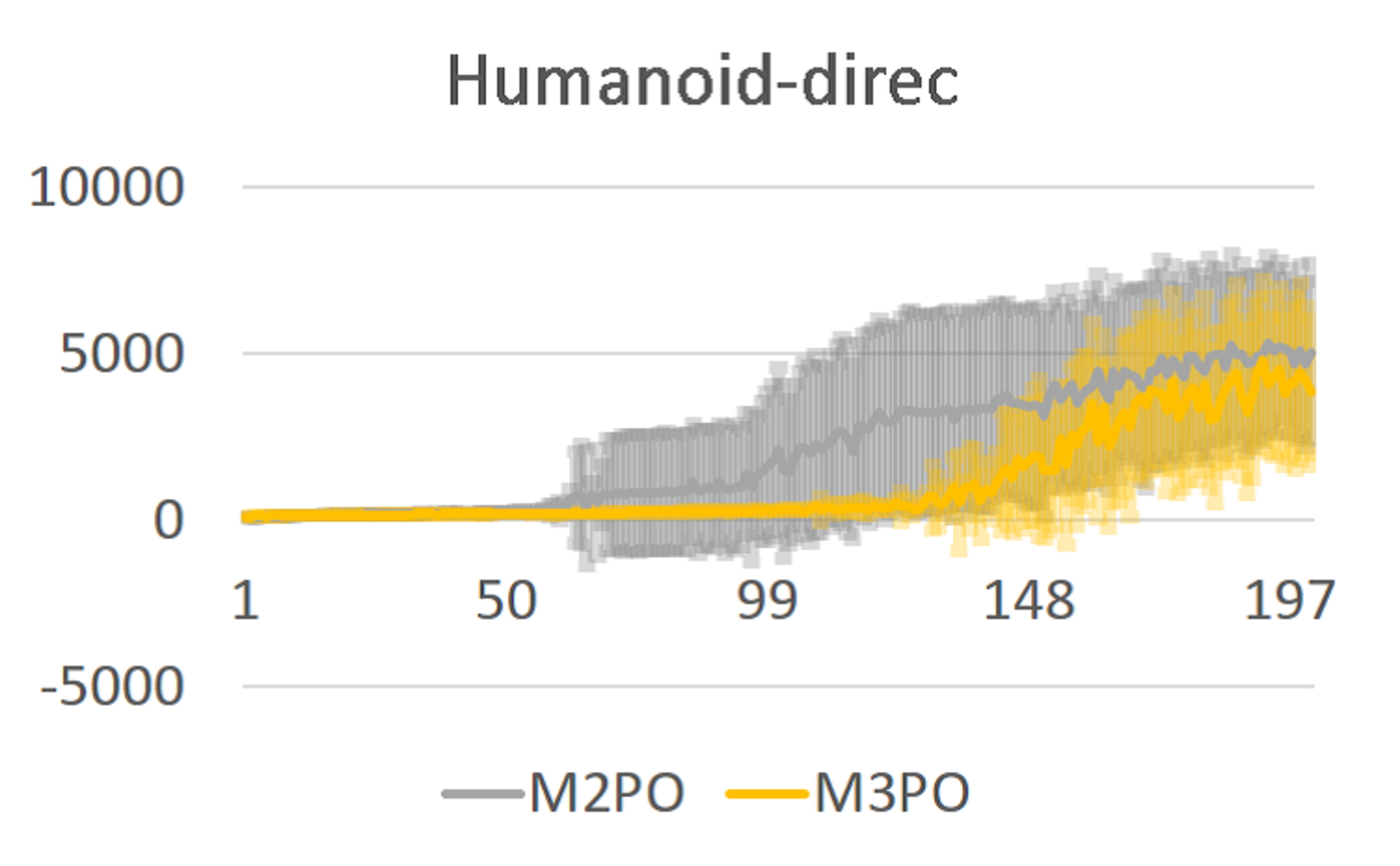}
\end{minipage}
\end{tabular}
\end{center}
\vspace{-0.8\baselineskip}
\caption{The learning curve of M3PO and M2PO. In each figure, the vertical axis represents expected returns and the horizontal axis represents the number of training samples (x1000). The meta-policy and models were fixed and their expected returns were evaluated on 50 episodes at every 1000 training samples for the other methods. In each episode, the task was initialized and changed randomly. Each method was evaluated in at least five trials, and the expected return on the 50 episodes was further averaged over the trials. The averaged expected returns and their standard deviations are plotted in the figures.}
\label{fig:lc-m2po}
\vspace{0\baselineskip}
\end{figure}

%-[result]
Our experimental result indicates that the use of a meta-model contributes to performance improvement in some of the environments. 
In Figure~\ref{fig:lc-m2po}, we can clearly see the improvement of M3PO against M2PO in Halfcheetah-fwd-bwd. 
In addition, in the Ant environments, although the M3PO's performance is seemingly the same as that of M2PO, the qualitative performance is quite different; 
the M3PO can produce a meta-policy for walking in the correct direction, while M2PO failed to do so (M2PO produces the meta-policy ``always standing'' with a very small amount of control signal). 
For Humanoid-direc, by contrast, M2PO tends to achieve better sample efficiency than M3PO. 
We hypothesize that the primary reason for this is that during the plateau at the early stage of training in Humanoid-direc, the model used in M2PO generates fictitious trajectories that make meta-policy optimization more stable. 
To verify this hypothesis, we compare TD-errors (Q-function errors), which are an indicator of the stability of meta-policy optimization, for M3PO and M2PO. The evaluation result (Figure~\ref{fig:td} in the appendix) shows that during the performance plateau (10--60 epoch), the TD-error in M2PO was actually lower than that in M3PO; this result supports our hypothesis. 
In this paper, we did not focus on the study of meta-model usage to generate the trajectories that make meta-policy optimization stable, but this experimental result indicates that such a study is important for further improving M3PO. 
\begin{figure}[h!]%[0pt]{0.4\textwidth}
  \centering
 \includegraphics[width=0.85\textwidth]{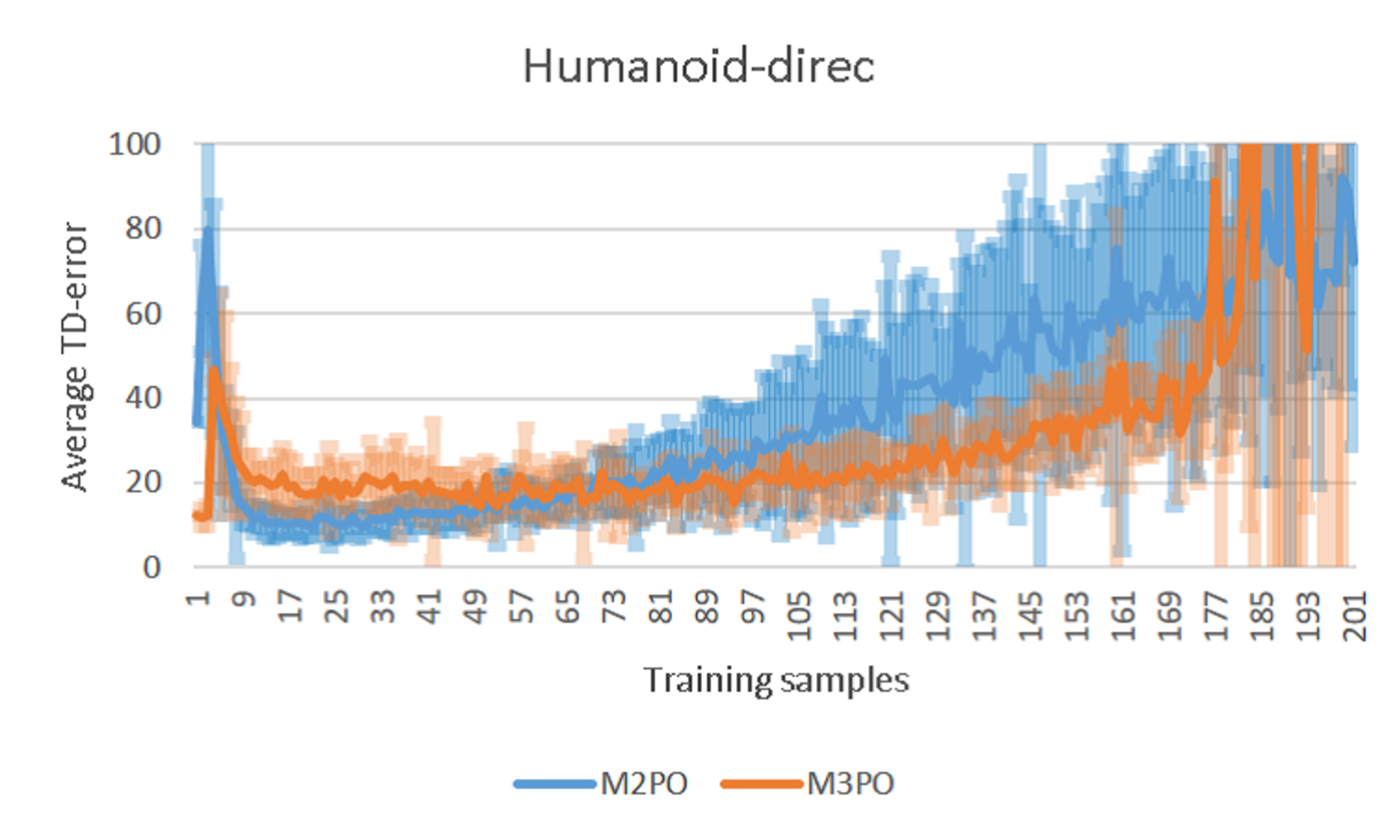}
\vspace{-0.5\baselineskip}
\caption{The transition of TD-errors (Q-function error) on training. In each figure, the vertical axis represents TD-error and the horizontal axis represents the number of training samples (x1000). We ran ten trials with different random seeds and plotted the average of their results. The error bar means one standard deviation.}\label{fig:td}
\end{figure}

\end{document}